\documentclass[12pt]{article}
\usepackage{amsmath,amssymb,amsthm,amsfonts,bm,mathrsfs,float,algorithm,algorithmic,caption,subcaption,setspace,comment}
\usepackage{avant,array,geometry,fontenc,inputenc,natbib,hyperref,multirow,enumerate,rotating,booktabs,color,latexsym,times,stmaryrd,authblk,xr}
\usepackage{chngcntr}
\counterwithin{figure}{section}
\newtheorem{lemma}{Lemma}[section]
\newtheorem{theorem}{Theorem}[section]
\newtheorem{remark}{Remark}

\numberwithin{equation}{section}

\usepackage{hyperref}
\hypersetup{colorlinks=true,
	linkcolor={magenta},
    filecolor={cyan},
	citecolor={blue},
}
\newcommand{\Mean}{{\mathbb{E}}}
\newcommand{\Var}{{\mbox{Var}}}

\newcommand{\prob}{{\mathbb{P}}}

\let\code=\texttt

\DeclareMathOperator*{\argmin}{arg\,min}
\DeclareMathOperator*{\argmax}{arg\,max}

\def\floor#1{\lfloor #1 \rfloor}

\newcommand{\blind}{1}

\newcommand{\change}[1]{{\leavevmode\color{black}{#1}}}

\renewcommand\thetable{\thesection.\arabic{table}}

\usepackage{xcite}
\usepackage{styles}
\usepackage{nomencl}

\makeatletter
\newcommand*{\addFileDependency}[1]{% argument=file name and extension
  \typeout{(#1)}% latexmk will find this if $recorder=0 (however, in that case, it will ignore #1 if it is a .aux or .pdf file etc and it exists! if it doesn't exist, it will appear in the list of dependents regardless)
  \@addtofilelist{#1}% if you want it to appear in \listfiles, not really necessary and latexmk doesn't use this
  \IfFileExists{#1}{}{\typeout{No file #1.}}% latexmk will find this message if #1 doesn't exist (yet)
}
\makeatother

\begin{document}

\if1\blind
{
\title{\Large{\textbf{Testing Stationarity and Change Point Detection in Reinforcement Learning}}}
\author[1]{Mengbing Li}
\author[2]{Chengchun Shi}
\author[3]{Zhenke Wu}
\author[4]{Piotr Fryzlewicz}
\affil[1,3]{University of Michigan, Ann Arbor}
\affil[2,4]{London School of Economics and Political Science}
\date{\empty}
\date{}
\maketitle
} \fi

\if0\blind
{
\title{\Large{\textbf{Testing Nonstationary in Reinforcement Learning}}}
\author{
\bigskip
\vspace{0.5in}
}
\date{}
\maketitle
} \fi

\begin{abstract}
We consider reinforcement learning (RL) in possibly nonstationary environments. Many existing RL algorithms in the literature rely on the stationarity assumption that requires the state transition and reward functions to be constant over time. However, this assumption is restrictive in practice and is likely to be violated in a number of applications, including traffic signal control, robotics and mobile health. In this paper, we develop a model-free test to assess the stationarity of the optimal Q-function based on pre-collected historical data, without additional online data collection. Based on the proposed test, we further develop a change point detection method that can be naturally coupled with existing state-of-the-art RL methods designed in stationary environments for online policy optimization in nonstationary environments. The usefulness of our method is illustrated by theoretical results, simulation studies, and a  real data example from the 2018 Intern Health Study. A Python implementation of the proposed procedure is publicly available at \url{https://github.com/limengbinggz/CUSUM-RL}. 
\end{abstract}


\section{Introduction}\label{sec:introduction}
Reinforcement learning \citep[RL, see][for an overview]{Sutton2018} is a powerful machine learning technique that allows an agent to learn and interact with a given environment, to maximize the cumulative reward the agent receives. It has been one of the most popular research topics in the machine learning and computer science literature over the past few years. 
Significant progress has been made in solving challenging problems across various domains using RL, including games, recommender systems, finance, healthcare, robotics, transportation, among many others \citep[see][for an overview]{li2019reinforcement}. 
In contrast, statistics as a field has only begun to engage with RL both in depth and in breadth. 
Most RL works in the statistics literature focused on developing data-driven methodologies for precision medicine 
\citep[see e.g.,][]{Murphy2003,robins2004,chakraborty2010inference,qian2011,zhang2013robust,zhao2015,Wallace2015,song2015penalized,Alex2016,zhu2017greedy,zhangyc2016,shi2018maximin,wang2018quantile,qi2020multi,nie2020learning,fang2023fairness}. See also \citet{tsiatis2019dynamic} and \citet{kosorok2019precision} for overviews. These aforementioned methods were primarily motivated by applications in precision medicine with only a few treatment stages. They require a large number of patients in the observed data to achieve consistent estimation and become ineffective in the long- or infinite-horizon setting where the number of decision stages diverges with the number of observations. The latter setting is widely studied in the RL literature to formulate many sequential decision making problems in games, robotics, ridesharing, etc. Recently, several algorithms have been proposed in the statistics literature for policy optimization or evaluation in long horizon settings \citep{Ertefaie2018,liao2019off,luckett2019,hu2021personalized,ramprasad2021online,liao2020batch,shi2022statistical,hu2023off,liu2023online,wang2023projected,chen2024reinforcement,li2024settling,zhou2024estimating}. 

%In the computer 

%In this paper, we focus on developing RL methods in offline nonstationary environments. 
Central to the empirical validity of most existing state-of-the-art RL algorithms is the stationarity assumption that requires the state transition and reward functions %, and the policy, 
to be constant functions of time. Although this assumption is valid in online video games, it can be violated in a number of other applications, including traffic signal control \citep{padakandla2020reinforcement}, robotic navigation \citep{niroui2019deep}, mobile health \citep[mHealth,][]{liao2020personalized}, e-commerce \citep{chen2020dynamic} and infectious disease control \citep{cazelles2018accounting}. According to \citet{Sutton2018}, ``{\textit{nonstationarity is the case most commonly encountered in reinforcement learning}}''. It is also a key challenge in lifelong RL where the tasks presented to the agent change over time \citep{silver2013lifelong}. 
We consider a few examples to elaborate the violation of the stationarity assumption. 
%the following example from mobile health applications to elaborate. 

%domains such as mobile health (mHealth), a rapidly expanding field that proposes to provide healthcare support and treatment interventions via mobile technologies such as smartphones, tablets and wearables. Advances in mobile technology allows us to collect rich longitudinal data that can be used to estimate the optimal individualized treatment intervention. %Different from the finite horizon setting, the number of treatment decision stages in not necessarily fixed. 
One motivating example considered in our paper comes from the Intern Health Study \citep[IHS;][]{necamp2020assessing}. The period of medical internship, which marks the initial phase of professional medical training in the United States, is a highly demanding and stressful time for physicians. During this phase, residents are confronted with challenging decisions, extended work hours, and sleep deprivation. In this ongoing prospective longitudinal study, one primary objective is to determine the optimal timing for providing smartphone-delivered interventions. These interventions send mobile prompts through a customized study app, aimed at offering timely tips to encourage interns to adopt anti-sedentary routines that may enhance their physical well-being.
Nonstationarity poses a significant challenge within the context of the mHealth study. Specifically, as individuals receive mobile-delivered interventions for longer duration, they may habituate to the prompts or become overwhelmed, resulting in reduced responsiveness to the contents of the suggestions \citep{klasnja2019efficacy,qian2022microrandomized}. Consequently, the treatment effect of activity suggestions may transition from positive to negative over time. To maximize the effectiveness of interventions, ideal treatment policies would be those adapting swiftly to the current circumstances of the subjects based on the most recent data collected. Failing to recognize potential nonstationarity in treatment effects over time may lead to policies that overburden medical interns, resulting in app deletion and study dropouts.

As another example, the coronavirus disease 2019 (COVID-19) emerged as one of the most severe global pandemics in history and infected hundreds of millions of people worldwide. 
In response to this crisis, there was a growing interest in utilizing RL to develop data-driven intervention policies to contain the spread of the virus \citep[see e.g.,][]{eftekhari2020markovian,kompella2020reinforcement,wan2020multi}. However, the dynamics of COVID-19 transmission were highly intricate and exhibited nonstationary over time. 
Initially, strict lockdown measures were proven to be highly effective in controlling the spread of the virus \citep{kharroubi2020lockdown}. However, these measures had significant economic costs \citep{eichenbaum2020macroeconomics}. As effective vaccines were developed and a substantial proportion of the population became vaccinated, a natural inclination was to ease these lockdown restrictions. Nonetheless, the efficacy of the vaccine was likely to diminish over time, particularly in the presence of new viral variants. To summarize, policymakers needed to dynamically adapt public health policies by taking the nonstationary nature of the COVID-19 spread into consideration, in order to enhance global health outcomes while carefully balancing the negative impacts on the economy and society.

In this paper, we consider situations where the optimal Q-function $Q^{opt}$ (the expected cumulative reward under the optimal policy, see Section \ref{sec:Q} for its detailed definition) is possibly nonstationary, as a result of potential %nonstationarity of the environment,  %including the state transition or reward distribution. 
%such as 
changes in the state transition or reward functions. These functions can change at an arbitrarily unknown time point (referred to as a change point), and the changes can be abrupt or smooth. A number of time series works have focused on testing the stationarity of a given time series and detecting the change point locations in various models, ranging from the simple piecewise-constant signal plus noise setting \citep{kfe12, f14} to complex high-dimensional panel data and time series \citep{cf15, ws18}; see \cite{ac17} and \cite{tov20} for comprehensive reviews. Different from the
aforementioned works in time series, $Q^{opt}$ takes %some value of 
a state-action pair as input. To test its stationarity, we need to check whether it is constant over time, for each possible value of the state-action pair. %Unlike these previous works, we focus on the nonstationarity of functions over time, as opposed to scalar or vector entities like time series. 
% This paper is concerned with offline RL in possibly nonstationary environments.

Our methodological contributions are summarized as follows:
\begin{enumerate}
\item We propose a novel %model-free 
test to assess the stationarity of the optimal Q-function. %(the expected cumulative reward under the optimal policy, see Section \ref{sec:Q} for its detailed definition) policy. 
To the best of our knowledge, this is the first work on developing statistically sound tests for stationarity in offline RL -- a domain where policies are learned from previously collected datasets, instead of actively collected data in real-time as in online RL. Our proposal is an example of harnessing the power of classical statistical inferential tools such as hypothesis testing to help address an important practical issue in RL.

%\item We use the method of sieves \citep{grenander1981abstract} to model the optimal Q-function, to address the challenge of examining whether the optimal Q-function remains stationary over time for each possible state-action value. We construct CUSUM-type test statistics and employ multiplier bootstrap to obtain the critical values. 
%Our method efficiently tests the  stationarity of the optimal Q-function without visiting each state-action pair.

\item We apply our proposed test to compute $p$-values and identify the most recent historical change point location from a set of potential change point candidates for subsequent online policy learning in nonstationary environments. 
\end{enumerate}

Our technical contributions are summarized as follows:
\vspace{-0.25cm}
\begin{enumerate}
    \item We present and systematically examine various types of stationarity assumptions, analyzing their interrelationships. 
    \item We establish the size and power properties of the proposed tests under a bidirectional asymptotic framework that allows either the number of data trajectories or the number of decision points per trajectory to diverge to infinity. This is useful for different types of applications. For example,
    disease management studies using registry data  \citep[e.g.,][]{cooperberg2004contemporary} or infectious disease control studies  \citep[e.g.,][]{lopes2020protocol} often
    % many mobile health studies 
    involve a large number of subjects and the objective is to develop an optimal policy at the population level to maximize the overall long-term reward.
    % , as in our real data application. 
    Conversely, there are other applications where the number of subjects is limited yet the number of decision points is large \citep[see e.g.,][]{marling2018ohiot1dm}. %as in our real data application.
    \item We develop a matrix concentration inequality for nonstationarity Markov decision processes %(see Lemma \ref{lemmamatrixnonstat} in the Supplementary Materials) 
    in order to establish the consistency of the proposed test. The derivation is non-trivial and naively applying the concentration inequality designed for scalar random variables \citep{alquier2019exponential} would yield a loose error bound; see Section \ref{sec:prooflemma2} of the Supplementary Materials for detailed illustrations. 
    \item We derive the limiting distribution of the estimated optimal Q-function computed via the fitted Q-iteration \citep[FQI,][]{Ernst2005} algorithm, one of the most popular Q-learning type algorithms. See %Theorem \ref{thmQ1} in Section \ref{sec:theoryQ} of the Supplementary Materials; see also 
    Equation \eqref{eqn:linearrep}. 
\end{enumerate}

The proposed test and change point detection procedure are useful in practical situations. In particular, the proposed test is useful in identifying nonstationary environments. Existing RL algorithms designed for stationary environments are no longer valid in nonstationary environments. While some recent proposals \citep{lecarpentier2019non,cheung2020reinforcement,fei2020dynamic,domingues2021kernel,wei2021non,xie2021deep,zhong2021optimistic,feng2023non} allow for nonstationarity, they may be less efficient in stationary settings. By examining the stationarity assumption, our proposed test provides valuable insights into the system's dynamics, enabling practitioners to select the most suitable state-of-the-art RL algorithms for implementation. Specifically, if the stationarity assumption is not rejected, standard RL algorithms (e.g., Q-learning, policy gradient methods) can be implemented to ensure efficiency of policy learning. Otherwise, RL algorithms designed for nonstationary environments, such as those mentioned above, should be preferred. 

Additionally, the proposed change point detection procedure identifies the most recent ``best data segment of stationarity''. It can be integrated with state-of-the-art RL algorithms for online policy learning in nonstationary environments. We apply this procedure to both synthetic and real datasets in Sections \ref{sec:num} and \ref{sec:data}, respectively. Results show that the estimated policy based on our constructed data segment is comparable or often superior to those computed by (i) stationary RL algorithms that ignore nonstationarity, (ii) nonstationary RL algorithms that do not perform change point detection and (iii) nonstationary RL algorithms that employ alternative tests for identifying change points. 
%baseline methods that either ignore nonstationarity, or do not perform change point detection, or employ alternative tests for identifying change points.
%neglect the identification of the best data segment. 
In the motivating IHS study, the proposed method reveals the benefit of nonstationarity detection for optimizing population physical activities for interns in some medical specialties. The promotion of healthy behaviors and the mitigation of negative chronic health outcomes typically require continuous monitoring over a long term where nonstationarity is likely to occur. As RL continues to drive development of optimal interventions in mHealth studies, this paper substantiates the need and effectiveness of incorporating a classical statistical inferential tool to accommodate nonstationarity.

The rest of the paper is organized as follows. In Section \ref{sec:pre}, we introduce the offline RL problem and review some existing algorithms. %In Section \ref{sec:learnnon}, we illustrate our main idea of learning the optimal policy in nonstationary environment. 
In Section \ref{sec:test}, we detail our procedures for hypothesis testing and change point detection. We establish the theoretical properties of our procedure in Section \ref{sec:theory}, conduct simulation studies in Section \ref{sec:num}, and apply the proposed procedure to the IHS data in Section \ref{sec:data}.
%Finally, we conclude the paper in Section \ref{sec:dis}. 

\section{Problem Formulation}\label{sec:pre}
We start by introducing the data generating process and the concept of policy in RL. 
We next review Q-learning \citep{watkins1992q}, a widely used RL algorithm that
is closely related to our proposal. Finally, we discuss five types of stationarity assumptions and introduce our testing hypotheses. 

\subsection{Data}\label{sec:dataintro}%Problem Formulation}
We consider an offline setting to learn an optimal policy based on a pre-collected dataset from a randomized trial or observational study. The offline dataset is summarized as $\mathcal{D} = \{(S_{i,t},A_{i,t},R_{i,t})\}_{1\le i\le N, 0\le t\le T}$, where $(S_{i,t},A_{i,t},R_{i,t})$ denotes the state-action-reward triplet of the $i$th subject at time $t$. Without loss of generality, we assume all subjects share the same termination time $T$, which is reasonable in many mHealth studies. In Section \ref{sec:excvary} of the Supplementary Materials, we extend our proposal to settings where subjects have different termination times.  
In IHS, the state corresponds to some time-varying covariates associated with each medical intern, such as their mood score, step counts, and sleep minutes. The action is a binary variable, corresponding to whether to send a certain text message to the intern or not. The immediate reward is the step counts. 
We assume all rewards are uniformly bounded, as commonly adopted in the RL literature to simplify theoretical analysis \citep[see e.g.,][]{fan2020theoretical}. 
%The trajectories of the $N$ interns $\{(S_{i,t},A_{i,t},R_{i,t})\}_{1\le i\le N, t\ge 0}$ 
These $N$ trajectories are assumed to be i.i.d copies of an infinite horizon Markov decision process \citep[MDP,][]{puterman2014markov} $\{(S_t,A_t,R_t)\}_{t\ge 0}$ %. Specifically, 
whose data generating process can be described as follows: 
\begin{enumerate}
    \item \textbf{State Presentation}: At each time $t$, the environment (represented as an intern in our example) is in a state $S_t\in \mathcal{S}$ where $\mathcal{S}\in \mathbb{R}^d$ denotes the state space and $d$ denotes the dimension.
    \item \textbf{Action Selection}: The agent (or decision maker) then selects an action $A_t$ from the action space $\cA$ based on the observed data history $H_t$ including $S_t$ and the state-action-reward triplets up to time $t-1$.
    \item \textbf{Reward Generation}: The agent receives an immediate reward $R_t\in \mathbb{R}$ with its expected value specified by an unknown reward function $r_t$:
\begin{eqnarray}\label{eqn:reward}
    \Mean (R_t|H_t,A_t)=r_t(A_t,S_t).
\end{eqnarray}
    \item \textbf{State Transition}: Subsequently, the environment transitions to the next state $S_{t+1}$, determined by an unknown transition function $\mathcal{T}_t$:
\begin{eqnarray}\label{eqn:transition}
    S_{t+1}=\mathcal{T}_t(A_t,S_t,\delta_t),
\end{eqnarray}
where $\{\delta_t\}_t$ is a sequence of i.i.d. random noises, with each $\delta_t$ being independent of $\{(S_j,A_j,R_j)\}_{j\le t}$. 
\end{enumerate}
\begin{remark}
By definition, $\mathcal{T}_t$ specifies the conditional distribution of the future state given the current state-action pair, and $r_t$ is the conditional mean function of the reward. 
Both \eqref{eqn:reward} and \eqref{eqn:transition} impose certain Markov or conditional independence assumptions on the data trajectories, implying that the future state and the conditional mean of the current reward are independent of the $H_t$ given the current state-action pair. 
These assumptions are testable from the observed data  \citep{chen2012testing,shi2020does,zhou2023testing}. 
\end{remark}

\subsection{Policy}\label{sec:policy} 
A policy defines how actions are chosen at each decision time. In particular:
% the way that a decision maker chooses an action at each decision time. %Formally speaking, each policy is a map that takes the observed state-action-reward history as input and outputs a probability distribution on the action space. 
\begin{enumerate}
\item A \textbf{history-dependent policy} $\pi$ is a sequence of decision rules $\{\pi_t\}_{t\ge 0}$ such that each $\pi_t$ takes %the observed data history 
$H_t$ as input, and outputs a probability distribution on the action space, denoted by $\pi_t(\bullet|H_t)$. Under $\pi$, the agent will set $A_t = a$ with probability $\pi_t(a|H_t)$ at time $t$. 

\item A \textbf{Markov policy} $\pi$ is a special history-dependent policy where each $\pi_t$ depends on $H_t$ only through the current state $S_t$. %and we use $\pi(\bullet|S_t)$ to denote the resulting decision rule. 

\item A \textbf{stationary policy} $\pi$ is a special Markov policy where $\{\pi_t\}_t$ are time-invariant, i.e., there exists some function $\pi^*$ such that $\pi_t(\bullet|H_t)=\pi^*(\bullet|S_t)$ almost surely for any $t$, and we use $\pi(\bullet|S_t)$ to denote the resulting decision rule.

\item An \textbf{optimal policy} $\pi^{opt}=\{\pi^{opt}_t\}_t$ maximizes the expected $\gamma$-discounted cumulative reward $J_{\gamma}(\pi)=\sum_{t\ge 0} \gamma^t \Mean^{\pi} (R_t)$ among all history-dependent policies $\pi$, given a discount factor $0<\gamma\le 1$ which balances the trade-off between the immediate and future rewards. The expectation $\Mean^{\pi}$ is calculated under the assumption that the agent makes decisions in accordance with the policy $\pi$.

\item The \textbf{behavior policy} $b=\{b_t\}_t$ denotes 
the policy the agent adopted for all individuals in the offline dataset. This policy is not necessarily optimal and may be a %fixed
 purely random policy in scenarios like sequential multiple assignment randomized trials \citep[SMARTs,][]{collins2007multiphase}. 
\end{enumerate}
\begin{remark}
Under \eqref{eqn:reward} and \eqref{eqn:transition}, there exists an optimal Markov policy $\pi^{opt}$ whose $J_{\gamma}(\pi^{opt})$ is no worse than that of any history-dependent policy; see Theorem 6.2.10 of \cite{puterman2014markov}. This substantially simplifies the calculation of the optimal policy. Hence, throughout this paper, `optimal policy' specifically refers to the optimal Markov policy, meaning that each $\pi_t^{opt}$ is a function of the current state $S_t$ only. Note that the proof in \citet{puterman2014markov} relies on the assumption that the reward is a deterministic function of the state-action-next-state triplet. However, this assumption can be effectively relaxed to \eqref{eqn:reward} while still preserving the validity of the proof.
\end{remark}

\subsection{Q-Learning}\label{sec:Q}
We review Q-learning, one of the most popular RL algorithms. It is model-free in that the optimal policy is derived without directly estimating the MDP model (i.e., transition and reward functions). Central to Q-learning is the state-action value function, commonly known as the Q-function. Given a policy $\pi$, its Q-function $Q^{\pi}=\{Q_{t}^{\pi}\}_{t\ge 0}$ is defined such that each $Q_{t}^{\pi}$ is the expected cumulative reward given a state-action pair following $\pi$, i.e., $Q_t^{\pi}(a,s)=\Mean^{\pi}\left(\sum_{k\ge 0}\gamma^k R_{t+k}|A_t=a,S_t=s\right)$.
% \begin{eqnarray*}
%     Q_{t:\infty}^{\pi}(a,s)=\Mean^{\pi}\left(\sum_{k\ge 0}\gamma^k R_{t+k}|A_t=a,S_t=s\right).
% \end{eqnarray*}
%In stationary environments, we have a stationary $Q$-function such that $Q_{t:\infty}^{\pi}=Q^{\pi}$ for any $t$. 
The optimal Q-function, denoted by $Q^{\pi^{\tiny{opt}}}$ or simply $Q^{\tiny{opt}}$, corresponds to the Q-function under the optimal policy. The optimal Q-function possesses two key properties\footnote{Typically, these properties are verified in a stationary setting, see e.g., Theorem 1.8 of \cite{agarwal2019reinforcement}. However, by simply incorporating the time index into the state definition, the proofs of these theorems can be readily adapted to nonstationary MDPs defined in Section \ref{sec:dataintro} as well.}: 
%that satisfies 
%(i) $Q^{\tiny{opt}}(a,s) = \sup\limits_\pi Q^{\pi} (a, s)$ for any $a,s$; %and that %exists. In this case, the optimal policy $\pi^{opt}$ is also stationary and satisfies 
% the optimal policy is stationary and that greedy with respect to $Q^{\tiny{opt}}$, i.e.,
(i) $\pi^{\tiny{opt}}$ is greedy with respect to $Q^{\tiny{opt}}$, i.e., for any $t\ge 0$, 
\begin{eqnarray}\label{eqn:optimalpolicy}
    \pi_t^{\tiny{opt}}(a|s)=\left\{
    \begin{array}{ll}
        1, & \textrm{if~}a=\argmax_{a'} Q_t^{\tiny{opt}}(a',s); \\
        0, & \textrm{otherwise}. 
    \end{array} 
    \right.
\end{eqnarray}
%In addition, $Q^{\tiny{opt}}$ satisfies
(ii) The Bellman optimality equation holds, stating that the expected Q-value at time $t$ equals the immediate reward plus the maximum Q-value of the next state:
\begin{eqnarray}\label{eqn:bell}
	\Mean \left\{ R_t+\gamma \max_{a} Q_{t+1}^{\tiny{opt}}(a,S_{t+1})|A_t, S_t \right\}=Q_t^{\tiny{opt}}(A_t,S_t),\qquad\forall t\ge 0. 
\end{eqnarray}

Equations \eqref{eqn:optimalpolicy} and \eqref{eqn:bell} form the basis of Q-learning, %, one of the most popular algorithms developed under stationarity. It is model-free in the sense that the optimal policy is derived without directly estimating the transition and reward functions. 
which estimates $\{Q_t^{\tiny{opt}}\}_t$ by solving the Bellman optimality equation \eqref{eqn:bell} and computes $\pi^{\tiny{opt}}$ based on the estimated optimal Q-function using \eqref{eqn:optimalpolicy}. Assuming the stationarity of the optimal Q-function, i.e., $Q_t^{opt}=Q^{opt}$ for any $t$, 
various algorithms have been developed under this framework, such as tabular Q-learning \citep{watkins1992q}, fitted Q-iteration (FQI), greedy gradient Q-learning \citep{maei2010,Ertefaie2018}, double Q-learning \citep{hasselt2010double} and deep Q-network \citep[DQN,][]{mnih2015human}.
In particular, FQI, which our paper implements, iteratively updates the optimal Q-function estimator based on \eqref{eqn:bell}. Beginning with an initial Q-function estimator $Q^{(0)}$ (typically set to zero), we compute $Q^{(k+1)}$ by minimizing 
\begin{eqnarray}\label{eqn:updateQinFQI}
    Q^{(k+1)}=\argmin_Q \sum_{i,t} \left\{R_{i,t}+ \gamma \max_a Q^{(k)}(a, S_{i,t+1})-Q(A_{i,t},S_{i,t}) \right\}^2,
\end{eqnarray}
during the $k$th iteration. The above optimization can be cast into a supervised learning problem with $\{R_{i,t}+ \gamma \max_a Q^{(k)}(a, S_{i,t+1})\}_{i,t}$ as the responses and $\{(A_{i,t},S_{i,t})\}_{i,t}$ as the predictors. 
%In Appendix \ref{sec:theoryQ}, 
We will establish the limiting distribution of the resulting Q-function estimator when employing the method of sieves for function approximation. %in a stationary MDP. 

\subsection{The Stationarity Assumption} 
Starting from a given time point $T_0\ge 0$, we introduce five types of stationarity assumptions:
\begin{enumerate}
    \item[\textbf{SA1}] (\textbf{Stationary MDPs}): The transition function $\mathcal{T}_t$ and the reward function $r_t$ remain constant over time, for all $t\ge T_0$.   
    \item[\textbf{SA2}] (\textbf{Stationary Q-functions}): For any stationary policy $\pi$ (see the definition in Section \ref{sec:policy}), the associated Q-function $Q_t^{\pi}$ is constant as a function of $t$, for all $t\ge T_0$. 
    \item[\textbf{SA3}] (\textbf{Stationary optimal Q-functions}):  $Q_{T_0}^{opt}=Q_{T_0+1}^{opt}=\cdots=Q_{t}^{opt}=\cdots$.
    \item[\textbf{SA4}] (\textbf{Stationary optimal policies}):  $\pi_{T_0}^{opt}=\pi_{T_0+1}^{opt}=\cdots=\pi_{t}^{opt}=\cdots$. 
    \item[\textbf{SA5}] (\textbf{Stationary behavior policies}):  $b_{T_0}=b_{T_0+1}=\cdots=b_{t}=\cdots$.
\end{enumerate}
The following theorem discusses the relationships among these assumptions. 
\begin{theorem}[Stationarity relationships]\label{thm1}
    Assume both the state space and the action space are finite, and the rewards are uniformly bounded. Then SA1 implies SA2, SA2 implies SA3, and SA3 implies SA4.
\end{theorem}
\begin{remark}
    Throughout this paper, %the term 
    \textit{stationary MDPs} refer to MDP models with stationary transition and reward functions. Therefore, SA1 represents a ``model-based'' stationarity assumption, as it directly relates to the %se functions. 
    MDP model. It is the most prevalently employed form of stationarity in the literature \citep{Sutton2018}.
    Importantly, this concept does not require the stationarity of the behavior policy (SA5), which characterizes the decisions or strategies put forth by the agent and operates independently of the environmental factors. Nor does SA5 imply SA2 -- SA4. As a result, the optimal policy may be stationary or nonstationary, regardless of whether the behavior policy is stationary or not.
\end{remark}
\begin{remark}
    %According to Theorem \ref{thm1}, the stationary environment assumption SA1 implies the stationarity of the (optimal) Q-function as well as that of the optimal policy. 
    SA2 and SA3 are characterized as model-free stationarity assumptions, as they are defined without direct reference to the state transition or reward functions. Theorem \ref{thm1} suggests that these assumptions are automatically satisfied under SA1. Furthermore, it is well-known that in stationary MDPs, the optimal policy is stationary as well \citep{puterman2014markov}. The proof that SA2 leads to SA3, however, is not straightforward. Drawing inspiration from the policy iteration algorithm \citep[Section 4.3]{Sutton2018}, we define a sequence of policies whose Q-functions converge to the optimal Q-function. This enables us to establish the connection between a non-optimal Q-function and the optimal one, thus proving the stationarity of the optimal Q-function in a potentially nonstationary MDP (i.e., when SA2 holds, but SA1 does not). 
    We refer readers to Appendix \ref{sec:proof:thm1} of the Supplementary Materials for details.
\end{remark}

\subsection{Hypothesis Testing under Nonstationarity}
\begin{figure}[!t]
	\centering
	\includegraphics[width=12cm]{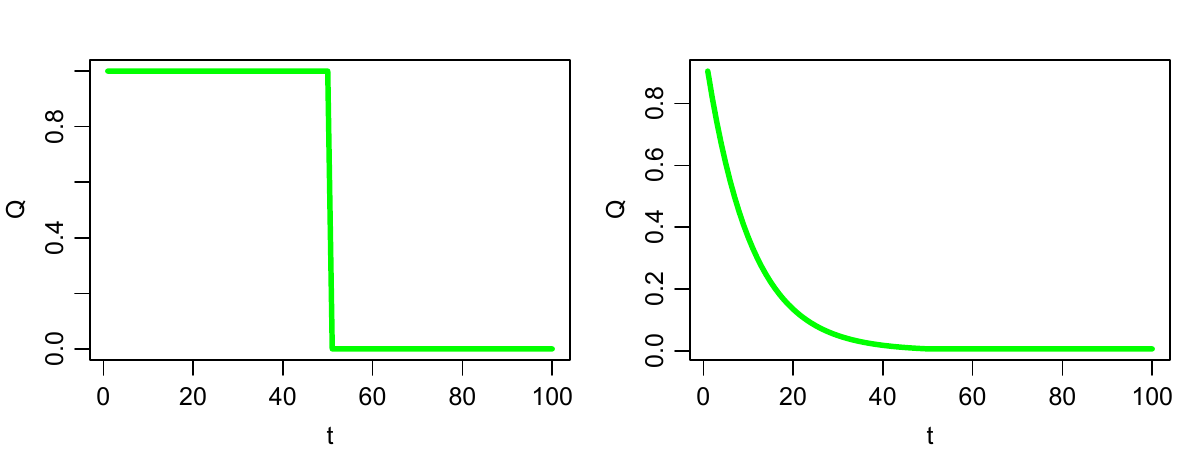}
	\caption{Examples of $Q^{opt}$ at a given state-action pair, with an abrupt change point (left panel) and a gradual change point (right panel) at $t=50$. $T_0=0$ in both examples.}\label{fig1}
\end{figure}

As commented in the introduction, the stationarity assumption can be restrictive in practice. This motivates us to test stationarity based on the offline dataset. Given the five different types of stationarity assumptions we have identified, there are correspondingly five different hypotheses to test. In this paper, we focus on testing SA3 due to the following considerations: 
\begin{itemize}
    \item Testing SA1 poses considerable challenges in scenarios with moderate to high-dimensional states, as the transition function's outputs are multidimensional with dimension matching that of the state.
    \item Testing SA2 is extremely difficult due to the need to enumerate over all possible policies, whose number increases exponentially with the cardinality of the state space.
    \item Based on \eqref{eqn:optimalpolicy}, the optimal policy is intrinsically tied to the optimal Q-function. Theorem \ref{thm1} further implies that SA4 follows from SA3. Thus, by testing SA3, we can effectively assess the stationarity of the optimal policy.
    \item Optimal testing of SA4 is complex since the optimal policy is a highly nonlinear functional of the observed data, complicating the derivation of its estimator's limiting distribution. Moreover, $\pi^{opt}$ may lack uniqueness even when $\{Q_t^{opt}\}_t$ is uniquely determined.
    \item Testing SA5 is meaningless because the stationarity of the behavior policy does not influence that of the optimal policy. 
\end{itemize}
Thus, our testing hypotheses are formulated as follows:
\begin{eqnarray}\label{eqn:H0}
\begin{split}
    \mathcal{H}_0: Q_t^{opt}=Q^{opt},\,\, \forall t\ge T_0, \,\,\textrm{versus}\,\,
    \mathcal{H}_1: Q_t^{opt}\,\textrm{has at least one smooth}\\ \textrm{or abrupt change point for}~t\ge T_0~(\textrm{see Figure}~ \ref{fig1}).
\end{split}    
\end{eqnarray}

\section{Proposed Stationarity Test and Change Point Detection}\label{sec:test}
% In nonstationary environments, we define $Q_t^{\tiny{opt}}$ to be a version of $Q^{\tiny{opt}}$ with the transition and reward function equal to $\mathcal{P}_t$ and $r_t$, respectively. As commented earlier, when $Q_t^{\tiny{opt}}$ is stationary, the optimal policy is stationary as well. Existing Q-learning type methods are applicable for policy learning. 
In this section, we start by proposing three types of test statistics for testing \eqref{eqn:H0}. Next, we present key computation steps in constructing these tests. Finally, we present our change point detection method, built upon the proposed test. 

\begin{algorithm}[!t]
\caption{Testing Stationarity of the Optimal Policy via the Optimal Q-Function.}
\label{alg:full}
\begin{algorithmic}
\item[]
\begin{description}
    \item[\textbf{Input}:] The offline data $\{(S_{i,t},A_{i,t},R_{i,t}) \}_{1\le i\le N, T_0\le t\le T}$, and the significance level $0<\alpha<1$. 
    
    %			\item[\textbf{Step 1}.] Randomly divide $\{1,2,\ldots,n\}$ into two disjoint subsets $\mathcal{I}_1\cup \mathcal{I}_2$ of equal sizes. 
    
    \item[\textbf{Step 1}.] For each $u \in [T_0+\epsilon T, (1-\epsilon) T]$, employ the fitted Q-iteration algorithm to compute the estimated Q-functions $\widehat{Q}_{[T_0,u]}$ and $\widehat{Q}_{[u,T]}$ on separate time segments. 
    %Compute an initial estimator $\widetilde{\bm{W}}^{(\ell)}$ for $\bm{W}_0$, given $\{\bm{X}_i\}_{i\in \mathcal{I}_{\ell}}$, $\ell=1,2$.
    
    \item[\textbf{Step 2}.] Construct one of the CUSUM-type test statistics $\textrm{TS}_1$, $\textrm{TS}_{\infty}$ or $\textrm{TS}_{n}$, according to \eqref{eqn:teststat1},  \eqref{eqn:teststatinf} or \eqref{eqn:teststatninf}. 
    
    \item[\textbf{Step 3}.] Employ multiplier bootstrap to compute the bootstrapped test statistics $\textrm{TS}_1^b$, $\textrm{TS}_{\infty}^b$ or $\textrm{TS}_{n}^b$, and calculate the $p$-value according to \eqref{eqn:calpvalue}.
    % according to \eqref{eqn:pvalue}. 
    
    \item[\textbf{Output}:] Reject the null hypothesis if the $p$-value is smaller than $\alpha$.
\end{description}
\end{algorithmic}
\end{algorithm}

\subsection{Test Statistics}\label{sec:mainidea}
%We focus on the null hypothesis that $Q_t^{\tiny{opt}}=Q^{\tiny{opt}}$ for any $t\in \{T_0,T_0+1, \ldots,T\}$ and 
% Recall that $T$ denotes the termination time in the offline data. %Let $T_0$ be a positive integer such that 
% Assume $T_0<T$. 
% We propose an integral-type and two maximum-type test statistics. 
All test statistics we propose require an estimated optimal Q-function, denoted as $\widehat{Q}_{[T_1,T_2]}(a,s)$ (here we drop the superscript $opt$ in $Q^{opt
}$ for simplicity), using data collected in the interval $[T_1, T_2] \subseteq [T_0, T]$.
For any candidate change point $u \in (T_0 +\epsilon T, (1-\epsilon) T)$, where $\epsilon > 0$ is a pre-specified boundary-removal constant, we use $\widehat{Q}_{[u,T]}(a,s)-\widehat{Q}_{[T_0,u]}(a,s)$ to measure the difference in the optimal Q-function after and before the candidate. 
Based on this measure, we introduce three test statistics: an $\ell_1$-type, a maximum-type, and a normalized maximum-type, given by 
\begin{align}
\begin{split}
\textrm{TS}_1 = \max_{T_0+\epsilon T<u<(1-\epsilon) T} \tau_u \Big[\frac{1}{N(T-T_0)} \sum_{i,t}|\widehat{Q}_{[T_0,u]}(A_{i,t},S_{i,t})-\widehat{Q}_{[u,T]}(A_{i,t},S_{i,t}) |\Big],  \label{eqn:teststat1}
\end{split}\\
\begin{split}
\textrm{TS}_{\infty} = \max_{T_0+\epsilon T<u<(1-\epsilon) T} \max_{a,s} \tau_u |\widehat{Q}_{[T_0,u]}(a,s)-\widehat{Q}_{[u,T]}(a,s) |, \label{eqn:teststatinf}
\end{split} \\
\begin{split}
\textrm{TS}_{n} = \max_{T_0+\epsilon T<u<(1-\epsilon) T} \max_{a,s} \tau_u \widehat{\sigma}^{-1}_u(a,s) |\widehat{Q}_{[T_0,u]}(a,s)-\widehat{Q}_{[u,T]}(a,s) |, \label{eqn:teststatninf}
\end{split}
\end{align}
%\ml{Should equation \eqref{eqn:teststat1} have factor $\frac{1}{N(T-T_0)}$ instead of $\frac{1}{NT}$? }
respectively, where $\tau_u = \sqrt{\frac{(u-T_0)(T-u)}{(T-T_0)^2}}$ is the scale factor being dependent on the proximity of $u$ to the interval boundary, $\widehat{\sigma}^2_u(a,s)$ denotes a consistent variance  estimator of $\widehat{Q}_{[T_0,u]}(a,s)-\widehat{Q}_{[u,T]}(a,s)$ whose detailed form is given in Appendix \ref{sec:proofmax}. 

Under a given significance level $\alpha$, the critical values and $p$-values for $\textrm{TS}_1$, $\textrm{TS}_{\infty}$ and $\textrm{TS}_{n}$ are computed using a multiplier bootstrap method (detailed in Section \ref{sec:boot}) to approximate their asymptotic distributions. 
This multiplier bootstrap is easy to implement. Unlike other methods such as the nonparametric bootstrap, it does not require to re-estimate the Q-function, thus simplifying the estimation.  
We reject the null hypothesis when the test statistic exceeds its corresponding critical value (or equivalently, the $p$-value falls below $\alpha$). 
%In practice, we only need to choose one of the test statistics as all three types produce asymptotically consistent tests.

%We make a few remarks. First, 
\begin{remark}
The three test statistics \eqref{eqn:teststat1}--\eqref{eqn:teststatninf} are very similar to the classical cumulative sum (CUSUM) statistic in change point analysis \citep{csorgo1997limit}. The weight scale $\tau_u$ assigns smaller weights on data points near the boundary of the interval $(T_0 + \epsilon T, (1-\epsilon) T)$. %. The boundary removal constant $\epsilon$ 
Removing the boundary is necessary as it is difficult to accurately estimate the Q-function when close to the boundary. Such practice is commonly employed in the time series literature for change point detection in non-Gaussian settings \citep[see e.g.,][]{cho2012multiscale,yu2021finite}. 
\end{remark}
\begin{remark}\label{remark:compare3tests}
In addition, the three test statistics differ in how they aggregate the estimated changes $|\widehat{Q}_{[T_0,u]} - \widehat{Q}_{[u,T]}|$ across different state-action pairs. The $\ell_1$-type test \eqref{eqn:teststat1} computes the average of the changes, weighted by the empirical state-action distribution. The two maximum-type tests \eqref{eqn:teststatinf} and \eqref{eqn:teststatninf} focus on the largest change in the (normalized) absolute value. Normalization can enhance efficiency, especially in cases where some state-action pairs $(a,s)$ are less frequently visited. In these cases, the standard error of the difference $\widehat{Q}_{[T_0,u]}(a,s) - \widehat{Q}_{[u,T]}(a,s)$ tends to be large. As a consequence, the estimated maximizer $\argmax_{a,s} |\widehat{Q}_{[T_0,u]}(a,s)-\widehat{Q}_{[u,T]}(a,s)|$ may differ significantly from the oracle maximizer \\$\argmax_{a,s} |Q^{\tiny{opt}}_{[T_0,u]}(a,s)-Q^{\tiny{opt}}_{[u,T]}(a,s)|$, leading to reduced power in the unnormalized test. We remark that such normalized supremum-type statistics have been commonly employed in the econometrics literature \citep[see e.g.,][] {belloni2015some,chen2015optimal,chen2018optimal}.
\change{On the other hand, the normalized test requires to compute the estimated variance $\widehat{\sigma}_u^2$, which introduces additional variability into the test statistic that might lower its type-I error control, and slightly increases computational time. Thus, we expect the normalized test to exhibit a larger type-I error and a larger power than the unnormalized and $\ell_1$-type tests, despite all the three tests are consistent. This is confirmed by our theoretical and numerical studies, which we will demonstrate later.
%This is confirmed in our real-data-based simulation, where the normalized test achieves slightly larger empirical type-I error and power than the other two tests; see Figure \ref{fig:real:result} of the Supplementary Materials.We also demonstrate in Table \ref{tab:sim:1d:computation_time} of the Supplementary Materials the unnormalized test is the most computationally efficient, while the integral test has similar efficiency to the unnormalized test when $\kappa$ is small and is similar to the normalized test when $\kappa$ is large.
}
\end{remark}

\begin{remark}
    To conclude this section, we remark that the proposed test is model-free, as it constructs the test statistic without directly estimating the reward and transition functions. Alternatively, one may consider model-based tests. We compare against an existing model-based test in Section \ref{sec:sim:1d} and provide a detailed discussion of model-free versus model-based tests in Section \ref{sec:modelbased} of the Supplementary Materials. 
\end{remark}

\subsection{Estimation of the Q-Function}\label{sec:estQ}
We provide more details on estimating $Q^{\tiny{opt}}$ in this section. In particular, we focus on a discrete action space $\mathcal{A}=\{0,1,\cdots,m-1\}$ with $m$ available actions. We employ the sieve method \citep{grenander1981abstract} to model $Q^{\tiny{opt}}$, primarily for two reasons. First, the sieve method ensures the resulting Q-estimator has a tractable limiting distribution, which allows us to derive the asymptotic distribution of the test statistic. Second, the sieve method is useful in mitigating bias caused by model misspecification, achieved by increasing the number of basis functions.

Specifically, we propose to model $Q^{\tiny{opt}}(a,s)$ by $\phi_L^\top(a,s) \beta^*$ for some $\beta^*\in \mathbb{R}^{mL}$ where
\begin{eqnarray} \label{eqn:basis}
	\phi_L(a,s) = [\mathbb{I}(a=0) \Phi^\top(s), \mathbb{I}(a=1) \Phi^\top(s), \cdots, \mathbb{I}(a=m-1) \Phi^\top(s) ]^\top,
\end{eqnarray}
is an $mL$-dimensional vector constructed using products between the action indicator $\mathbb{I}(a=\bullet)$ and a vector of $L$ basis functions $\Phi$ on the state space. Several choices can be considered here for $\Phi$. For continuous state spaces, options for $\Phi$ include power series, Fourier series, splines or wavelets \citep[see e.g.,][]{judd1998numerical}. 
For discrete state spaces, one could use a lookup table and set
$\phi_L(a,s) = [\mathbb{I}\{(a,s)=(a', s')\}; a' \in \cA, s' \in \cS  ]^\top$.
In Section \ref{sec:testcon}, we show that the proposed test is not overly sensitive to the choice of the number of basis functions $L$. %An appropriate choice of $L$ balances off the bias and standard deviation of the Q-function estimator.
% , yet a smaller approximation error to the oracle $Q^{\tiny{opt}}(a,s)$ leads to a higher power. 
In practice, we can determine $L$ using cross-validation, as illustrated in our simulation studies. 

\change{For a given time interval $[T_1, T_2] \subseteq [T_0, T]$, we compute an estimator $\widehat{\beta}_{[T_1,T_2]}$ for $\beta^*$ using data collected from this interval. 
Specifically, we employ FQI to iteratively update $\widehat{\beta}_{[T_1,T_2]}$ as outlined in \eqref{eqn:updateQinFQI}. With a linear model, we perform ordinary least square regression at the $k$th iteration with $\{R_{i,t}+ \gamma \max_a \phi^\top(a, S_{i,t+1})\beta^{(k-1)}\}_{1\le i\le N,T_1\le t<T_2}$ as the responses and $\{\phi(A_{i,t},S_{i,t})\}_{1\le i\le N,T_1\le t<T_2}$ as the predictors to compute the regression coefficients $\beta^{(k)}$. The procedure stops after $K$ iterations and we set $\widehat{\beta}_{[T_1,T_2]}$ to $\beta^{(K)}$.}

\subsection{Bootstrap Approach to Critical Value}\label{sec:boot}
We employ a multiplier bootstrap method \citep{wu1986jackknife,cherno2014} to obtain the $p$-values. The idea is to simulate Gaussian random noises to approximate the limiting distribution of the Q-function estimator, and subsequently to approximate that of the test statistic.
% As we will show in Appendix \ref{sec:theoryQ}, each estimator $\widehat{\beta}_{[T_1,T_2]}$ computed by solving \eqref{eqn:betahat} is asymptotically normal. As such, the estimated Q-function is asymptotically normal as well. This motivates us to do employ the multiplier bootstrap to approximate the asymptotic distribution of the Q-estimator and the resulting test statistics. 
% The idea is to generate bootstrap samples to approximate the limiting distribution of each of $\textrm{TS}_1$, $\textrm{TS}_{\infty}$ and $\textrm{TS}_n$. 
A key observation is that, under the null hypothesis, when the Q-function is well-approximated and the optimal policy is unique, the estimated Q-function $\phi_L(a,s)^\top \widehat{\beta}_{[T_1,T_2]}$ has the following linear representation:
\begin{eqnarray}\label{eqn:linearrep}
\begin{split}
%	\begin{aligned}
    &\phi_L(a,s)^\top \widehat{\beta}_{[T_1,T_2]}-Q^{\tiny{opt}}(a,s)\\
    =&\frac{1}{N(T_2-T_1)}\phi_L^\top(a,s)W_{[T_1,T_2]}^{-1} \sum_{i=1}^N \sum_{t=T_1}^{T_2-1} \phi_{L}(A_{i,t},S_{i,t})\delta_{i,t}^*+o_p(1),
%	\end{aligned}
\end{split}
\end{eqnarray}
where 
\begin{eqnarray*}
	W_{[T_1,T_2]}=\frac{1}{T_2-T_1}\sum_{t=T_1}^{T_2-1} \Mean \phi_{L}(A_{i,t},S_{i,t})\{\phi_{L}(A_{i,t},S_{i,t})-\gamma \phi_{L}(\pi^{\tiny{opt}}(S_{i,t+1}),S_{i,t+1})\}^\top,
 \end{eqnarray*}
$\delta_{i,t}^*=R_{i,t}+\gamma \max_a Q^{\tiny{opt}}(a,S_{i,t+1}) - Q^{\tiny{opt}}(A_{i,t}, S_{i,t})$ is the temporal difference error. 
By the Bellman optimality equation \eqref{eqn:bell}, the leading term on the right-hand-side (RHS) of \eqref{eqn:linearrep} forms a mean-zero martingale. When its quadratic variation process converges, it follows from the martingale central limit theorem \citep{McLeish1974} that $\widehat{\beta}_{[T_1,T_2]}$ is asymptotically normal. As such, the estimated Q-function is asymptotically normal as well. Refer to Lemma \ref{lemmaQ} in the Supplementary Materials for details. 
\begin{remark}
To the best of our knowledge, the limiting distribution of the Q-function estimated via FQI has not been established in the existing RL literature. Most papers focus on establishing non-asymptotic error bound of the estimated Q-function \citep[see e.g.,][]{munos2008finite,chen2019information,fan2020theoretical, uehara2021finite}. One exception is a recent proposal by \citet{hao2021bootstrapping} that studied the asymptotics of Q-estimators computed via the fitted Q-evaluation \citep[FQE,][]{le2019batch} algorithm. We note that FQE is similar to FQI but is designed for the purpose of policy evaluation. 
\end{remark}
In addition, it follows from \eqref{eqn:linearrep} that
\begin{eqnarray}\label{eqn:1.5}
	\begin{aligned}
		\widehat{Q}_{[T_0,u]}(a,s)-\widehat{Q}_{[u,T]}(a,s)=	\frac{1}{N(u-T_0)}\phi_{L}^\top(a,s)W_{[T_0,u]}^{-1} \sum_{i=1}^N \sum_{t=T_0}^{u-1} \phi_{L}(A_{i,t},S_{i,t})\delta_{i,t}^*\\-\frac{1}{N(T-u)}\phi_{L}^\top(a,s)W_{[u,T]}^{-1} \sum_{i=1}^N \sum_{t=u}^{T-1} \phi_{L}(A_{i,t},S_{i,t})\delta_{i,t}^*+o_p(1).
	\end{aligned}
\end{eqnarray}
This motivates us to %employ a Gaussian multiplier bootstrap to approximate the asymptotic distribution of the Q-estimator and the resulting test statistics.
%Multiplier bootstrap has shown success in approximating the distribution of sum of asymptotically normal random variables \citep{Cherno2018}.We can 
construct $B$ bootstrap samples to approximate the asymptotic distribution of the leading term on the RHS of \eqref{eqn:1.5}. Specifically, at the $b$th iteration, $b = 1, \ldots, B$, we compute a bootstrap sample $\widehat{Q}^b_{[T_0,u]}(a,s)-\widehat{Q}^b_{[u,T]}(a,s)$ where
\begin{eqnarray*}
    \widehat{Q}^b_{[T_1,T_2]}(a,s)=\frac{1}{N(T_2-T_1)}\phi_{L}^\top(a,s)\widehat{W}_{[T_1,T_2]}^{-1} \sum_{i=1}^N \sum_{t=T_1}^{T_2-1} \phi_{L}(A_{i,t},S_{i,t})\delta_{i,t}(\widehat{\beta}_{[T_1,T_2]}) e_{i,t}^b,\,\forall T_1,T_2,
\end{eqnarray*}
where $\widehat{W}_{[T_1,T_2]}$ denotes a consistent estimator for $W_{[T_1,T_2]}$ (refer to Lemma \ref{lemmamatrixnonstat} for a detailed upper bound on the estimation error) given by
\begin{eqnarray*}
\begin{split}
    \widehat{W}_{[T_1,T_2]}=\frac{1}{N(T_2-T_1)}\sum_{i=1}^N \sum_{t=T_1}^{T_2-1}   \phi_{L}(A_{i,t},S_{i,t}) 
     \{\phi_{L}(A_{i,t},S_{i,t})-\gamma \phi_{L}(\pi_{\widehat{\beta}_{[T_1,T_2]}}(S_{i,t+1}),S_{i,t+1})\}^\top,
\end{split}
\end{eqnarray*}
$\delta_{i,t}(\beta) = R_{i,t}+\gamma \max_a \beta^\top \phi_L(a,S_{i,t+1})-\beta^\top \phi_L(A_{i,t},S_{i,t})$, %is the temporal difference error, 
% \begin{eqnarray*}%\label{eqn:tde}
%     \delta_{i,t}(\beta)=R_{i,t}+\gamma \max_a \beta^\top \phi_L(a,S_{i,t+1})-\beta^\top \phi_L(A_{i,t},S_{i,t}),
% \end{eqnarray*}
%$\pi_{\beta}$ is defined in \eqref{eqn:gradient}, and 
$\{e_{i,t}^b\}_{i,t}$ is a sequence of i.i.d. standard Gaussian random variables independent of the observed data, and $\pi_{\widehat{\beta}_{[T_1,T_2]}}$ denotes the greedy policy with respect to the estimated Q-function (see \eqref{eqn:optimalpolicy}), $\pi_{\widehat{\beta}_{[T_1,T_2]}}(s)=\argmax_{a} \phi_L^\top(a,s) \widehat{\beta}_{[T_1,T_2]}$. This yields the bootstrapped statistics,
\begin{align*}
\begin{split}
\textrm{TS}_1^b = \max_{T_0+\epsilon T<u<(1-\epsilon) T} \tau_u \Big[\frac{1}{N(T-T_0)} \sum_{i,t}|\widehat{Q}_{[T_0,u]}^b(A_{i,t},S_{i,t})-\widehat{Q}^b_{[u,T]}(A_{i,t},S_{i,t}) |\Big],
\end{split}\\
\begin{split}
\textrm{TS}_{\infty}^b = \max_{T_0+\epsilon T<u<(1-\epsilon) T} \max_{a,s} \tau_u  |\widehat{Q}_{[T_0,u]}^b(a,s)-\widehat{Q}_{[u,T]}^b(a,s) |,
\end{split} \\
\begin{split}
\textrm{TS}_{n}^{b} = \max_{T_0+\epsilon T<u<(1-\epsilon) T} \max_{a,s} \tau_u  \widehat{\sigma}^{-1}_u(a,s) |\widehat{Q}_{[T_0,u]}^b(a,s)-\widehat{Q}_{[u,T]}^b(a,s) |.
\end{split}
\end{align*}
The random noise $e_{i,t}^b$ in $\widehat{Q}^b_{[T_1,T_2]}$ plays a crucial role in the approximation of the asymptotic distribution. In particular, in Step 3 of the proof of Theorem \ref{thm:size} (see Section \ref{sec:proofunmax} of the Supplementary Materials), we show that the conditional variance of the difference in the bootstrap sample $\widehat{Q}_{[T_0,u]}^b(a,s)-\widehat{Q}_{[u,T]}^b(a,s)$  given the data, is asymptotically equivalent to the asymptotic variance of the difference in the actual Q-function estimator $\widehat{Q}_{[T_0,u]}(a,s)-\widehat{Q}_{[u,T]}(a,s)$. Meanwhile, $\widehat{Q}_{[T_0,u]}^b(a,s)-\widehat{Q}_{[u,T]}^b(a,s)$ follows a normal distribution given the data, due to the injected Gaussian noises $\{e_{i,t}^b\}_{i,t}$, while $\widehat{Q}_{[T_0,u]}(a,s)-\widehat{Q}_{[u,T]}(a,s)$ is asymptotically normal. These alignments justify the use of the multiplier bootstrap. We thus set the critical value of each test to the $\alpha$th upper quantile of the bootstrapped samples. The $p$-values can be computed by
\begin{eqnarray}\label{eqn:calpvalue}
    \frac{1}{B}\sum_{b=1}^B \mathbb{I}(\textrm{TS}_1^b>\textrm{TS}_1), 
    \frac{1}{B}\sum_{b=1}^B \mathbb{I}(\textrm{TS}_{\infty}^b>\textrm{TS}_{\infty})\,\,\hbox{and}\,\,
    \frac{1}{B}\sum_{b=1}^B  \mathbb{I}(\textrm{TS}_{n}^b>\textrm{TS}_{n}),
\end{eqnarray}
respectively.

% \begin{eqnarray*}%\label{eqn:ts1b}
% 	\textrm{TS}_1^b=\max_{T_0+\epsilon T<u<(1-\epsilon) T} \sqrt{\frac{(u-T_0)(T-u)}{(T-T_0)^2}}\left\{\frac{1}{NT}\sum_{i,t}|\widehat{Q}_{[T_0,u]}^b(A_{i,t},S_{i,t})-\widehat{Q}^b_{[u,T]}(A_{i,t},S_{i,t}) |\right\},\\%\label{eqn:tsinftyb}
% 	\textrm{TS}_{\infty}^b=\max_{T_0+\epsilon T<u<(1-\epsilon) T} \max_{a,s} \sqrt{\frac{(u-T_0)(T-u)}{(T-T_0)^2}} |\widehat{Q}_{[T_0,u]}^b(a,s)-\widehat{Q}_{[u,T]}^b(a,s) |,\\%\label{eqn:tsinftybstar}
% 	\textrm{TS}_{n,\infty}^{b}=\max_{T_0+\epsilon T<u<(1-\epsilon) T} \max_{a,s} \sqrt{\frac{(u-T_0)(T-u)}{(T-T_0)^2}} \widehat{\sigma}^{-1}_u(a,s) |\widehat{Q}_{[T_0,u]}^b(a,s)-\widehat{Q}_{[u,T]}^b(a,s) |.
% \end{eqnarray*}
% In Section \ref{sec:testcon}, we show that under the null hypothesis, the asymptotic distributions of $\textrm{TS}_1$, $\textrm{TS}_{\infty}$ and $\textrm{TS}_{n}$ can be well-approximated by the conditional distributions of $\textrm{TS}_1^b$, $\textrm{TS}_{\infty}^b$ and $\textrm{TS}_{n}^{b}$ given the observed data. The corresponding $p$-values are given by
% \begin{eqnarray}\label{eqn:pvalue}
% 	\prob(\textrm{TS}_1^b > \textrm{TS}_1|\textrm{Data}),\,\,\,\,\prob(\textrm{TS}_{\infty}^b > \textrm{TS}_{\infty}|\textrm{Data})\,\,\hbox{and}\,\,\prob(\textrm{TS}_{n}^{b} > \textrm{TS}_{n}|\textrm{Data})
% \end{eqnarray}
% respectively. We reject the null when the $p$-value is smaller than a given significance level $\alpha$. 

\subsection{Change Point Detection}\label{sec:changedetection}

Given the offline data collected from $N$ subjects up to time $T$, we aim to learn an optimal ``warm-up''  policy, i.e., the optimal policy that maximizes these subjects' long-term rewards starting from time $T$, until the reward or transition function changes. %-- that can be used for treatment recommendation for these subjects. 
This goal aligns with our motivating mHealth study setting where the researchers want to design the best policy based on pre-collected data %(which can be expensive to collect) 
and extends the policy to the same group of subjects after the study ends. To this end, we focus on identifying the most recent change point $T^*$ such that $Q^{\tiny{opt}}_{T^*}=Q^{\tiny{opt}}_{T^*+1}=\cdots=Q^{\tiny{opt}}_T$ and apply state-of-the-art Q-learning to the data subset $\{(S_{i,t},A_{i,t},R_{i,t})\}_{1\le i\le N, T^*\le t\le T}$ to learn $\pi^{opt}_T$. 

To estimate $T^*$, we apply any of the three proposed tests to a sequence of candidate change points from the back. We start by specifying a monotonically increasing sequence $\{\kappa_j\}_j \subseteq (0, T)$ and apply the test to intervals $[T - \kappa_j, T]$. The estimator $\widehat{T}^*$ is then set to the candidate change point before the first rejection, i.e., $\widehat{T}^* = T - \kappa_{j_0-1}$ where the test is first rejected at $\kappa_{j_0}$. If no changes are detected, we
propose to use all the observed data for policy optimization. 
%\change{
%We recommend specifying as many $\kappa_j$'s as possible at which the test is applied, to ensure the most recent change point is precisely identified.
%}

\change{Though offline, the proposed change point detection method is an integral part in batch online RL settings via the following strategy. Step 1) A behavior policy is used to collect experiences; 2) After certain amount of experiences is collected, apply the proposed approach to detect the most recent change point; 3) If there exists a change point, use the data after the most recent change point to update the policy; 4) The new policy, together with $\epsilon$-greedy algorithm, is then used to collect more data; Repeat Steps 2) to 4) for a pre-specified or indefinitely long duration of interactions with the environment. See Section \ref{subsect:onlineevaluation} %and Supplementary Section \ref{sec:sim:evaluation:implementation} 
for detailed simulation %real-data based 
experiments that demonstrate this strategy and utility.}

\section{Consistency of the Test}\label{sec:theory}
We proceed by first investigating the size (i.e., the rejection probability or type-I error) of the proposed test, and next establishing its power property. To simplify the theoretical analysis, we focus on the setting where the state space $\mathcal{S}=[0,1]^d$, and $\Phi$ denotes the tensor product of B-spline basis functions, motivated by their popularity in the sieve estimation literature \citep[see e.g., Section 6 of][for a review]{chen2015optimal}. We use $p_t(\bullet|a,s)$ to denote the probability density function of $\mathcal{T}_t(a,s,\delta_1)$. %which is bounded on $\mathcal{S}$ and has a bounded derivative $\partial p_t(s'|a,s)/\partial s$. 
In other words, $p_t$ corresponds to the density function of $S_{t+1}$ given $(A_t,S_t)=(a,s)$. 
For each $s$ and $t$, we use $\pi^{opt}_t(s)$ to denote the greedy action $\argmax_a Q_t^{opt}(a,s)$ that $\pi^{opt}_t$ picks (see \eqref{eqn:optimalpolicy}). Let $\delta_t^*$ denote the temporal difference error $R_t+\max_a Q_t^{\tiny{opt}}(a,S_{t+1})-Q_t^{\tiny{opt}}(A_t,S_t)$. 
As commented in the introduction, all the theories in this section are established under a bidirectional asymptotic framework, which is to say that they are valid as either $N$ or $T$ diverges to infinity.

\subsection{Size of the Test}\label{sec:techcond}
We introduce the following assumptions: 
\begin{enumerate}
    \item[\textbf{A1}] (\textbf{CUSUM statistics}): The boundary removal parameter $\epsilon$ in the proposed test statistics 
    \eqref{eqn:teststat1}, \eqref{eqn:teststatinf} and \eqref{eqn:teststatninf},
    is proportional to $\log^{-c_1} (NT)$ for some $c_1\ge 0$.
    \change{\item[\textbf{A2}] (\textbf{Realizability}): Assume that there exist some constants $p,c_2>0$, such that for any $a\in \mathcal{A}$ and $t\ge T_0$, the reward function $r_t(a, \bullet) \in \Lambda(p, c_2)$, where $\Lambda(p, \bullet)$ is the H{\"o}lder class with the smoothness parameter $p$; see the Supplementary Materials for its definition.
    \item[\textbf{A3}] (\textbf{Completeness}): For any $\beta \in \mathbb{R}^{mL}$ whose $\ell_2$ norm $\|\beta\|_2$ is less than or equal to $1$, there exists some $\beta^*$ with $\|\beta^*\|_2\le 1$ such that the function $\mathcal{B}_t \phi_L^\top \beta$ can be uniformly approximated by $\phi_L^\top \beta^*$ with an approximation error $O(L^{-p/d})$, 
    %function $g$ such that $g(\bullet,a)\in \Lambda(p,c)$ for any action $a$ and some constant $c>0$, we have $\mathcal{B}_t g(\bullet,a)\in \Lambda(p,c)$ for any $t\ge T_0$ 
    where $\mathcal{B}_t$ denotes the operator $(\mathcal{B}_t g) (a,s)=\Mean [\max_{a'} g(a',\mathcal{T}_t(a,s,\delta))]$.} 
    \item[\textbf{A4}] (\textbf{Transition}): (i) $\sup_{t,a} \Mean \|\mathcal{T}_t(a,s,\delta)-\mathcal{T}_t(a,s',\delta)\|_2\le \rho \|s-s'\|_2$ for some $0\le \rho<1$, $\sup_{t,a,s} \|\mathcal{T}_t(a,s,\delta)-\mathcal{T}_t(a,s,\delta')\|_2=O(\|\delta-\delta'\|_2)$%, where $\|\cdot\|_2$ denotes the $\ell_2$ norm of a vector
    ; (ii) Suppose $\delta_1$ has sub-exponential tails, i.e., for any $j$th element $\delta_{1,j}$, $\Mean |\delta_{1,j}|^k=O (k! c_3^{k-2})$ for some constant $c_3>0$; (iii) $p_t(s'|a,s)$ is bounded and is Lipschitz continuous as a function of $s$; (iv) $\inf_{t,a,s}\Var((1-\gamma)\delta_t^*|A_t=a,S_t=s)$ is bounded away from zero. 
    
    \item[\textbf{A5}] (\textbf{Behavior policy}): (i) $\pi^b$ is a Markov policy; (ii) $\inf_{t\ge T_0,0\le \gamma'\le \gamma}(1-\gamma')^{-1}\lambda_{\min} [\Mean \phi_L(A_t,S_t) \\\phi_L(A_t,S_t)^\top-(\gamma')^2 \Mean \phi_{L}(\pi_{\tiny{opt}}(S_{t+1}),S_{t+1})\phi_{L}^\top(\pi_{\tiny{opt}}(S_{t+1}),S_{t+1})]$ 
    is uniformly bounded away from zero where $\lambda_{\min}[\bullet]$ denotes the minimum eigenvalue of a given matrix. %and the infimum over $\pi_{\beta}$ is taken over the class of linear policies $\pi_{\beta}(s)=\argmax_{a} |\phi_L^{\top}(a,s)\beta|$. 
    
    \item[\textbf{A6}] (\textbf{Optimal policy}): The optimal policy $\pi^{\tiny{opt}}_t$ is unique for all $t$. %(ii) 
    \item[\textbf{A7}] (\textbf{Optimal Q-function}): The margin $Q_t^{\tiny{opt}}(\pi^{\tiny{opt}}_t(s), s)-\max_{a\in \mathcal{A}\setminus \pi^{\tiny{opt}}_t(s)}Q_t^{\tiny{opt}}(a,s)$ is bounded away from zero, uniformly for all $s$ and $t$. 
    \item[\textbf{A8}] (\textbf{Computation}): The number of FQI iterations $K$ to produce the estimated Q-function $\widehat{Q}$ is much larger than $\log(NT)$. 
    \item[\textbf{A9}] (\textbf{Basis functions}): $L$ is proportional to $(NT)^{c_4}$ for some $0<c_4<1/4$. Under the null SA3 (see \eqref{eqn:H0}), we additionally require $c_4>d/(2p)$ for all three types of tests. %However, such a lower bound requirement is not needed under SA1 and SA5.
   
    %Under the null SA3 (see \eqref{eqn:H0}), we additionally require $c_4>d/(2p)$ for all three types of tests. For the unnormalized test \eqref{eqn:teststatinf}, it is further required that $c_4>d/(2p-d)$. However, these lower bound requirements are not needed under SA1 and SA5. 
\end{enumerate}

\begin{remark}\label{remark:A1}
    As commented earlier, assumptions similar to A1 are commonly adopted in the change point detection literature \citep{yu2021finite}. \change{This assumption can be easily satisfied as the boundary removal parameter $\epsilon$ is user-specified}. 
\end{remark}
\begin{remark}
    \change{The realizability and completeness assumptions are commonly imposed in the RL literature \citep[see e.g.,][]{chen2019information,uehara2021finite}. In our context, realizability requires the H{\"o}lder class to be sufficiently rich to contain the reward functions. The completeness assumption requires the linear function class to be ``approximately complete” in the sense that it remains closed under the operator $\mathcal{B}_t$ up to certain approximation error. It holds automatically when the transition density function $p_t$ belongs to the H{\"o}lder class as well \citep[see e.g.,][]{fan2020theoretical,shi2022statistical}. Such smoothness conditions are commonly imposed in the sieve estimation literature as well \citep[see e.g.][]{Huang1998,chen2015optimal}. They are mild as the smoothness parameter $p$ remains unspecified and can be adjusted to be arbitrarily small to meet the two conditions.}
\end{remark}
\begin{remark}
    Conditions A4(i) and (ii) are needed to establish concentration inequalities for nonstationary Markov chains \citep{alquier2019exponential}. They allow us to develop a matrix concentration inequality with nonstationary transition functions, which is needed to prove the validity of the bootstrap method (see  Lemma \ref{lemmamatrixnonstat} in the Supplementary Materials for details). %Suppose the behavior policy is stationary over time and is Lipschitz continuous as a function of the state. 
These assumptions are automatically satisfied when the state satisfies a time-varying AR(1) process: 
\begin{eqnarray*}
	S_{t+1}=\rho_t S_t+\beta_t A_t+\delta_t,
\end{eqnarray*}
for some $\{\rho_t\}_t$ and $\{\beta_t\}$ such that $\sup_t |\rho_t|<1$, and $\delta_t$ has sub-exponential tails. More generally, it also holds when the auto-regressive model is given by
\begin{eqnarray*}
	S_{t+1}=f_t(A_t,S_t)+\delta_t,
\end{eqnarray*}
with $\sup_{a,t} |f_t(a,s)-f_t(a,s')|\le \rho \|s-s'\|_2$ for some $\rho<1$. When the transition functions are stationary over time, it essentially requires the Markov chain to possess the exponential forgetting property \citep{dedecker2015deviation}. Condition A4(iii) is automatically satisfied when $p_t$ belongs to the H{\"o}lder class with the smoothness parameter $p\ge 1$. Condition A4(iv) requires the transition to be stochastic so that the variance of the temporal difference error remains strictly positive. 

\change{We also remark that the sub-exponential tail assumption is generally met in mHealth studies, particularly in the IHS. %This is because the state variables, which include sleeping minutes and mood scores, are bounded. 
This is because the state variable such as the mood score is  bounded. Additionally, after cubic root transformation of weekly average step count and square root transformation of weekly average sleep minutes, %the upper tails of 
these two variables exhibit light upper tails, similar to those of a normal distribution, as can be seen from Figure \ref{fig:data:qqplot} of the Supplementary Materials.}
\end{remark}

%Fourth, %The last part of (A4) holds since each function in $\Phi_L$ is Lipschitz continuous.  
\begin{remark}
    A5(i) allows the behavior policy that generates the data to be nonstationary over time. %It is automatically satisfied in randomized studies where the behavior policy is usually a constant function of the state. 
    Assumptions to A5(ii) are commonly imposed in the statistics literature on RL \citep[see e.g.,][]{Ertefaie2018,luckett2019,shi2022statistical}. These assumptions are typically satisfied in mobile health where the behavior policy is usually a constant policy -- as in the IHS study -- which is thus Markovian (fulfilling A5(i)) and is expected to cover the optimal policy (fulfilling A5(ii)).
\end{remark}
%Fifth, 
\begin{remark}
A6 is a necessary condition for establishing the limiting distribution of $\widehat{\beta}_{[T_1,T_2]}$ computed based on FQI. It is widely imposed in the statistics literature  \citep{Ertefaie2018,luckett2019}, but can be violated in nonregular settings where the optimal policy is not unique  \citep{chakraborty2013inference,Alex2016,shi2020breaking,guo2021inference}. \change{Our proposal could be further coupled with data splitting to derive a valid test in nonregular settings without A6. Specifically, we divide all trajectories into two parts: one to estimate the optimal policy and the other to construct the test.} However, the resulting test might suffer from a loss of power, due to the use of data splitting. 
\end{remark}
%We discuss this in Section \ref{sec:nonregular} in detail. 
%Seventh,
\begin{remark}
The margin $Q_t^{\tiny{opt}}(\pi^{\tiny{opt}}_t(s), s) -\max_{a\in \mathcal{A} \setminus \{\pi^{\tiny{opt}}_t(s)\}} Q_t^{\tiny{opt}}(a,s)$ in A7 measures the difference between the state-action value under the best action and the second best action. This condition is imposed to simplify the theoretical analysis. \change{It can be relaxed to require the probability that the margin approaches zero to converge to zero at certain rate \citep[see e.g.,][]{qian2011,Alex2016,shi2022statistical}, when coupled with data splitting}.
\end{remark}

\begin{remark}
    %As shown in A9, %A7, the $\ell_1$-type test \eqref{eqn:teststat1} and normalized maximum-type test \eqref{eqn:teststatninf} require weaker conditions than the unnormalized test \eqref{eqn:teststatinf}. Specifically, for the former two tests, the condition $d/(2p)<c_4<1/2$ is sufficient. In contrast, the unnormalized test requires a more stringent condition, namely $d/(2p-d)<c_4<1/3$. Further, 
    \change{Both A8 and A9 are mild as $K$ and $L$ are user-specified}. Meanwhile, it is also possible to remove the lower bound requirements on $c_4$ in A9 under SA1 and SA5. In such cases, it suffices for the sieve approximation error to simply tend towards zero, as opposed to being $o\{(NT)^{-1/2}\}$. 
    The latter is required to ensure the bias of the Q-estimator converge to zero at a faster rate than its standard deviation. %As such, the proposed test is not overly sensitive to the number of basis functions $L$. 
    Consequently, our CUSUM-type test statistics ensure that the proposed tests remain valid under weaker assumptions about the approximation error and are not overly sensitive to the number of basis functions $L$. This is because at each hypothesized error location, the test statistics implement scaled differencing, requiring the difference in approximation errors, instead of  these errors themselves, to converge at a specific order. %Such an order requirement is automatically satisfied 
    Under SA1 and SA5, all the approximation errors are equal, and such an order requirement is automatically satisfied. 

    % This latter requirement, typically necessitating undersmoothing to accelerate the convergence of the bias of the Q-estimator to zero relative to its standard deviation, is thus not obligatory. Consequently, undersmoothing is not a prerequisite, and the proposed test is not overly sensitive to the number of basis functions. The rationale behind the proposed test's validity under weaker assumptions about the approximation error can be attributed to the employment of CUSUM-type statistics. These statistics implement scaled differencing at each hypothesized error location, thus requiring the difference in approximation errors (as opposed to the absolute values of these errors) to conform to a specific order. Such an order requirement is automatically satisfied under SA1 and SA5.
    %the constant $c_1$ in (A1) to be larger than $1/2$ whereas the maximum-type test requires $(NT)^{2c_1-1}\gg L$.
\end{remark}
%We focus on the following pair of hypothesis:
%\begin{eqnarray}\label{eqn:null}
%\begin{split}
%	\mathcal{H}_0: Q_t^{\tiny{opt}}=Q^{\tiny{opt}},\,\,\,\,\forall t\in \{T_0,T_0+1,\cdots,T\};\\
%	\mathcal{H}_1: \textrm{There~exists~some~change~point}~T^*~\textrm{such~that}
%\end{split}	
%\end{eqnarray}

% {\color{blue}[Piotr, would such a condition holds for the Wavelet basis? Could you please make some comments here?]}

%\subsection{Consistency of the Test}\label{sec:testcon}
%We derive the size of the proposed test in this section.
\begin{theorem}[Size]\label{thm:size}
    \change{Recall that $\mathcal{D}$ denote the observed data. Suppose A1-A9 and the null hypothesis SA3 hold,  %(A1)-(A5), (A7)-(A10) hold and $L$ is proportional to $(NT)^{c_5}$ for some $0<c_5<1/4$. Suppose under the null hypothesis, $\max_{a,s,u} |\phi_L^\top(a,s) (b_{[T_0,u]}-b_{[u,T]})|=O\{(NT)^{-c_6}\}$ and $\max_{a,s,u}  \|\phi_L^\top(a,s)(\beta_{[T_0,u]}^*-\beta_{[u,T]}^*)\|_2=O\{(NT)^{-c_6}\}$ for some $c_6>1/2$ where $b$ is defined in (A1) and $\beta^*$ is defined in \eqref{eqn:betastar}. Then under the null, 
    we have
	\begin{eqnarray*}
		&&\sup_{z} |\prob(\textrm{TS}_1^{b}\le z|\mathcal{D})-\prob(\textrm{TS}_1\le z)|=O\Big(\frac{\sqrt{L}\log^{2}(NT)}{(1-\gamma)^{3/4}(\epsilon NT)^{1/8}}\Big)+O\Big(\frac{L^{-p/d}\sqrt{NT\log(NT)} }{(1-\gamma)^2}\Big),\\
		&&\sup_{z} |\prob(\textrm{TS}_{n}^{b}\le z|\mathcal{D})-\prob(\textrm{TS}_{n}\le z)|=O\Big(\frac{L^{1/4}\log^{2}(NT)}{(1-\gamma)^{1/4}(\epsilon NT)^{1/8}}\Big)+O\Big(\frac{L^{-p/d}\sqrt{NT\log(NT)} }{(1-\gamma)^3}\Big),\\
		&&\sup_{z} |\prob(\textrm{TS}_{\infty}^{b}\le z|\mathcal{D})-\prob(\textrm{TS}_{\infty}\le z)|=O\Big(\frac{\sqrt{L}\log^2(NT)}{(1-\gamma)^{3/4}(\epsilon NT)^{1/8}})+O\Big(\frac{L^{-p/d}\sqrt{NT\log(NT)} }{(1-\gamma)^2}\Big),
	\end{eqnarray*}
	with probability at least $1-O(N^{-1}T^{-1})$.} 
\end{theorem}

\change{Theorem \ref{thm:size} derives the upper error bounds on the differences in distribution between the proposed test statistics and 
%implies that the limiting distribution of the proposed test can be well-approximated by 
the conditional distributions of the bootstrapped statistic given the data. In particular, the error bounds depend on five factors: (i) the minimal sample size $\epsilon NT$ in estimating the Q-function over specified time intervals; (ii) the number of basis functions $L$; (iii) the $(1-\gamma)^{-1}$ term which has a similar interpretation as the ``horizon'' in episodic tasks; (iv) the smoothness of the system dynamics, measured by $p$; (v) the dimension of the state space $d$. Under the given conditions in $L$, $\epsilon$, $p$, $d$, and as $\gamma$ remains bounded away from $1$, these bounds decay to zero, 
%It in turn 
implying that the size of the proposed test approaches the nominal level as the total number of observations diverges to infinity}. 

\change{Additionally, it is important to note that the second error bound for the normalized test statistics depends more heavily on the horizon $(1-\gamma)^{-1}$ compared to the other two test statistics. Specifically, this error term for the normalized test statistics is proportional to $(1-\gamma)^{-3}$, while for the other two, it is $(1-\gamma)^{-2}$. This increased dependence is due to the estimation of variance in the normalized test procedure. As such, in settings where the system is not overly smooth -- i.e., $p$ is small -- the second error term becomes the dominant factor in the error bound, leading to a higher type-I error for the normalized test. 
Such a finding aligns with our results in the real-data-based simulation; see Figure \ref{fig:real:result} of the Supplementary Materials.}

Finally, as commented in the introduction, the derivation of the consistency of the proposed test is complicated due to that we allow $L$ to grow with the number of observations. Specifically, when $L$ is fixed, the test statistic's limiting distribution can typically be derived using classical weak convergence theorems \citep{van1996}. However, these theorems become inapplicable as $L$ diverges with the sample size. This complexity arises because the dimension of the estimator $\widehat{\beta}$ also expands with $L$. To prove its asymptotic normality, it is necessary to demonstrate that $\widehat{\beta}$ converges to a Gaussian vector despite the growth of its dimension. To address this challenge, we develop a matrix concentration inequality for nonstationary MDPs in Lemma \ref{lemmamatrixnonstat}.

\subsection{Power of the Test}\label{sec:testcon}
We next establish the power property of the proposed test. In our theoretical analysis, we focus on a particular type of alternative hypothesis $\mathcal{H}_a$ where there is a single change point $T^* \in (T_0, T)$ such that 
\begin{eqnarray}\label{eqn:Ha}
Q_{T_0}^{\tiny{opt}}=Q_{T_0+1}^{\tiny{opt}}=\cdots=Q_{T^*-1}^{\tiny{opt}}\neq Q_{T^*}^{\tiny{opt}}=Q_{T^*+1}^{\tiny{opt}}=\cdots=Q_{T}^{\tiny{opt}}. 
\end{eqnarray}
%Under the asymptotics $T\to \infty$, we require $T^*\to\infty$ as well. 
Let $\Delta_1=T^{-1}\sum_{t=0}^{T-1}\Mean |Q_{T_0}^{\tiny{opt}}(A_t,S_t)-Q_T^{\tiny{opt}}(A_t,S_t)|$ and $\Delta_{\infty}=\sup_{a,s}|Q_{T_0}^{\tiny{opt}}(a,s)-Q_T^{\tiny{opt}}(a,s)|$ characterize the degree of nonstationarity. Specifically, the null holds if $\Delta_1$ or $\Delta_{\infty}$ equals zero and the alternative hypothesis $\mathcal{H}_a$ holds if $\Delta_1$ or $\Delta_{\infty}$ is positive. However, we remark that the proposed test is consistent against more general alternative hypothesis as well. See Section \ref{sec:num} for details. %Notice that in the definition of $\Delta_1$, we integrate over the observed state-action distribution $T^{-1} \sum_{t=0}^{T-1} \pi_t(a|s)p_t^b(s)$ where $p_t^b$ denotes the marginal distribution of $S_t$ under the behavior policy. %Alternatively, we can integrate over any reference distribution, in addition to the observed distribution, 
% the reference distribution can be taken for any measure that is absolutely continuous with respect the observed state-action distribution. 
For any two positive sequences $\{a_{N,T}\}_{N,T}$ and $\{b_{N,T}\}_{N,T}$, the notation $a_{N,T}\gg b_{N,T}$ means that $b_{N,T}/a_{N,T}\to 0$ as $NT\to \infty$. 

\smallskip

%\noindent \textbf{A1' (Uniform consistency under $\mathcal{H}_a$)}: Under the alternative hypothesis defined in \eqref{eqn:Ha}, both sets of Q-estimators $\{\phi_L^\top(a,s) \widehat{\beta}_{[T_0,T_1]}: T_1<T^* \}$ and $\{\phi_L^\top(a,s) \widehat{\beta}_{[T_1,T]}: T_1\ge T^* \}$ are uniformly consistent in $\ell_{\infty}$ norm.
%\noindent \textbf{A9' (Basis functions)}: $L$ is proportional to $(NT)^{c_5}$ for some $0<c_5<1/2$. \\
\noindent \textbf{A10 (Change point)}: $T_0 +\epsilon T<T^*<(1-\epsilon) T$ given the boundary removal parameter $\epsilon$.

\begin{theorem}[Power]\label{thm:power}
	Suppose A1-A10 hold. %Additionally, suppose that . 
	\begin{itemize}
		\item If $\Delta_1\gg (1-\gamma)^{-2}\sqrt{L(\epsilon NT)^{-1}\log (NT)}+(1-\gamma)^{-2}L^{-p/d}$, then the power of the test based on $\textrm{TS}_1$ \eqref{eqn:teststat1} is at least $1-O(N^{-1}T^{-1})$; 
		\item If $\Delta_{\infty}\gg (1-\gamma)^{-2}\sqrt{L(\epsilon NT)^{-1}\log (NT)}+(1-\gamma)^{-2}L^{-p/d}$, then the power of the test based on $\textrm{TS}_{\infty}$ \eqref{eqn:teststatinf} is at least $1-O(N^{-1}T^{-1})$; 
		\item If $\Delta_{\infty}\gg (1-\gamma)^{-2}\sqrt{L(\epsilon NT)^{-1}\log (NT)}+L^{-p/d}$, then the power of the test based on $\textrm{TS}_{n}$ \eqref{eqn:teststatninf} is at least $1-O(N^{-1}T^{-1})$.
	\end{itemize} 
\end{theorem}

%The assumption $T_0 +\epsilon T<T^*<(1-\epsilon) T$  
Assumption A10 is reasonable as we allow $\epsilon$ to decay to zero as the number of observations grows to infinity (see A2). %Under the given assumptions, the bias and standard deviation of the Q-function estimator are proportional to $O(\sqrt{L(\epsilon NT)^{-1}})$ and $O(L^{-p/d})$, respectively, up to some logarithmic factors. 
\change{The conditions on $\Delta_1$ and $\Delta_{\infty}$ essentially require the signal under $\mathcal{H}_a$ to be much larger than a certain lower bound to detect the change. In settings where the system is not overly smooth and $p$ is small, the second term will dominate the lower bound. It is evident that the lower bounds for the $\ell_1$-type and unnormalized maximum-type tests depend more heavily on the horizon $(1-\gamma)^{-1}$ than the normalized test. As such, when $\Delta_1$ and $\Delta_{\infty}$ are of the same order of magnitude, both the $\ell_1$-type test \eqref{eqn:teststat1} and unnormalized maximum-type test \eqref{eqn:teststatinf} require stronger conditions on the signal strength to detect the alternative hypothesis than the normalized test. This observation aligns with our empirical study, where we find that the normalized test generally achieves better power properties than the other two tests (see Figure \ref{fig:real:result} of the Supplementary Materials).} To %guarantee the proposed three tests have good power properties
further boost power, we use cross-validation to select the number of basis functions for all three tests, as discussed in Section \ref{sec:imp} of the Supplementary Materials. This ensures the bias and standard deviation of the Q-function estimator are approximately of the same order of magnitude. 
%In A1', we further require the Q-estimators before and after the change to be consistent. Similar to Theorem \ref{thmQ}, we can show that it holds when FQI is employed for estimating the optimal Q-function.

%To conclude this section, we remark that we focus on establishing the consistency of the test in this section. The change point detection procedure 
%It is worthwhile to mention that establishing the power property of the test requires a less stringent condition on $L$ than deriving the size property. Specifically, we require $L$ to grow at a rate of $o(\sqrt{NT})$ in Theorem \ref{thm:power}. In contrast, this condition is strengthened to $L=o(N^{1/4}T^{1/4})$ in Theorem \ref{thm:size}. 
%The condition $L\asymp (NT)^{c_3}$ implies that $L$ is proportional to $(NT)^{c_3}$. 

%The derivation of the asymptotic property is complicated due to 

%%%%%%%%%%%%%%%%%%%%%%%%%%%%%%%%%%%%%%%%%%%%%%%%%%%%%%%%%%%%%
\section{Simulations}\label{sec:num}

In this section, we conduct simulation studies to evaluate the finite sample performance of the proposed method and compare against common alternatives. %In Section \ref{sec:imp}, we detail the implementation of the proposed tests. 
Section \ref{sec:sim:1d} presents results of the proposed offline
testing and change point detection methods based on four generative models with different nonstationarity scenarios (see Table \ref{tab:simscenario1d}). Section \ref{subsect:onlineevaluation} further demonstrates the usefulness of the proposed method in an online setting as data accumulate. 
%for toy simulation examples in 1-dimensional state space, and 
In %Section \ref{sec:sim:real}, 
Section \ref{sec:realdatabasedsimu} of the Supplementary Materials, 
we simulate data to mimic the data setup in the motivating application of IHS. %We detail our implementation in Section \ref{sec:imp} of the Supplementary Materials. 
%contains results for higher dimensional state space to mimic IHS scenario. 
%All simulation settings involve only one change point. 
All simulation results are aggregated over 100 replications. %{\color{red} [Is this 500 overall; if not, can remove this sentence.]}
%All statistical tests are rejected at $\alpha = 0.05$ level. Each simulation scenario is repeated 500 times. 
%\textcolor{blue}{[All: how many figures are we allowed to include in the main paper and Appendix?]}

%%%%%%%%%%%%%%%%%%%%%%%%%%%
%\subsection{Simulation I} \label{sec:sim:1d}
%We begin with the following simple generative model with a one-dimensional state variable. The initial state $S_0$ is generated according to $N(0, 0.5)$. 
%We design four generative models to 

%In the following, we 
% conduct two sets of simulation studies to 
%a) assess the performance of the proposed offline testing and change point detection methods %with empirical demonstration of detection delay, %using offline data, 
%and b) %compare the value of optimal policies estimated against alternative methods 

\smallskip

\subsection{Offline Testing and Change Point Detection}\label{sec:sim:1d} 
We consider four nonstationary data generating processes with one-dimensional states and binary actions where the nonstationarity occurs in either the state transition function or the reward function, as listed in Table \ref{tab:simscenario1d}. For nonstationary functions, both abrupt (e.g., the underlying function is piece-wise constant) and smooth changes are considered. 
\change{Specifically, in the first two scenarios, the transition function $\mathcal{T}_t$ is stationary whereas the reward function $r_t$ is piecewise constant or varies smoothly over time, respectively}. The last two scenarios involve stationary reward functions and nonstationary transition functions. See Section \ref{sec:smoothtransition} of the Supplementary Materials for more details about the data generating processes in these four scenarios. 

In all scenarios, we set $T = $100 and simulated offline data with sample sizes $N=25, 100$. The true location of the change point $T^*$ was set to $50$. We first applied each of the proposed three tests to the time interval $[T-\kappa,T]$ to detect nonstationarity, where $\kappa$ took value from a equally-spaced sequence between $25$ and $75$ with increments of $5$. According to the true data generating mechanisms, when $\kappa \leq 50$, the null of no change point over $[T-\kappa,T]$ holds; the alternative hypothesis holds if $\kappa > 50$. 
The actions were generated i.i.d. according to a Bernoulli random variable with a success probability of $0.5$.

\begin{table}[!t]
	\centering
	\begin{tabular}{r c c} 
		\toprule
		& State transition function & Reward function \\
		\midrule
		(1) & Time-homogeneous & Piecewise constant  \\ 
		(2) & Time-homogeneous & Smooth \\
		(3) & Piecewise constant & Time-homogeneous \\
		(4) & Smooth & Time-homogeneous \\
		\bottomrule
	\end{tabular}
	\caption{Simulation scenarios with different types of nonstationarity in Sections \ref{sec:sim:1d} and \ref{subsect:onlineevaluation}.}
	\label{tab:simscenario1d}
\end{table}

%We generate the initial state as $S_{0,0} \sim \cN (0, 0.5)$ and independent noise variables $z_{0,t} \sim \cN (0, 0.25), t \in [T] \equiv \{ 0, 1, \ldots, T \}$. The binary random actions are $A_{0,t} \in \{-1, 1\}$ with $ P(A_{0,t} = 1) =  P(A_{0,t} = -1) = 0.5, t \in [T-1]$. A collection of state and reward trajectories $\{ S_{i,t}, R_{i,t}, A_{i,t}, S_{i, t+1} \}_{i \in [N], t \in [T]}$ are generated with $N \in \{25, 100\}$. We set the discount factor as $\gamma \in \{0.8, 0.9, 0.95\}$. We test the null hypothesis of stationarity at a sequence of $\kappa$'s between $t = 25$ and 75 at every 5 time points.

Figures \ref{fig:1d:rej} and \ref{fig:1d:N100}(a) in the Supplementary Materials show the empirical rejection probabilities of each proposed test, when $N=25$ and $100$ respectively. 
First, in all settings, each test properly controls the type-I error. Second, the power increases with $\kappa$ due to inclusion of more pre-change-point data into the interval $[T-\kappa,T]$. The power also increases with the sample size $N$, demonstrating the consistency of our tests. 
Third, as expected, gradual changes are more difficult to detect than abrupt changes. This is evident in Figure \ref{fig:1d:rej}, where when $\kappa = 55$, the power of the proposed test in scenarios with a smooth reward or state transition function is smaller than scenarios with a piecewise constant function. 
Finally, the maximum-type tests \eqref{eqn:teststatinf} and \eqref{eqn:teststatninf} achieve slightly higher power than the $\ell_1$-type test \eqref{eqn:teststat1} when $N = 25$, whereas the powers of the three tests become indistinguishable when $N = 100$. 

% \blue{
% The change point estimation results are consistent across the three tests in the majority of cases, as displayed in Figure \red{XX} of the supplementary material.
% We now look at cases where the results of the three tests do not agree with each other. In particular, we focus on the four synthetic data settings in Section \ref{sec:sim:1d} of the revised paper. 

%     \begin{figure}
%         \centering
%         \includegraphics[width=0.5\linewidth]{}
%         \caption{Caption}
%         \label{fig:enter-label}
%     \end{figure}
% }

\begin{figure}[!t]
	\centering
	\includegraphics[width=0.9\textwidth]{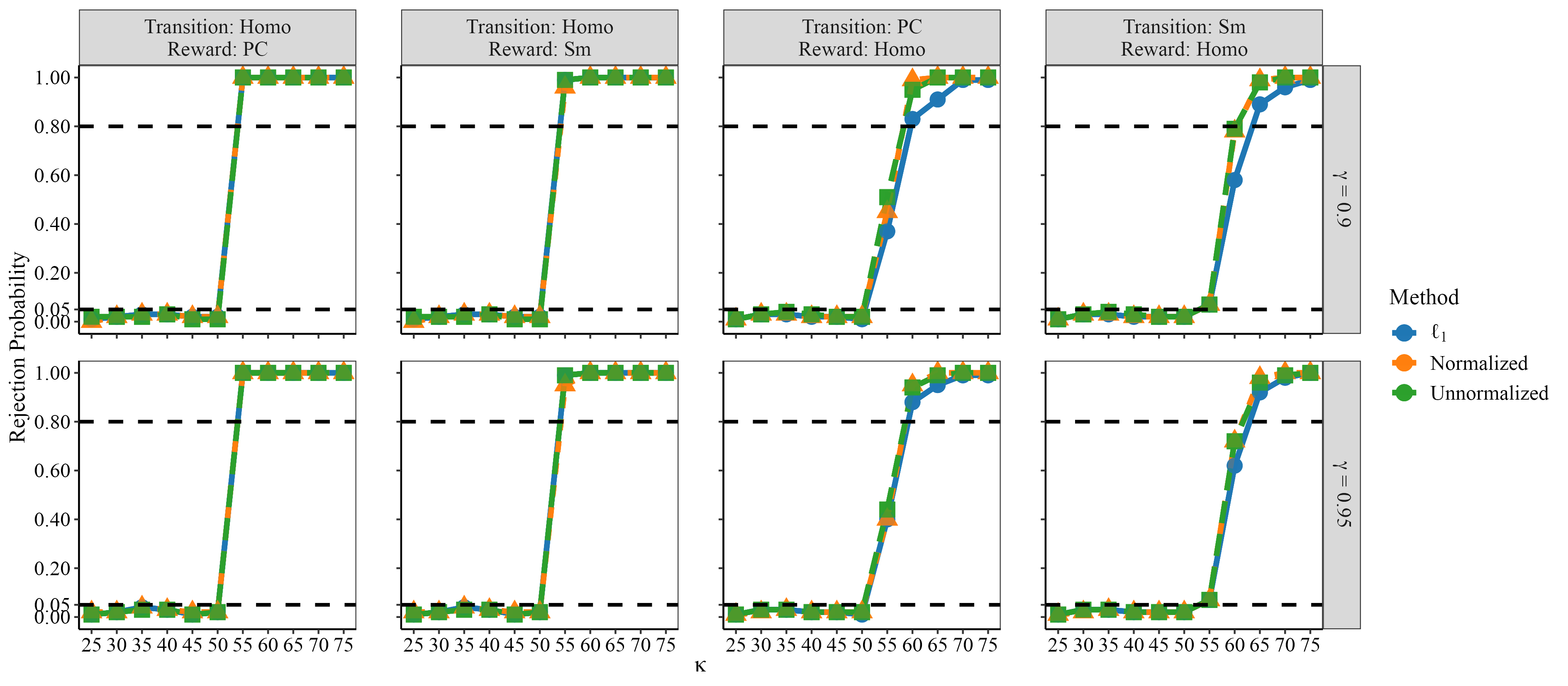} 
\caption{Empirical type-I errors and powers of the proposed test and their associated 95\% confidence intervals under settings described in Section \ref{sec:sim:1d}, with $N=25$. Abbreviations: Homo for homogeneous, PC for piecewise constant, and Sm for smooth.}
\label{fig:1d:rej}
\end{figure}

Next, we investigate the finite sample performance of the estimated change point location  $\widehat{T}^*$.  
Figures \ref{fig:1d:cpdist} and \ref{fig:1d:N100}(b) in the Supplementary Materials depict the distribution of $\widehat{T}^*$ in each simulation scenario. It can be seen that in the first two scenarios with abrupt changes, the estimated change points concentrate on $50$, which is the true change point location, yielding a minimal detection delay. In the last two scenarios with smooth changes, the estimated change points have a wider spread especially when $N=25$. This results in a marginally extended detection delay. However, in the majority of cases, these estimators are still close to 50.

\begin{figure}[!t]
    \centering
    \includegraphics[width=0.9\textwidth]{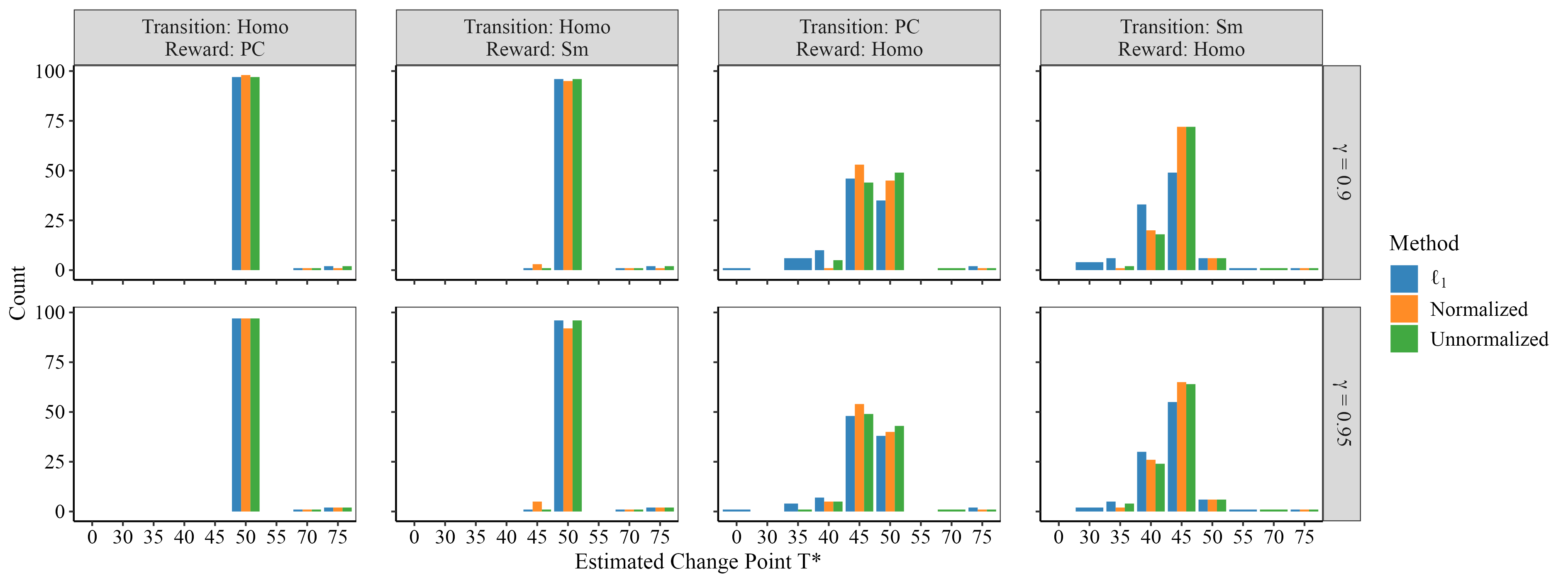}
    \caption{Distribution of detected change points under simulation settings in Section \ref{sec:sim:1d} with $N=25$.}
    \label{fig:1d:cpdist}
\end{figure}

\subsection{Online Evaluation}\label{subsect:onlineevaluation} 
Finally, we %perform value estimation and 
illustrate how the proposed change point detection method can be coupled with existing state-of-the-art RL algorithms for policy learning in nonstationary environments. In each simulation, we first simulated an offline dataset as discussed earlier with $T=100$ and $N=200$. We next applied our proposal to identify the most recent change point location $\widehat{T}^*$ and estimated the optimal policy 
%compute the optimal policy based on the estimated change point and compare it with some baseline methods. \violet{We estimate the optimal value using an online evaluation scheme, where we update the optimal policy as a batch of new data are collected. }In each simulation, after computing $\widehat{T}^*$ \violet{from the initial simulated data on $[0, T]$}, we estimate the optimal mean policy 
using the data subset $\{ (S_{i,t}, R_{i,t}, A_{i,t}): 1 \leq i \leq N, \widehat{T}^* \leq t \leq T \}$. As commented earlier, the resulting estimated policy can be used for treatment recommendation after study end time $T$. Specifically, we used a decision tree model \citep{myles2004introduction} to approximate $Q^{\tiny{opt}}$ to obtain interpretable policies for healthcare researchers. 
%We employed decision tree regression to approximate the Q-function and estimated the tree using FQI.
% We couple FQI with decision tree regression to update the Q-estimator in each iteration 
Through FQI, we transformed the estimation of $Q^{\tiny{opt}}$ into an iterative regression problem (see \eqref{eqn:updateQinFQI}) and employed decision tree regression to update the Q-estimator at each iteration. 
The decision tree model involves hyperparameters such as the maximum tree depth and the minimum number of samples on each leaf node. We used 5-fold cross validation to select these hyperparameters from $\{3,5,6\}$ and $\{ 50, 60, 80\}$, respectively. %(see Supplementary Section \ref{sec:imp}). 
%using the criterion in \eqref{eq:cv:kerneldist}. 

Next, for each of the 200 subjects, we 
sequentially applied our procedure for online policy learning as data accumulated to maximize their cumulative reward. Specifically, we %considered an online setting and 
assumed the number of change points after $T=100$ followed a Poisson process with rate $1/50$. In other words, we expected a new change point to occur every 50 time points. We set the termination time $T_{end} = 300$, yielding 3 to 4 change points in most simulations. 
We similarly considered four different types of change points listed in Table \ref{tab:simscenario1d}. Whenever a new change point occured, the effect of the action on the state transition or reward function was reversed. We further considered three different settings with strong, moderate and weak signals by varying the magnitude of treatment effect.
For instance, suppose we have two change points $T_1$ and $T_2$ after $T=100$. In Scenario (1) with a piecewise constant reward function, we set $r(s,a)$ to $1.5 \delta a s$ when $t\in [100, T_1)$ or $[T_2,300]$ and $-1.5\delta as$ when $t\in (T_1, T_2]$ where $\delta$ measures the treatment effect equals $1,0.5$ and $0.3$ in strong-, moderate- and weak-signal settings, respectively.

\change{Finally, we assumed that online data came in batches regularly at every $L = 15$ time points starting from $T=100$. The first online data batch was generated according to an $\epsilon$-greedy policy that selected actions using the estimated optimal policy $\widehat{\pi}$ computed based on the data subset in the time interval $[\widehat{T}^*,T]$ with probability $1-\epsilon$ and a uniformly random policy with probability $\epsilon$. Let $T^{*(0)}=\widehat{T}^*$. 
Suppose we have received $k$ batches of data. %and we have performed $d$ times of policy learning. 
We first apply the proposed change point detection method on the data subset in $[T^{*(d-1)}, T+dL]$ to identify a new change point $T^{*(d)}$. If no changes are detected, we set $T^{*(d)}=T^{*(d-1)}$. 
We next update the optimal policy based on the data subset in $[T^{*(d)}, T+dL]$ and use this estimated optimal policy (combined with the $\epsilon$-greedy algorithm) to generate the $(k+1)$-th data batch. We repeat this procedure until the termination time $T_{end}$ is reached and aggregate all immediate rewards obtained from time $T$ to $T_{end}$ over the $200$ subjects to estimate the average value.}

Comparison is made among the following methods: 
% \smallskip
% After computing the estimated policy, we simulate 300 new subjects following such a policy for 100 time points after $T^*$ and aggregate the discounted rewards over these subjects to estimate the expected return (e.g., value) under that policy. The proposed three tests yield similar policies. We report the results based on the integral-type test only and
%We further 
% compare them with the following baseline methods: %take the Monte Carlo mean value under each policy as an estimate of the value of that policy.

% \begin{figure}[t]
%     \centering
%     \includegraphics[width=\textwidth]{fig/1d_box_optvalue_dt_N25.pdf}
%     \caption{Distribution of Monte Carlo differences in mean rewards under the proposed policy and those under policy computed based on other baseline methods, under settings in Section \ref{sec:sim:1d} with $N=25$.} 
%     \label{fig:1d:value}
% \end{figure}

% \violet{\noindent {\textbf{Overall}}: After each new batch of data are collected, standard policy optimization method that uses all prior data;  \\
% \noindent {\textbf{Random}}: After each new batch of data are collected, policy optimization with a randomly assigned change point location with 0.5 probability, and with no change point (using all data) with 0.5 probability ; \\
% \noindent {\textbf{Kernel}}: The kernel-based approach developed by \cite{domingues2021kernel};\\
% \noindent {\textbf{Oracle}}: After each new batch of data are collected, policy optimization method as if the oracle change point location were known in advance.}
\noindent {\textbf{Proposed}}: The proposed $\ell_1$-type test \eqref{eqn:teststat1} (the other two tests yield similar change points and  policies. Their results are not reported to save space);  \\
\noindent {\textbf{Oracle}}: The ``oracle'' policy optimization method that works as if the oracle change point location were known in advance; \\
\noindent {\textbf{Overall}}: Standard policy optimization method that uses all the data;  \\
\noindent {\textbf{Random}}: Policy optimization with a randomly assigned change point location;\\
\noindent {\textbf{ODCP}}: The online Dirichlet change point approach proposed by \cite{padakandla2020reinforcement}; \\
\noindent {\textbf{MBCD}}: The model-based RL context detection approach proposed by \cite{Alegre2021Minimum}; \\
\noindent {\textbf{Kernel}}: The kernel-based approach developed by \cite{domingues2021kernel}.

For fair comparisons, we used FQI and decision tree regression to compute the optimal Q-function for all methods. %Specifically, 
To implement the random method, after a new batch of data arrived, 
we randomly picked a time point uniformly from the new batch
% interval $[T^{*(k-1)}, T+kL]$ 
as the next change point location 
% $T^{*(k)}$ 
and computed the optimal Q-function based on the observations that occurred afterwards.
% time $T^{*(k)}$.
% at the $l$-th FQI iteration, we consider the following objective function,
% \begin{eqnarray}\label{eqn:FQIkernel}
% Q^{(l+1)}=\argmin_Q \sum_{i,t} K\left(\frac{T-t}{Th}\right)\left\{R_{i,t}+ \gamma \max_a Q^{(l)}(a, S_{i,t+1})-Q(A_{i,t},S_{i,t}) \right\}^2,
% \end{eqnarray}
% where $K(\cdot)$ denotes the Gaussian RBF basis and $h$ denotes the associated bandwidth parameter taken from the set $\{0, 0.1, 0.2,0.4,0.8,1.6\}$. According to \eqref{eqn:FQIkernel}, the kernel-based method assigns larger weights to more recent observations to deal with nonstationarity.
% After we receive the $k$th data batch, we sample $B \gg T$ data slices across all individuals from $\{(S_{i,t},A_{i,t},R_{i,t},S_{i,t+1}; 1 \leq i \leq N)\}_{0 \le t < T+kL}$
% with weights proportional to $K((T-t)/(Th))$ and apply the decision tree regression to these samples to solve \eqref{eqn:FQIkernel}. 
The oracle method was implemented by repeatedly using observations that occurred after the oracle change point to update the optimal policy. 

\change{The last three methods -- ODCP, MBCD and Kernel -- are existing state-of-the-art nonstationary RL approaches. In particular, similar to ours, both ODCP and MBCP are change-point-based methods that apply stationary RL to the best data segment of stationarity identified by a change point detection algorithm. Among the two algorithms, MBCD is model-based, which uses neural networks for estimating the reward and transition functions, along with a likelihood ratio test for detecting change point. ODCP 
% On the contrary, ODCP is model-free. It 
was originally proposed by \citet{kj2022change} for handling compositional multivariate data and later adapted by \cite{padakandla2020reinforcement} for RL in nonstationary environments. %After mapping the state-action pairs into the unit cube, ODCP assumes that the joint state-action distribution follows a Dirichlet distribution, and applies
%the Dirichlet likelihood test to 
It is model-free. Specifically, it does not model the reward and transition functions, but applies the likelihood ratio test to detect changes in the marginal state-reward distribution. As such, this algorithm is not consistent: it might detect changes in the behavior policy rather than in the Q- or transition function \citep{wang2023robust}. 
Finally, the kernel-based method is non-change-point-based. It uses a kernel to assign larger weights to more recent observations, and smaller weights to past observations to deal with nonstationarity.
To implement this method, we used the Gaussian RBF kernel with bandwidth parameters chosen from $\{0, 0.1, 0.4, 1.6\}$. More details about these three methods can be found in Section \ref{sec:sim:evaluation:baseline} of the Supplementary Materials.}  
%Implementations of the MBCD and kernel methods are detailed in Supplementary Section \ref{sec:sim:evaluation:implementation}.
%}
% To implement the oracle method, we use observations that occur after the oracle change point to compute the optimal Q-function. 
%Each of the aforementioned method requires to estimate the optimal Q-function. To ensure consistency, 

\begin{figure}[tp]
\centering
    \includegraphics[width=\textwidth]{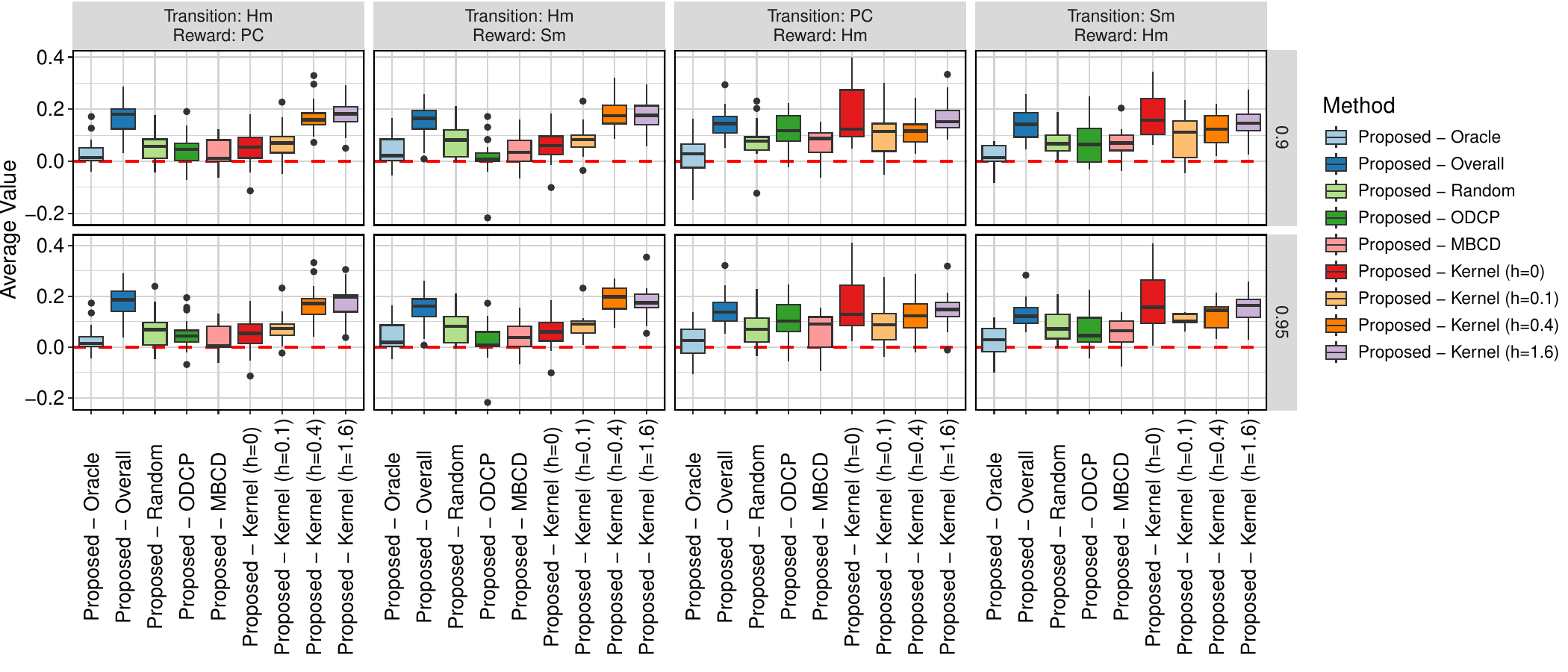}
\caption{
% \textcolor{red}{ML: to be changed. Still waiting for results to finish running.} 
Distribution of the difference between the average value $(T_{end}-T)^{-1}\sum_{t=T+1}^{T_{end}} \Mean R_t$ under the proposed policy and those under policies computed by other baseline methods, under settings in Section \ref{sec:sim:1d} with strong signal-to-noise ratio. The proposed policy is based on the change point detected by the $\ell_1$-type test \eqref{eqn:teststat1}. In all scenarios, we find the normalized or unnormalized tests \eqref{eqn:teststatinf} and \eqref{eqn:teststatninf} yield similar average values.}
\label{fig:1d:value}
\end{figure}

Figure \ref{fig:1d:value} reports the difference between the average value under the proposed policy and those under policies estimated based on these baseline methods 
% with $N=25$ and $100$ 
in the ``strong signal'' setting (i.e., the treatment effect $\delta=1$). Results with moderate and weak signals %-to-noise ratios 
are available 
in Supplementary Section \ref{app:sec:sim:result}. We briefly summarize a few notable findings: 
\begin{enumerate}
\item First, the proposed method achieves much larger average values compared to the ``overall'' method, %indicating 
demonstrating %detrimental consequences 
the inferiority of the best policy learned without acknowledging the nonstationarity. %in policy learning if  nonstationarity is ignored. 
\item Second, the proposed method is no worse and often better than kernel-based approaches in most cases. In addition, as shown in Figure \ref{fig:1d:value}, kernel-based method can be sensitive to the choice of the kernel bandwidth %. A poor choice of $h$ would yield a poor policy, 
and it remains unclear how to determine this tuning parameter in practice. These results highlight the necessity of change point detection in policy learning, demonstrating the benefits of change-point-based methods in nonstationary RL. 
\item Third, the proposed method outperforms the ``random'' method in almost all cases. It also achieves larger average values than ODCP in most cases. As ODCP is not guaranteed to consistently detect the change point, these results imply that correctly identifying the change point location is essential to policy optimization in nonstationary environment. 
\item Finally, the model-based MBCD method produces smaller average rewards in general than the proposed method. This demonstrates the advantage of our model-free method, which is less prone to model misspecification. %does not rely on parametric assumptions, which are prone to misspecification, about the state and reward dynamics.
\end{enumerate}

\begin{comment}
\begin{table}[t]
\centering
\begin{tabular}{lcc}
	\toprule
	Difference  & \multicolumn{2}{c}{Mean value (s.e)} \\
	(in value)       & $\gamma=$0.9 & $\gamma=$0.95 \\
	\midrule
	Proposed - Oracle & -0.06 (0.04) &  -0.03 (0.03) \\
	Proposed - Overall & 22.92 (1.04) &  35.92 (2.42) \\
	Proposed - Random &  5.75 (0.86) &   9.62 (1.49) \\
	\bottomrule
\end{tabular}
\caption{Simulation II: Value difference between the proposed method and the oracle, overall and random method. %The numbers are produced with 100 data replications; 
	Positive numbers indicate higher values based on the proposed method.}
\label{tab:real:value}
\end{table}
\end{comment}

%%%%%%%%%%%%%%%%%%%%%%%%%%%%%%%%%%%%%%%%%%%%%%%%%%%%%%%%%%%%%
\section{Application to Intern Health Study} \label{sec:data}

% In recent years, mobile technology has the potential to revolutionize the format of healthcare by providing convenient access to services and personal data tracking. Physical activity is one of the areas that benefits from the advantage of mobile devices which can deliver just-in-time adaptive interventions (JITAIs) to individuals at the right time with the right content. 

The 2018 Intern Health Study (IHS) is a micro-randomized trial (MRT) that seeks to evaluate the efficacy of push notifications sent via a customized study app upon proximal physical and mental health outcomes \citep{necamp2020assessing}, a critical first step for designing effective just-in-time adaptive interventions. Over the $26$ weeks, each study subject was re-randomized weekly to receive or not to receive activity suggestions; daily self-reported mood scores were assessed via ecological momentary assessments, a method repeatedly recording subjects' behaviors in real time and in their natural environment; step count and sleep duration were measured by Fitbit. In this paper, we focus on policy optimization for improving time-discounted cumulative step counts under the infinite horizon setting. 
As discussed in Section \ref{sec:introduction}, determining the optimal policy for delivering prompts is challenging due to potential nostationarity that results in changes in treatment effects.
Here we demonstrate how to use the proposed method to detect change point and perform optimal policy estimation in the presence of potential temporal nonstationarity.

\subsection{Data and method: Setup} \label{sec:data_setup}

\change{Let $S_t$ denote a $4$-dimensional state vector comprised of the following: (i) the square root of average step count in the previous week $t-1$; (ii) the cubic root of average sleep minutes in week $t - 1$; (iii) the average mood score in week $t-1$; (iv) the square root of average step count in week $t-2$. All these variables are centered and scaled  \citep{necamp2020assessing}. The binary action $A_t = 1$ ($0$) corresponds to pushing (not pushing) an activity message at the start of week $t$. The randomization probabilities are known under MRT: $\prob(A_t = 1) = 1 - \prob(A_t = 0) = 1/4$. The reward $R_t$ is defined as the average step count in week $t$. }
To resemble a real-time evaluation scenario, we divide the data of 26 weeks into two trunks: we perform change point detection and estimate the optimal policy based on data collected in the first $T = 22$ weeks (training data batch), and then evaluate the estimated policy on data in the remaining 4 weeks (evaluation data batch) assuming that there is no change point in the final 4 weeks. 
To implement change point detection, we set the boundary removal parameter $\epsilon = 0.08$ and search for change points within $[5, 18]$. 
% The number of basis functions is selected via 5-fold cross validation (
%See Section \ref{sec:imp} of the Supplementary Materials for implementation details. 
We focus on three specialties: emergency ($N = 141$), pediatrics ($N = 211$), and family practice ($N = 125$). One consideration is that work schedules and activity levels vary greatly across different specialties, and thus medical interns might experience distinct change points. Stratification by specialty may improve homogeneity of the study groups so that the assumption of a common change point is more plausible. %see Section \ref{sec:hetero} for discussions on potential extensions to heterogeneous change points. %{\color{blue}[All: should we discuss change points that may differ by subjects or unknown clusters of subjects?]}

% in which subjects likely had the same change point, Similar to many other mHealth studies, the percentage of missing data in IHS increases nearly linearly with week in study. We adopt the multiple imputation method in \cite{necamp2020assessing} to address the issue. 

\subsection{Results}
\label{sec:data_results}
\begin{figure}[t]
\begin{minipage}[c]{\linewidth}
	\centering
	\includegraphics[width=0.8\textwidth]{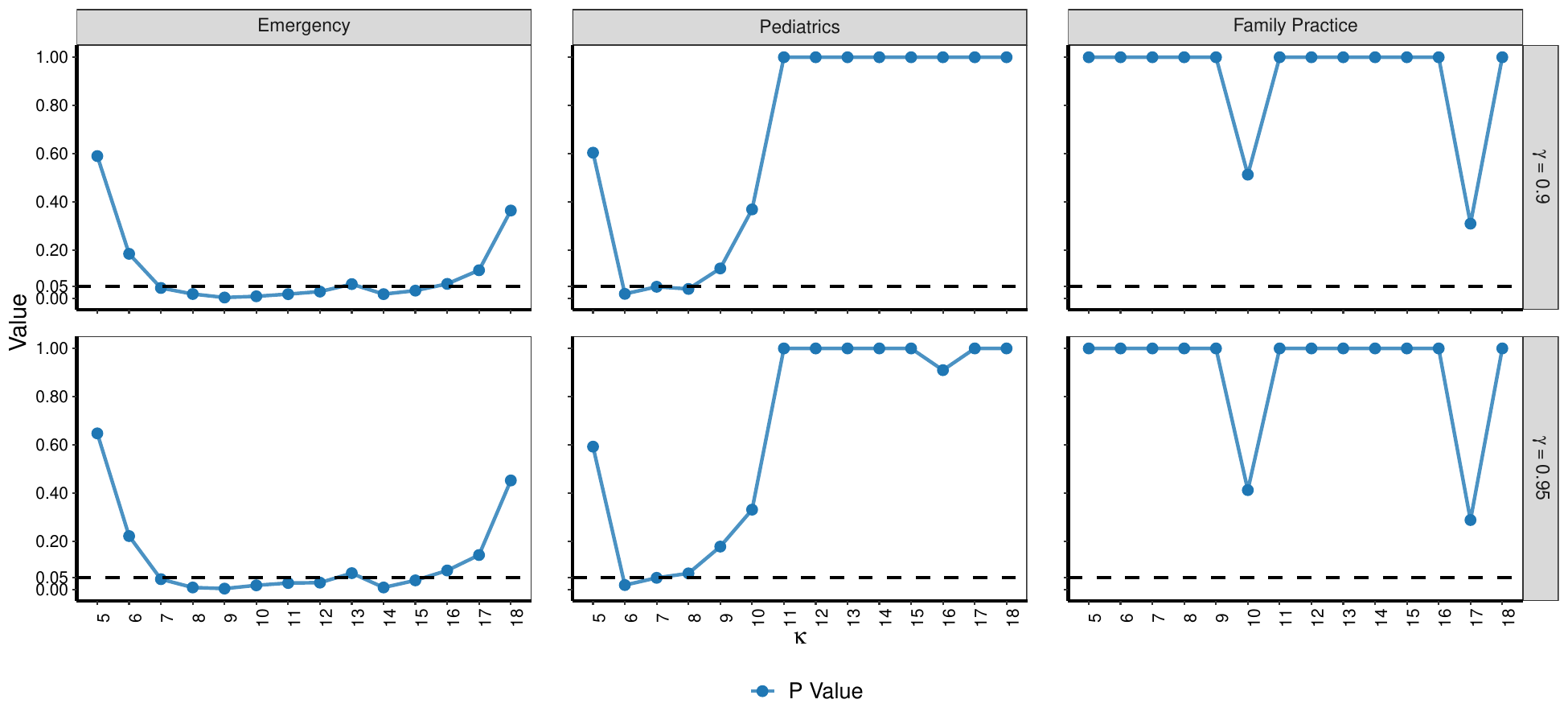}
\end{minipage}
\caption{$p$-values over different values of $\kappa$ (the number of time points from the last time point $T$) under $\gamma=0.9$ (top) and $0.95$ (bottom) among the three specialties considered in IHS. 
% The specialties are emergency (left), pediatrics (middle) and family practice (right). 
% Many insignificant $p$-values are close to 1, due to the use of the aggregation method (see Section \ref{sec:imp}), which tends to increase insignificant $p$-values. 
}
%for three specialties in IHS and two discount factors $\gamma=0.9,0.95$. }
\label{fig:data:pvalue}
\end{figure}

Figure \ref{fig:data:pvalue} plots the trajectories of $p$-values using $\ell_1$-type test statistic \eqref{eqn:teststat1}; the results are similar when maximum-type tests \eqref{eqn:teststatinf} and \eqref{eqn:teststatninf} were applied to the data (not reported here). We consider $\gamma = 0.9$ or $0.95$, which produce similar results. First, we notice that in the pediatrics and family practice specialties, many $p$-values are close to 1. 
This is due to the use of the $p$-value aggregation method \citep[see Section \ref{sec:imp} of the Supplementary Materials for details]{meinshausen2009p}, which tends to increase insignificant $p$-values and reduce the type-I error. 
Second, the emergency specialty displays roughly monotonically decreasing $p$-values over time, whereas at the largest few $\kappa$ values the $p$-values rise up due to the limited effective sample size at the boundary. %, resulting in a wide U-shape of the $p$-value curves.
% The emergency specialty displays roughly monotonically decreasing $p$-values over time, indicating a change point at $\kappa = 6$ for both $\gamma = 0.9$ and $\gamma = 0.95$. Nevertheless, at the largest few $\kappa$ values the $p$-values rise up, indicating a possible second change point at around $\kappa = 16$, after which the time series dynamics shifts back to the one before the first change point $\kappa = 6$. 
Third, the U-shaped $p$-value trajectory of the pediatrics specialty shows evidence for multiple change points. Specifically, when only a single change point exists, the significant $p$-values are likely to decrease with $\kappa$. The U-shaped $p$-value trajectory can occur only when the data interval contains at least two change points and the system dynamics after the second change point is similar to that before the first change occurs, yielding a small CUSUM statistics. Because we focus on the latest detected change point (first $\kappa_{j_0-1}$ where $\kappa_{j_0}$ results in a rejection of the null) to inform the latest data segment to use for optimal policy estimation, we find $\kappa_{j_0-1} = 6$ for the emergency specialty and $\kappa_{j_0-1} = 5$ for the pediatrics specialty, for both choices of $\gamma$. %under $\gamma = 0.9$ and $\gamma = 0.95$. 
Fourth, the $p$-value trajectories of the family practice specialty (mostly close to 1) are above the significance threshold, indicating the stationarity assumption is compatible with this data subset. We therefore estimate the optimal policy using data from all the time points for the family practice specialty.

%\violet{
We next compare the proposed policy optimization method with three other methods: 1) overall, 2) random, which were described in Section \ref{sec:sim:1d}) and 3) behavior, which is the treatment policy used in the completed MRT. The data of the first 22 weeks are used to learn an optimal policy $\hat{\pi}^{opt}$ through FQI and decision tree regression. In particular, the proposed method uses data after the estimated change point location whereas the overall method uses all the training data. Similar to simulations, hyperparameters of the decision tree regression are selected via 5-fold cross-validation. %similar to the simulation sections. 
Next, based on the evaluation data, we %employed 
applied FQE 
to the testing data of the remaining 4 weeks to evaluate the $\gamma$-discounted values (i.e., $J_{\gamma}(\pi)$) of these estimated optimal policies. Results are reported in Table \ref{tab:data:values}. It can be seen that the proposed method achieves larger values when compared to ``random'' and ``behavior'' in all cases, demonstrating the need for change point detection and data-driven decision making. In the following, we focus on comparing the proposed method against ``overall''. 
In the emergency specialty, the optimal policy estimated using data after the detected change point improves weekly average step count per day by about 130 $\sim$ 170 steps relative to the estimated policy based on the overall method. %using data from all time points (``Overall''). 
In the pediatrics specialty, however, %the policy under the overall method outperforms the policy under our proposed method 
the overall method achieves a larger weekly average step count by about 40 $\sim$ 112 steps per day. %on weekly average. 
Recall that the proposed method only uses data on $t \in [T - \kappa_{j_0-1}, T] = [17, 22]$ for policy learning. %are involved in the proposed policy estimation, whereas data from all time points are involved in the overall policy estimation. 
As commented earlier, there are likely two change points in the pediatrics specialty and according to Fig \ref{fig:data:pvalue}, the system dynamics after the first most recent change point %($T - \kappa_{j_0-1} = 16$ for emergency and $17$ for pediatrics) and 
are very similar to those before the second most recent change point. %(around 6 for emergency and 14 for pediatrics) were likely highly similar as shown in Fig \ref{fig:data:pvalue}. 
As a result, the overall method pools over more data from similar dynamics, %than the proposed policy and results in a better optimal policy by utilizing more information effectively
resulting in a better policy. This represents a bias-variance trade-off. In settings %where the most recent stationary segment is short, 
with a U-shaped $p$-value trajectory and the most recent change point is close to the second most recent one, it might be sensible to borrow more information from the historical data. 
%Therefore, under circumstances where the most recent stationary segment in the system dynamics is short in length, 
%we need to be cautious about whether to optimize the policy using a relatively small amount of stationary data, or to borrow more information from the possibly nonstationary segment, i.e., a bias-variance trade-off.
Finally, the proposed and overall methods have equal values in the family practice specialty since no change point is identified. %These results show that policy improvement can be achieved by identifying a change point and applying standard reinforcement learning to data after the detected change point that is compatible with the stationarity assumption.%}

% We next compare the proposed policy optimization method with two other methods: 1) overall, which was described in Section \ref{sec:sim:1d}) and 2) behavior, which is the treatment policy used in the completed MRT. We first split all the subjects into training and evaluation data sets with a ratio of 3/2. %: one for policy learning and the other two for The training data are used to learn an optimal policy $\hat{\pi}^{opt}$ through FQI and decision tree regression, based on the estimated change point location. Hyperparameters of the decision tree regression are selected via 5-fold cross-validation, similar to the simulation sections. Next, based on the evaluation data, we employed fitted-Q evaluation \citep[FQE,][]{le2019batch}, which is designed for off-policy value evaluation. 
% We normalize the value estimates by $1 - \gamma$ to reflect value per time point, as shown in Table \ref{tab:data:values}. In the emergency specialty, the optimal policy estimated using data after the detected change point improves weekly average step count per day by about 170 $\sim$ 200 steps relative to the estimated policy using data from all time points (``Overall''). The proposed and overall methods have equal values in the family practice specialty because no change point is identified. These results show that policy improvement can be achieved by identifying a change point and applying standard reinforcement learning to data after the detected change point that is compatible with the stationarity assumption.

\begin{table}[!ht]
\centering
% \begin{tabular}{@{}cccc@{}}%
\begin{tabular}{@{}ccccc@{}}%
\toprule%
Number of Change Points&Specialty&Method&$\gamma=0.9$&$\gamma=0.95$\\%
\midrule%
\multirow{4}{*}{$\geq 1$}&\multirow{4}{*}{Emergency}&Proposed&8237.16&8295.99\\%
&&Overall&8108.13&8127.55\\%
&&Behavior&7823.75&7777.32\\%
&&Random&8114.78&8080.27\\%
\midrule%
\multirow{4}{*}{$\geq 2$}&\multirow{4}{*}{Pediatrics}&Proposed&7883.08&7848.57\\%
&&Overall&7925.44&7960.12\\%
&&Behavior&7730.98&7721.29\\%
&&Random&7807.52&7815.30\\%
\midrule%
\multirow{4}{*}{0}&\multirow{4}{*}{Family Practice}&Proposed&8062.50&7983.69\\%
&&Overall&8062.50&7983.69\\%
&&Behavior&7967.67&7957.24\\%
&&Random&7983.52&7969.31\\\bottomrule%
\end{tabular}%
\caption{Mean value estimates using decision tree in analysis of IHS. Values are normalized by multiplying $1 - \gamma$. %{\color{blue}[Chengchun, add more details about the caption]} 
	All values are evaluated over 10 splits of data.}
\label{tab:data:values}
\end{table}
% \vspace{-1cm}

\section{Discussion}\label{sec:dis}
\change{We propose three tests for assessing the stationarity assumption in RL, including an $\ell_1$-type test, a maximum-type test and a normalized maximum-type test. In our numerical experiments, these tests generally lead to the same conclusions. To illustrate this, we calculate the percentage of times any of the two tests produce concordant results -- either both rejecting or both not rejecting the null hypothesis -- across 100 simulation replications and visualize them in Figure \ref{fig:sim:synthetic:test_agreement} of the Supplementary Materials. The agreement rates exceed 95\% for any two tests in most scenarios. Specifically, the $\ell_1$-type and unnormalized maximum-type tests exhibit particularly high consistency, with agreement rates over 97.5\% in the majority of cases. Most inconsistencies primarily arise between the normalized test and the other two. This behavior is expected since our theoretical and empirical results suggest that the normalized test tends to achieve a larger type-I error and a larger power. As such, it is more likely to reject the null, leading to these inconsistencies.}
%almost always agree at all candidate change points. However, the normalized test sometimes yields different results than the other two. 
%Specifically, the normalized test is more likely to reject the null hypothesis when the candidate change point is close to the terminal time point (i.e., $\kappa$ is small). Conversely, it is also less likely to reject the null when $\kappa$ is larger. This behavior arises because the variance estimate used in the normalized test statistic may be subject to larger variability near the boundary. Yet with a more stable variance estimate as we move further away from the boundary, normalization increases the test sensitivity to detect deviations from stationarity. }
% {\color{blue}Need to discuss more here}. 

\change{To address such inconsistency, we further outline two approaches in Section \ref{sec:inconsistency} of the Supplementary Materials. The first approach chooses one out of the three tests, depending on the application scenarios. The second approach aggregates results from all three tests to produce a final $p$-value. This approach leverages the advantages of the three tests and outperforms each of them individually, according to our empirical study; see Figure \ref{fig:sim:synthetic:aggregation} of the Supplementary Materials for more details.}

\change{Given an offline dataset, we focus on detecting the most recent change point. That is, regardless of how many change points there are in the past, we aim to identify the most recent one. Meanwhile, our procedure can be easily adapted to identify all past change points. Specifically, as described in Section \ref{sec:changedetection},
given a pre-specified monotonically increasing sequence $\{\kappa_j\}_j \subseteq (0, T)$, our proposed test is applied to each interval $[T - \kappa_j, T]$. We define the most recent change point as $\widehat{T}^* = T - \kappa_{j_0-1}$ where the test is first rejected at $\kappa_{j_0}$. In general when there are multiple change points, we can continue applying our procedure to the interval $(0, \widehat{T}^*)$ and identify the second most recent change point $\widehat{T}_2^*$. This process can be repeated until the remaining interval does not contain other change points. When the change points are dense, we recommend to specify as many $\kappa_j$'s as possible at which the test is applied, to ensure they can be precisely identified.}

\bibliographystyle{apalike}
\bibliography{nonstationary}

\newpage
\baselineskip=22pt
\appendix

This supplement is organised as follows. We begin with a list of commonly used notations in the supplement. We next compare model-free tests against mode-based tests, and discuss some extensions in Section \ref{sec:adddis}. In Section \ref{sec:proof}, we present the proofs of our theorems. 
Finally, in Appendix \ref{sec:morenum}, we detail the simulation setting and present some additional empirical results.

\makenomenclature
\renewcommand{\nomname}{Notations and Technical Definitions}
\nomenclature[01]{\(S_t\)}{The state vector at time $t$, from a trajectory randomly selected from the population}
\nomenclature[02]{\(A_t\)}{The action taken at time $t$, from a trajectory randomly selected from the population}
\nomenclature[03]{\(R_t\)}{The immediate reward at time $t$, from a trajectory randomly selected from the population}
\nomenclature[04]{\(S_{i,t}\)}{The state vector at time $t$ in the $i$th observed data  trajectory}
\nomenclature[05]{\(A_{i,t}\)}{The action taken at time $t$ in the $i$th observed data trajectory}
\nomenclature[06]{\(R_{i,t}\)}{The immediate reward at time $t$ in the $i$th observed data trajectory}
\nomenclature[07]{\(r_t\)}{The reward function at time $t$, i.e., $r(A_t,S_t)=\mathbb{E}(R_t\mid A_t,S_t)$}
\nomenclature[08]{\(\mathcal{T}_t\)}{The state transition function at time $t$, i.e., $S_{t+1}=\mathcal{T}_t(S_t,A_t,\delta_t)$}
\nomenclature[09]{\(\gamma\)}{The discounted factor, between 0 and 1}
\nomenclature[10]{\(V_t^{\pi}\)}{The state value function under $\pi$, starting from a given state at time $t$}
\nomenclature[11]{\(Q_t^{\pi}\)}{The Q-function under $\pi$, starting from a given state-action pair at time $t$}
\nomenclature[12]{\(\pi^{opt}\)}{The optimal policy}
\nomenclature[13]{\(Q_t^{\tiny{opt}}\)}{The optimal Q-function}
\nomenclature[14]{\(\widehat{Q}_{[T_1,T_2]}\)}{The estimated optimal Q-function using data collected from the interval $[T_1,T_2]$}
\nomenclature[15]{\(\widehat{\beta}_{[T_1,T_2]}\)}{The estimated regression coefficients using data collected from the interval $[T_1,T_2]$}
\nomenclature[16]{\(\beta^*\)}{The population limit of $\widehat{\beta}_{[T_1,T_2]}$ under the null}
\nomenclature[17]{\(\phi_L(a,s)\)}{The set of sieve basis functions of length $L$}
\nomenclature[18]{$\widehat{\Sigma}_{[T_1,T_2]}$}{The random matrix $N^{-1}(T_2-T_1)^{-1}\sum_{i=1}^N\sum_{t=T_1}^{T_2-1} \phi_L(A_{i,t},S_{i,t})\phi_L^\top(A_{i,t},S_{i,t})$}
\nomenclature[19]{$\Sigma_{[T_1,T_2]}$}{The population limit of $\widehat{\Sigma}_{[T_1,T_2]}$, given by $(T_2-T_1)^{-1}\sum_{t=T_1}^{T_2-1} \Mean \phi_L(A_{t},S_{t})\phi_L^\top(A_{t},S_{t})$}
\nomenclature[20]{$\widehat{W}_{[T_1,T_2]}$}{The random matrix $$\frac{1}{N(T_2-T_1)}\sum_{i=1}^N\sum_{t=T_1}^{T_2-1} \phi_L(A_{i,t},S_{i,t})[\phi_L(A_{i,t},S_{i,t})-\gamma \phi_L(\pi_{\widehat{\beta}_{[T_1,T_2]}}(S_{i,t+1}),S_{i,t+1})]$$ where $\pi_{\beta}(s)=\argmax_s \phi_L^\top(a,s) \beta$}
\nomenclature[21]{$W_{[T_1,T_2]}$}{The population limit of $\widehat{W}_{[T_1,T_2]}$}
\nomenclature[22]{\(u\)}{A candidate change point location}
\nomenclature[23]{\(\epsilon\)}{The boundary removal parameter}
\nomenclature[24]{\(\delta_{i,t}^*\)}{The oracle temporal difference error $R_{i,t}+\gamma \max_a Q^{\tiny{opt}}(a,S_{i,t+1})-Q^{\tiny{opt}}(A_{i,t},S_{i,t})$}
\nomenclature[25]{\(\delta_{i,t}(\beta)\)}{The estimated temporal difference error $R_{i,t}+\gamma \max_a \phi_L^\top(a,S_{i,t+1})\beta-\phi_L^\top(A_{i,t},S_{i,t})\beta$ using the parameter $\beta$}
%\nomenclature[26]{\(\delta_{i,t}(\beta)\)}{The estimated temporal difference error $R_{i,t}+\gamma \max_a \phi_L^\top(a,S_{i,t+1})\beta-\phi_L^\top(A_{i,t},S_{i,t})\beta$ using the parameter $\beta$}
\nomenclature[26]{$p_t(\bullet\mid a,s)$}{The density function of $S_{t+1}$ given $(A_t,S_t)=(a,s)$}
\nomenclature[27]{\(\textrm{TS}\)}{The test statistic}
\nomenclature[28]{\(\textrm{TS}^b\)}{The bootstrap test statistic}
\nomenclature[29]{\(\Lambda(p,c)\)}{The class of $p$-smoothness functions, defined below} 
\begin{singlespace}
\printnomenclature
\end{singlespace}

%\section{Some Technical Definitions}\label{sec:def}
For a $J$-tuple $\alpha=(\alpha_1,\dots,\alpha_J)^{\top}$ of nonnegative integers and a given function $h$ on $\mathcal{S}$, let $D^{\alpha}$ denote the differential operator:
\begin{eqnarray*}
	D^{\alpha}h(s)=\frac{\partial^{\|\alpha\|_1} h(s)}{\partial s_1^{\alpha_1}\cdots\partial s_J^{\alpha_J}}.
\end{eqnarray*}
Here, $s_j$ denotes the $j$th element of $s$. For any $p>0$, let $\floor{p}$ denote the largest integer that is smaller than $p$. The class of $p$-smooth functions is defined as follows:
\begin{eqnarray*}
	\Lambda(p,c)=\left\{h:\sup_{\|\alpha\|_1\le \floor{p}} \sup_{s\in \mathcal{S}} |D^{\alpha} h(s)|\le c, \sup_{\|\alpha\|_1=\floor{p}} \sup_{\substack{s_1,s_2\in \mathcal{S}\\ s_1\neq s_2}} \frac{|D^{\alpha} h(s_1)-D^{\alpha} h(s_2)|}{\|s_1-s_2\|_2^{p-\floor{p}}}\le c \right\},
\end{eqnarray*}
for some constant $c>0$. 

\section{Additional Discussions}\label{sec:adddis}
\subsection{Model-free Tests vs. Model-based Tests}\label{sec:modelbased}
As commented in the main text, we focus on model-free tests in this paper, constructing the test statistic without directly estimating the reward and transition functions. Alternatively, one may study model-based tests \citep[see e.g.,][]{Alegre2021Minimum,wang2023robust}, which directly evaluate the stationarity of the MDP model (i.e., reward and transition functions). 
Both model-based and model-free methods offer distinct advantages in RL. For example, model-free RL focuses on learning a one-dimensional optimal Q-function, eliminating the need to model a complex transition function that has an output dimension equal to the state's. This is particularly beneficial given the challenges in modeling transition functions, which can be prone to misspecification, %especially in high-dimensional state spaces. 
as demonstrated in our simulation studies. 
More specifically, consider the case where the future state follows a conditional Gaussian distribution. Specifying the $d$-dimensional mean and $d\times d$-dimensional covariance functions of the state-action pair can be quite complex. Conversely, the estimation the Q-function is challenging. A limited number of visits to certain state-action pairs can lead to inaccurate estimations of the Q-function's value across other pairs.

\subsection{Extensions to Settings with Varying Termination Times}\label{sec:excvary}
Our proposed method is readily adaptable to scenarios where subjects have varying termination times. To elaborate, let \(T^{(i)}\) represent the termination time for the \(i\)th subject, and \(T = \max_i T^{(i)}\) be the maximum of these times. In estimating \(\widehat{\beta}_{[T_1,T_2]}\), we can modify the empirical sum operator \(\sum_{i=1}^N \sum_{t=T_1}^{T_2-1}\) to \(\sum_{i: 1\le i\le N, T^{(i)}>T_1} \sum_{t=T_1}^{T^{(i)}-1}\). Similarly, for the construction of the \(\ell_1\)-type test, the empirical average operator \(1/(N(T-T_0)) \sum_{i=1}^N \sum_{t=T_0}^{T-1}\) can be substituted with \(n^{-1}(T_0) \sum_{i: 1\le i\le N, T^{(i)}>T_0} \sum_{t=T_0}^{T^{(i)}-1}\), where \(n(T_0)\) denotes the total count of observations in the interval between \(T_0\) and \(T\). 

\subsection{Inconsistency, Comparison and Aggregation of the Three Tests}\label{sec:inconsistency}
\begin{figure}[t]
    \centering
    \includegraphics[width=\linewidth]{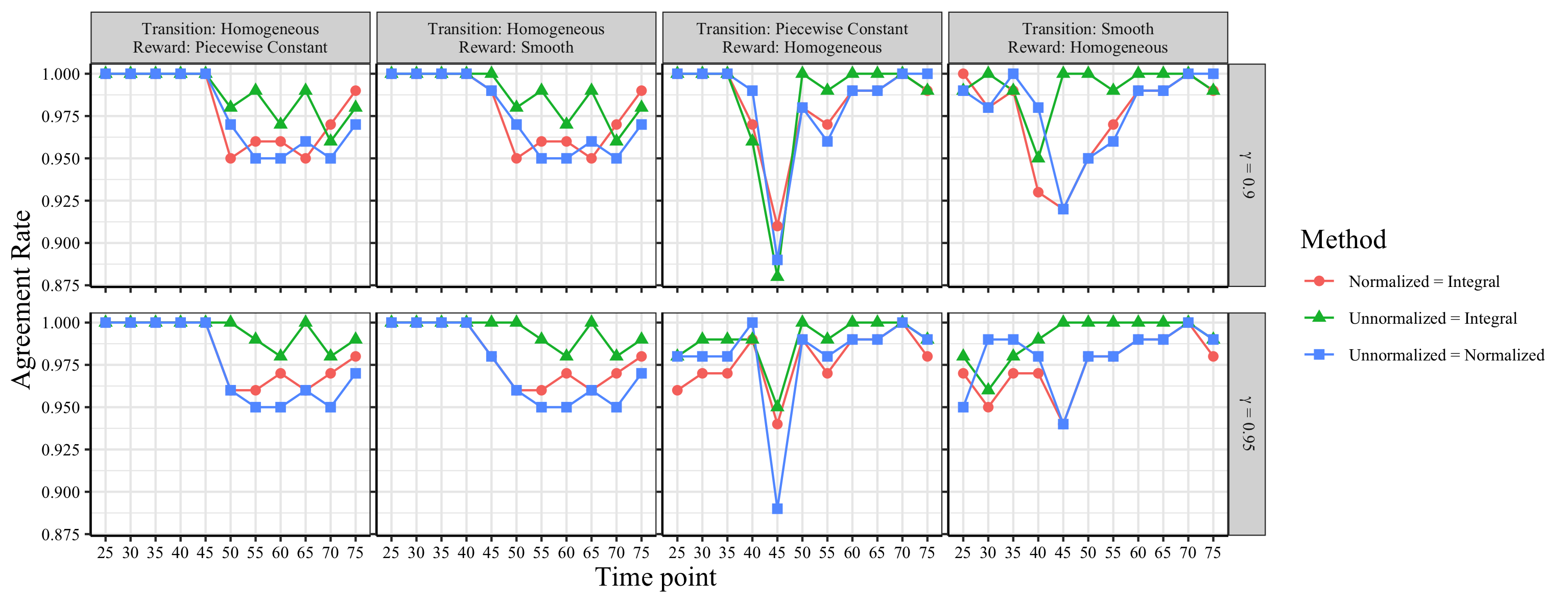}
    \caption{Agreements rates of the three tests under settings in Section \ref{sec:sim:1d} with $N = 25$.
    }
    \label{fig:sim:synthetic:test_agreement}
\end{figure}
For each time point that is a candidate change point, we calculate the percentage of times any of the two tests produce consistent results -- both rejecting or both not rejecting the null hypothesis -- out of 100 replications and visualize the results in Figure \ref{fig:sim:synthetic:test_agreement}. All three tests are agree in most cases. In particular, the unnormalized and $\ell_1$-type tests have the highest agreement rates in most scenarios. The normalized test has a relatively lower agreement rates, since it is more likely to reject the null hypothesis compared to the other two.

To address such inconsistency, we outline two approaches below. The first approach chooses one out of the three tests, depending on the application scenarios. In scenarios where our goal lies in identifying meaningful change points in scientific discoveries, the $\ell_1$-type or unnormalized maximum-type test is preferable for lower susceptibility to type-I errors (see Remark \ref{remark:compare3tests}). In biomedical applications like mobile health, type-I errors are more detrimental than type-II errors. 
% This is because to gain clinical insights with batch data collected in biomedical applications such as mobile health, type-I errors are more detrimental than type-II errors. 
In scenarios where our goal is to train an optimal policy that maximizes the cumulative reward, the normalized test will be preferred. Overlooking a potential change point (type-II error) may lead to a sub-optimal policy and can be more critical than incorrectly identifying a non-change point as a change (type-I error). 
% Ignoring a change point results in usage of nonstationary data for learning, leading to a sub-optimal policy. 
In contrast, incorrectly identifying a stationary time period still uses stationary data, leading to consistent, though inefficient policy learning. 

\begin{figure}[t]
    \centering
    \begin{subfigure}{0.9\textwidth}
        \includegraphics[width=\linewidth]{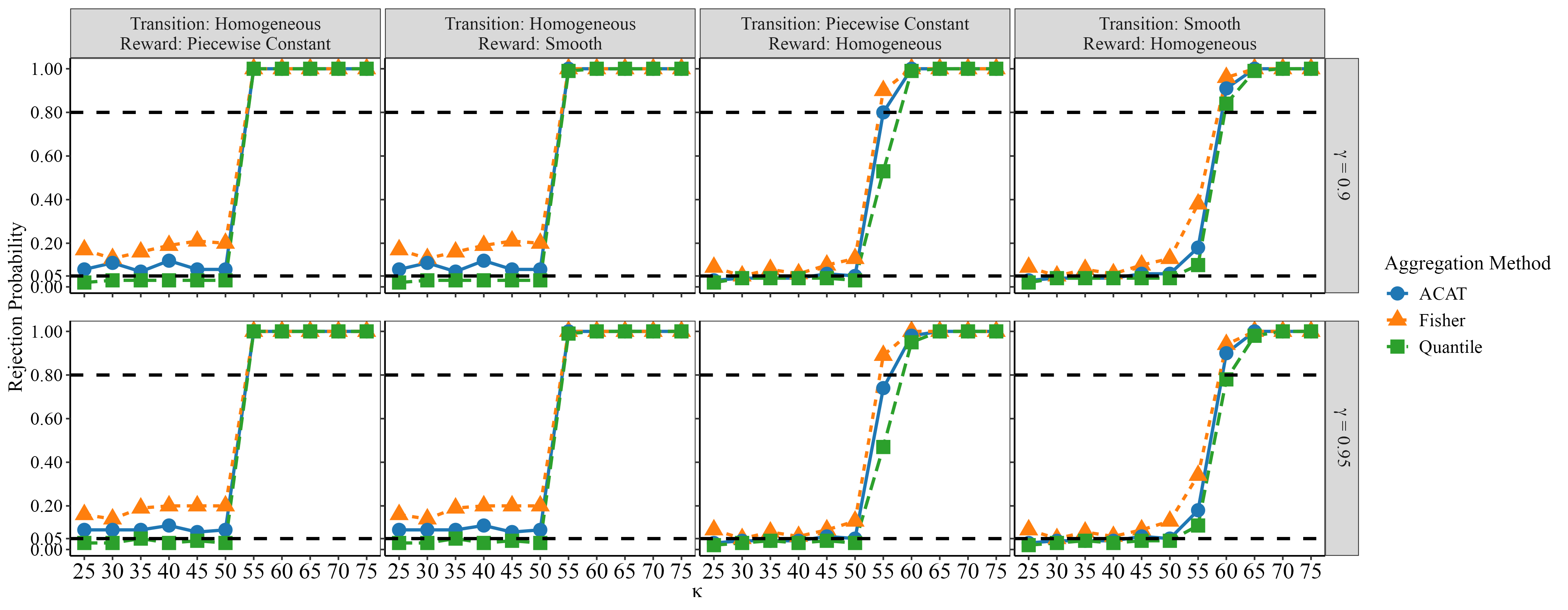}        \subcaption{Rejection probabilities under $N = 25$.}
        % \label{fig:arm1}
    \end{subfigure}

    \medskip
    \begin{subfigure}{0.9\textwidth}
        \includegraphics[width=\linewidth]{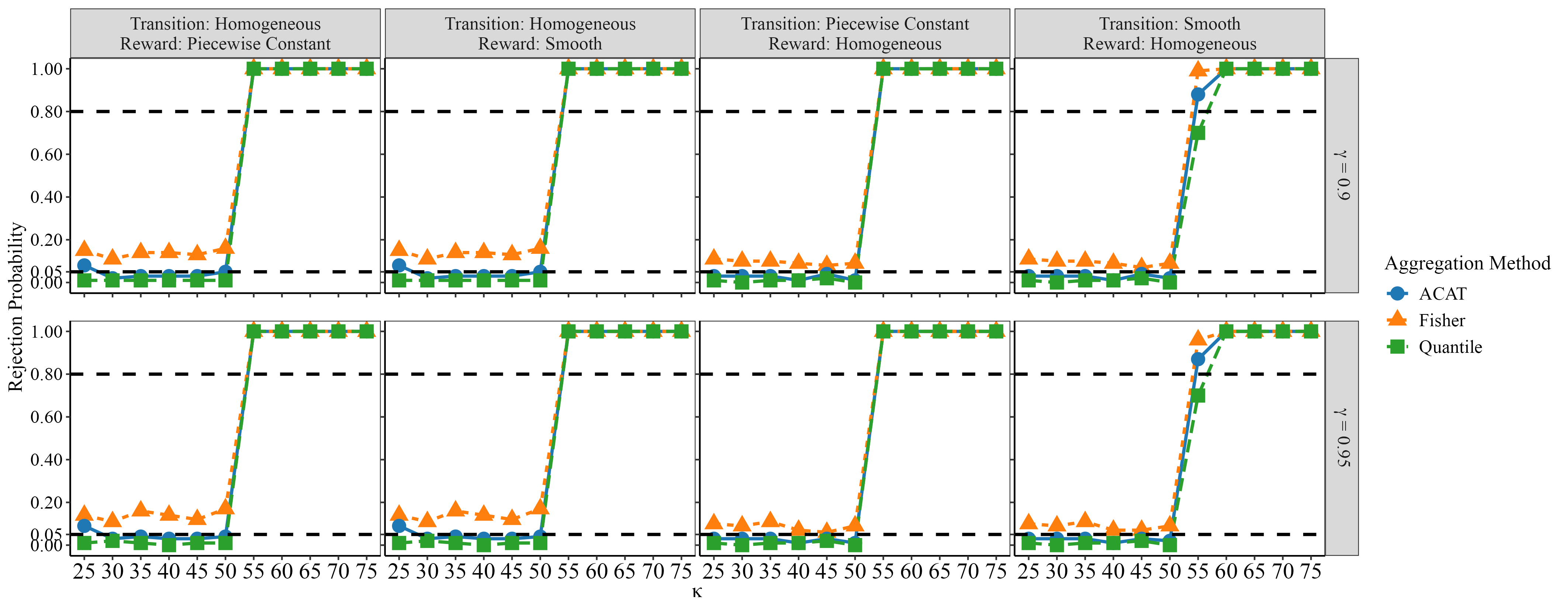}        \subcaption{Rejection probabilities under $N = 100$.}
    \end{subfigure}
    \caption{Test results by aggregating p-values from the three tests using Fisher's method, the Cauchy combination method, and quantile-based method (with 0.1 quantile), under simulation settings in Section \ref{sec:sim:1d} with $N = 25$ and $100$. True change point is 50 in all scenarios.}
    \label{fig:sim:synthetic:aggregation}
\end{figure}

The second approach aggregates results from all three tests to produce a final $p$-value. This approach leverages the advantages of the three tests and is expected to outperform each of them individually. 
Various $p$-value aggregation methods are applicable, including the Fisher's method \citep{fisher1928statistical}, the quantile-based method \citep{meinshausen2009p} and the Cauchy combination method \citep[ACAT; ][]{Liu2020cauchy}. We implement these three methods under settings in Section \ref{sec:sim:1d} and report the results in Figure \ref{fig:sim:synthetic:aggregation}. It can be seen that while the Fisher's method is subject to inflated type-I errors, both the quantile method and ACAT asymptotically control type-I errors and achieve comparable or better power to that of the individual tests in all scenarios. Notably, in the last two settings with nonstationary transition functions, ACAT and the quantile methods are more powerful when the change point is near the endpoint of the interval being tested (i.e., $\kappa$ is close to 50). These results empirically verify the benefits of combining the three tests. 
%are the more appropriate for aggregating the three tests. 
%As shown in Figure \ref{fig:sim:synthetic:aggregation}(a), the quantile-based method is best in controlling type I errors, while Fisher's method is subject to inflated type I errors in all scenarios. ACAT controls type I errors properly except when $\kappa = 25$ with a small testing interval. On the other hand, Fisher's method produces the largest power and quantile method produces the smallest power, particularly in the most difficult scenario with a smooth state transition and a homogeneous reward function. Overall, 

\begin{table}[b]
\centering
\caption{Mean (SD) computation time ($\times 10^{-3}$ seconds) for three test statistics in synthetic data simulation across 10 replicates} 
\label{tab:sim:1d:computation_time}
\small
\begin{tabular}{rlll}
$\kappa$ & Integral & Normalized & Unnormalized \\ 
  \midrule
 25 & 35.42 (15.19) & 51.7 (5.35) & 35.16 (5.11) \\ 
   50 & 54.2 (11.21) & 68.54 (7.73) & 49.46 (6.77) \\ 
   75 & 72.17 (10.26) & 73.02 (9.43) & 52.1 (6.61) \\ 
   \bottomrule
\end{tabular}
\end{table}
Finally, as commented in Remark \ref{remark:compare3tests} of the main paper, the normalized test statistics requires to compute the estimated variance $\widehat{\sigma}_u^2$, which slightly increases the computational time. Here we investigate the time required to compute each test statistic in Scenario 4 of the synthetic data simulation, with smooth transition function and homogeneous reward function. Specifically, we focus on the test when $\kappa = 25, 50$, and 75 with sample size $N = 100$. Table \ref{tab:sim:1d:computation_time} reports the average time used to calculate the three test statistics over 10 replications. The experiments were conducted on Apple M4 Pro chip with 24GB memory. 
It can be seen that the normalized test is indeed the most computationally expensive among the three tests, even though all three tests are computed very fast.

\section{Proofs}\label{sec:proof}
Throughout the proof, we use $c$, $\bar{c}$, $C$, $\bar{C}$ to denote some generic constants whose values are allowed to vary from place to place. Additionally, the discount factor $\gamma$ is set to a fixed constant that is less than 1. This is to simplify our finite-sample error bounds by omitting some higher-order remainder terms, which are more heavily dependent on $(1-\gamma)^{-1}$. However, for the leading remainder terms, we will explicitly allow them to depend on $(1-\gamma)^{-1}$ to illustrate their dependence with the horizon. Finally, recall that for any two positive sequences $\{a_{N,T}\}_{N,T},\{b_{N,T}\}_{N,T}$, the notation $a_{N,T} \preceq b_{N,T}$ means that there exists some constant $C>0$ such that $a_{N,T}\le C b_{N,T}$ for any $N$ and $T$.
\subsection{Proof of Theorem \ref{thm1}}\label{sec:proof:thm1}
The derivation of SA2 from SA1 is straightforward, given the definition of $Q_t^{\pi}$. Similarly, the derivation from SA3 to SA4 is straightforward, as it directly follows from Equation \eqref{eqn:optimalpolicy}. Therefore, we aim to show SA3 under SA2 in the rest of the proof. 

Without loss of generality, assume $T_0=0$. The main idea of our proof lies in the use of policy iteration \citep{Sutton2018} to establish the connection between a (non-optimal) Q-function and its optimal counterpart. We initiate this process with an arbitrarily chosen stationary policy $\pi_1$. Subsequently, we define $\pi_2$ as the greedy policy with respect to $Q_t^{\pi_1}$, i.e., 
\begin{eqnarray*}
    \pi_{2,t}(a|s)=\left\{
    \begin{array}{ll}
        1, & \textrm{if~}a=\argmax_{a'} Q_t^{\pi_1}(a',s); \\
        0, & \textrm{otherwise}. 
    \end{array} 
    \right.
\end{eqnarray*}
When the argmax is non-unique, we select the smallest maximizer. Under SA2, $Q_t^{\pi_1}$ is stationary, so is $\pi_2$. We next repeat this process by defining $\pi_{k}$ to the greedy policy with respect to $Q_t^{\pi_{k-1}}$ for $k=3,4,\cdots$. Similarly, we can show that all these policies and Q-functions are stationary. To ease the notation, we remove the subscript $t$ from $\pi_{k,t}$ and $Q_t^{\pi_{k-1}}$. Thus, they are represented as $\pi_k$ and $Q^{\pi_{k-1}}$, respectively. 

According to the policy improvement theorem \citep[see e.g.,][Equation 4.8]{Sutton2018}, we obtain $V_t^{\pi_k}(s)\ge V_{t}^{\pi_{k-1}}(s)$ for any $k$, $t$ and $s$. Here, $V_t^{\pi}(\bullet)$ denotes the (state) value function $\Mean^{\pi} (\sum_{k\ge t} \gamma^{k-t} R_k|S_t=s)$ starting from a given state $s$ at time $t$. Additionally, the Q-function is connected to the value function through the following equation: 
\begin{align}\label{eqn:QVrelation}
    Q_t^{\pi}(a,s)=\Mean [R_t+\gamma V_{t+1}^{\pi}(S_{t+1})|A_t=a,S_t=s].
\end{align}
As such, we obtain that $Q^{\pi_k}(a,s)\ge Q^{\pi_{k-1}}(a,s)$ for any $k$,  $a$ and $s$. For a given state-action pair, the sequence $\{Q^{\pi_k}(a,s)\}_k$ is monotonically non-decreasing. Since all rewards are uniformly bounded, so are these Q-function. Consequently, $Q^{\pi_k}$ converges to a bounded function $Q^*$. The convergence is uniform given that the state-action space is finite. 

To complete the proof, we aim to show that $Q^*=Q^{opt}_t$ for any $t$. According to Theorem 6.2.10 of \cite{puterman2014markov}, $\pi^{opt}$ maximizes the value function among all policies, i.e., $V_{t}^{\pi_{opt}}(s)\ge V_t^{\pi}(s)$ for any $t,s$ and $\pi$. This together with \eqref{eqn:QVrelation} yields 
\begin{align}\label{eqn:somerelation}
    Q^{\pi_k}\le Q^{opt}_t,    
\end{align}
for any $k,t$. Additionally, it follows from \eqref{eqn:QVrelation} that
\begin{align*}
    Q^{\pi_k}(a,s)=\Mean [R_t+\gamma Q^{\pi_k}(\pi_k(S_{t+1}),S_{t+1})|A_t=a,S_t=s].
\end{align*}
Since $Q^{\pi_k}$ is monotonically non-decreasing, we obtain
\begin{align*}
    Q^{\pi_k}(a,s)\ge \Mean [R_t+\gamma Q^{\pi_{k-1}}(\pi_k(S_{t+1}),S_{t+1})|A_t=a,S_t=s]\\
    =\Mean [R_t+\gamma \max_{a'} Q^{\pi_{k-1}}(a',S_{t+1})|A_t=a,S_t=s].
\end{align*}
This together with the Bellman optimality equation yields, 
\begin{align*}
    Q^{\pi_k}(a,s)-Q^{opt}_t(a,s)\ge \gamma \Mean [\max_{a'} Q^{\pi_{k-1}}(a',S_{t+1})-\max_{a'}Q_{t+1}^{opt}(a',S_{t+1})|A_t=a,S_t=s].
\end{align*}
In view of \eqref{eqn:somerelation}, we obtain
\begin{align*}
    |Q^{\pi_k}(a,s)-Q^{opt}_t(a,s)|\le \gamma \Mean [|\max_{a'} Q^{\pi_{k-1}}(a',S_{t+1})-\max_{a'}Q_{t+1}^{opt}(a',S_{t+1})||A_t=a,S_t=s]\\
    \le \gamma \max_{a',s'} |Q^{\pi_{k-1}}(a',s')-Q_{t+1}^{opt}(a',s')|,
\end{align*}
and hence 
\begin{align*}
    \max_{a,s} |Q^{\pi_k}(a,s)-Q^{opt}_t(a,s)|\le \gamma \max_{a,s} |Q^{\pi_{k-1}}(a,s)-Q_{t+1}^{opt}(a,s)|.
\end{align*}
Notice that $Q^{\pi_k}$ converges uniformly to $Q^*$. By letting $k\to \infty$, we obtain
\begin{align*}
    \max_{a,s} |Q^*(a,s)-Q^{opt}_t(a,s)|\le \gamma \max_{a,s} |Q^*(a,s)-Q_{t+1}^{opt}(a,s)|\le \gamma^K \max_{a,s} |Q^*(a,s)-Q_{t+K}^{opt}(a,s)|,
\end{align*}
for any $K>1$. Under the bounded reward assumption, by letting $K\to \infty$, we obtain $Q^*=Q^{opt}_t$ for any $t$. This yields SA3. The proof is hence completed. 

\subsection{Proof of Theorem \ref{thm:size}}\label{sec:proofthmsize}
We begin by introducing the following auxiliary lemmas. Specifically, Lemma \ref{lemmabasis} lists the properties of B-spline basis function. Lemma \ref{lemmaQ} derives the uniform rate of convergence of $\{\widehat{Q}_{[T_1,T_2]}\}_{T_1,T_2}$ and obtain their asymptotic linear representations. It in turn implies the asymptotic normality of these estimated Q-functions.   
Lemma \ref{lemmamatrixnonstat} provides a uniform upper error bound on $\|\widehat{W}_{[T_1,T_2]}-W_{[T_1,T_2]}\|_2$. Without loss of generality, assume $T_0=0$. Their proofs are provided in Sections \ref{sec:prooflemmaB1} and \ref{sec:prooflemma2}, respectively. 
\begin{lemma}\label{lemmabasis}
    Under the null hypothesis defined in \eqref{eqn:H0} that $Q_t^{opt}=Q^{opt}$ for all $ t\ge T_0$, there exists some $\beta^*\in \mathbb{R}^L$ such that
    \begin{eqnarray}\label{eqn:basis1}
        \sup_{a,s} |Q^{opt}(a,s)-\phi_L^\top(a,s) \beta^*|=O\Big(\frac{L^{-p/d}}{(1-\gamma)^2}\Big).
    \end{eqnarray}
    Additionally, we have
    \begin{eqnarray}\label{eqn:basis2}
        \max_a \lambda_{\max}\Big[\int_s \phi_L(a,s) \phi^\top(a,s) ds\Big]=O(1),\\\label{eqn:basis2.5}
        1\le \|\phi_L(a,s)\|_2\le \|\phi_L(a,s)\|_1=\sqrt{L},\quad \forall a,s,
    \end{eqnarray}
    and
    \begin{eqnarray}\label{eqn:basis3}
        \max_a \sup_{s_1\neq s_2} \frac{\|\phi_L(a,s_1)-\phi_L(a,s_2)\|_2}{\|s_1-s_2\|_2}=O(\sqrt{L}).
    \end{eqnarray}
\end{lemma}

\begin{lemma}\label{lemmaQ}
Under the null hypothesis, the set of Q-estimators $\{\phi_L^\top(a,s) \widehat{\beta}_{[T_1,T_2]}\}$ are uniformly consistent in $\ell_{\infty}$ norm and satisfy the following uniform rate of convergence 
\begin{eqnarray}\label{eqn:rateofconvergence}
    \max_{\substack{[T_1,T_2]\subseteq [T_0,T]\\ T_2-T_1>\epsilon T}}\sup_{a,s} |\phi_L^\top(a,s) \widehat{\beta}_{[T_1,T_2]}-Q^{opt}(a,s)|=O\Big(\frac{L^{-p/d}}{(1-\gamma)^2}\Big)+O\Big(\frac{\sqrt{L\log (NT)}}{(1-\gamma)^2\sqrt{\epsilon  NT}}\Big),
\end{eqnarray}
with probability at least $1-O(N^{-1}T^{-1})$. 
 %there exists some constant $c>0$ such that
Additionally, 
\begin{eqnarray}\label{eqn:asymptoticexpansion}
\begin{split}
    \phi_L^\top(a,s)\widehat{\beta}_{[T_1,T_2]}-Q^{\tiny{opt}}(a,s)=\frac{\phi_L^\top(a,s)}{N(T_2-T_1)}W_{[T_1,T_2]}^{-1}\sum_{i=1}^N\sum_{t=T_1}^{T_2-1}\phi_L(A_{i,t},S_{i,t})\delta_{i,t}^*\\
    +O\Big(\frac{L^{-p/d}}{(1-\gamma)^3}\Big)+O\Big(\frac{L^{3/2}\log (NT)}{(1-\gamma)^3\epsilon  NT}\Big),
\end{split}
\end{eqnarray}
where the big-$O$ terms hold uniformly for any pair $(T_1,T_2)$ such that $T_2-T_1\ge \epsilon T$ with probability at least $1-O(N^{-1}T^{-1})$, and we recall that $\delta_{i,t}^*$ denotes the temporal difference error $R_{i,t}+\gamma \max_a Q^{\tiny{opt}}(a,S_{i,t+1})-Q^{\tiny{opt}}(A_{i,t},S_{i,t})$. 
	%for some $\kappa>1/2$ 
    %WPA1. %where $\beta^*$ corresponds to the least-false parameter defined in Lemma \ref{lemmabasis}. Here, the exponent $\kappa$ can be set to $\min(p/d, 0.5c_4^{-1}-0.5-\epsilon^*)$ for any sufficiently small $\epsilon^*>0$. 
\end{lemma}

\begin{lemma}\label{lemmamatrixnonstat}
Under the null hypothesis, there exists some constant $\bar{c}>0$ such that $\|W_{[T_1,T_2]}^{-1}\|_2\le \bar{c}(1-\gamma)^{-1}$ and that $\max_{T_2-T_1\ge \epsilon T} \|\widehat{W}_{[T_1,T_2]}-W_{[T_1,T_2]}\|=O\{(\epsilon NT)^{-1/2}  \sqrt{L\log (NT)}\}$ with probability at least $1-O(N^{-1} T^{-1})$. Here, $\|W_{[T_1,T_2]}^{-1}\|_2$ corresponds to the matrix operator norm of $W_{[T_1,T_2]}^{-1}$.
\end{lemma}

\subsubsection{Unnormalized Maximum-type Tests}\label{sec:proofunmax}
We first provide an outline of the proof. Define $\textrm{TS}_{\infty}^{*}$ to be a version of $\textrm{TS}_{\infty}$ with $\widehat{Q}_{[T_1,T_2]}(a,s)$ replaced by the leading terms in the asymptotic expansion \eqref{eqn:asymptoticexpansion}, i.e., 
\begin{eqnarray*}
    \frac{1}{N(T_2-T_1)}\sum_{i=1}^N \sum_{t=T_1}^{T_2-1} \phi_L^\top(a,s)W_{[T_1,T_2]}^{-1}\phi_L(A_{i,t},S_{i,t})\delta_{i,t}^*.
\end{eqnarray*}
It follows from \eqref{eqn:asymptoticexpansion} that
\begin{eqnarray}\label{eqn:step0}
\begin{aligned}
\prob\Big(|\textrm{TS}_{\infty}-\textrm{TS}_{\infty}^{*}|\le \frac{C L^{-p/d}}{(1-\gamma)^3}+\frac{C L^{3/2} \log (NT)}{(1-\gamma)^3\epsilon  NT} \Big) =1-O\Big(\frac{1}{NT}\Big),
\end{aligned}
\end{eqnarray}
for some constant $C>0$.

The rest of the proof is divided into three steps. In the first step, we establish a uniform upper error bound for $\max_{i,t,T_1,T_2} |\widehat{Q}_{[T_1,T_2]}^{b}(A_{i,t},S_{i,t})-\widehat{Q}_{[T_1,T_2]}^{b,0}(A_{i,t},S_{i,t})|$, where \begin{eqnarray*}
	\widehat{Q}^{b,0}_{[T_1,T_2]}(a,s)=\frac{1}{N(T_2-T_1)}\phi_{L}^\top(a,s)W_{[T_1,T_2]}^{-1} \sum_{i=1}^N \sum_{t=T_1}^{T_2-1} \phi_{L}(A_{i,t},S_{i,t})\delta_{i,t}^* e_{i,t},\,\,\,\,\forall T_1,T_2,
\end{eqnarray*}
a version of $\widehat{Q}^{b}_{[T_1,T_2]}(a,s)$ with $\widehat{W}_{[T_1,T_2]}^{-1}$ and $\delta_{i,t}(\widehat{\beta}_{[T_1,T_2]})$ replaced by their oracle values. 
This in turn leads to the following bound for the difference between $\textrm{TS}_{\infty}^{b}$ and $\textrm{TS}_{\infty}^{b,*}$,
\begin{eqnarray}\label{eqn:step1}
\prob\Big(|\textrm{TS}_{\infty}^{b}-\textrm{TS}_{\infty}^{b,*}|\le \frac{C L^{3/2}\log(NT)}{(1-\gamma)^3\epsilon NT}+\frac{CL^{-p/d}\sqrt{\log(NT)}}{(1-\gamma)^3\sqrt{\epsilon NT}}\Big)= 1-O\Big(\frac{1}{NT}\Big),
\end{eqnarray}
for some constant $C>0$, where
\begin{eqnarray*}
	\textrm{TS}_\infty^{b,*}=\max_{\epsilon T<u<(1-\epsilon) T} \max_{a,s} \sqrt{\frac{u(T-u)}{T^2}}|\widehat{Q}_{[0,u]}^{b,0}(a,s)-\widehat{Q}_{[u,T]}^{b,0}(a,s)|.
\end{eqnarray*}

Notice that in the test statistic, the maximum is taken over all state-action pairs. Recall that the state space is $[0,1]^d$. In the second step, we discretize the state space by considering an $\varepsilon$-net of $[0,1]^d$ -- denoted by $\mathcal{S}_{\varepsilon}$ -- with $\varepsilon=\sqrt{d}/(NT)^4$ so that for any $s\in [0,1]^d$, there exists some $s'$ in the $\varepsilon$-net such that $\|s-s'\|_2\le \varepsilon$. Let $\textrm{TS}_{\infty}^{**}$ and $\textrm{TS}_{\infty}^{b,**}$
be versions of $\textrm{TS}_{\infty}^{*}$ and $\textrm{TS}_{\infty}^{b,*}$ where the maximum is taken over the $\varepsilon$-net. The focus of this step is to upper bound $|\textrm{TS}_{\infty}^*-\textrm{TS}_{\infty}^{**}|$ and $|\textrm{TS}_{\infty}^{b,*}-\textrm{TS}_{\infty}^{b,**}|$. Due to the choice of $\varepsilon$, these bounds are minimal. Together with \eqref{eqn:step0} and \eqref{eqn:step1}, we can show that
\begin{eqnarray}\label{eqn:step3initial}
\begin{aligned}
&\prob\Big(|\textrm{TS}_{\infty}-\textrm{TS}_{\infty}^{**}|\le \frac{c L^{-p/d}}{(1-\gamma)^3}+\frac{c L^{3/2} \log (NT)}{(1-\gamma)^3\epsilon  NT} \Big)= 1-O\Big(\frac{1}{NT}\Big),\\
&\prob\Big(|\textrm{TS}_{\infty}^{b}-\textrm{TS}_{\infty}^{b,**}|\le \frac{c L^{-p/d}}{(1-\gamma)^3}+\frac{c L^{3/2} \log (NT)}{(1-\gamma)^3\epsilon  NT} \Big)= 1-O\Big(\frac{1}{NT}\Big),
\end{aligned}
\end{eqnarray}

In the last step, we aim to show the proposed test controls the type-I error. 
A key step in our proof is to bound the Kolmogorov distance between $\textrm{TS}_{\infty}^{**}$ and $\textrm{TS}_{\infty}^{b,**}$, which together with \eqref{eqn:step3initial} yields the validity of the proposed test. Notice that $\textrm{TS}_{\infty}^{**}$ can be viewed as the maximum of a set of mean zero random vectors
\begin{eqnarray}\label{eqn:maxrandomvectors}
\begin{split}
	\Big\{Z_{u,a,s,j}\stackrel{\Delta}{=}(-1)^j\tau_u\phi_L^\top(a,s)\Big[\frac{W_{[0,u]}^{-1} }{Nu}\sum_{i=1}^N \sum_{t=0}^{u-1} \phi_{L}(A_{i,t},S_{i,t})\delta_{i,t}^*-\frac{W_{[u,T]}^{-1}}{N(T-u)}\\ \times \sum_{i=1}^N \sum_{t=u}^{T-1} \phi_{L}(A_{i,t},S_{i,t})\delta_{i,t}^*\Big]: j\in \{0,1\},\epsilon T<u<(1-\epsilon) T,a\in \mathcal{A},s\in \mathcal{S}_{\varepsilon} \Big\},
\end{split}
\end{eqnarray}
Similarly, $\textrm{TS}_\infty^{b,*}$ can be represented as a function of the bootstrapped samples
\begin{eqnarray}\label{eqn:maxrandomvectors0}
\begin{split}
	\Big\{Z_{u,a,s,j}^b\stackrel{\Delta}{=}(-1)^j\tau_u \Big[\frac{W_{[0,u]}^{-1} }{Nu} \sum_{i=1}^N \sum_{t=0}^{u-1} \phi_{L}(A_{i,t},S_{i,t})\delta_{i,t}^*e_{i,t}-\frac{W_{[u,T]}^{-1}}{N(T-u)} \sum_{i=1}^N \sum_{t=u}^{T-1} \phi_{L}(A_{i,t},S_{i,t})
    \\\times \delta_{i,t}^*e_{i,t}\Big]: j\in \{0,1\},\epsilon T<u<(1-\epsilon) T,a\in \mathcal{A},s\in \mathcal{S}_{\varepsilon} \Big\}.
\end{split}
\end{eqnarray}
When $T$ and $L$ are fixed, the classical continuous mapping theorem can be applied to establish the weak convergence results. However, in our setting, $L$ needs to diverge with the number of observations to alleviate the model misspecification error. We also allow $T$ to approach infinity.
Hence, classical weak convergence results cannot be applied. Toward that end, we establish a nonasymptotic error bound for the Kolmogorov distance as a function of $N,T$ and $L$, and show that this bound decays to zero under the given conditions. The proof is based on the high-dimensional martingale central limit theorem (CLT) developed by \cite{belloni2018high}; see also the high-dimensional CLT by \cite{cherno2014}.

We next detail the proof for each step. 

\noindent \textbf{Step 1}. By definition, $\widehat{Q}_{[T_1,T_2]}^{b,0}(a,s)-\widehat{Q}_{[T_1,T_2]}^{b}(a,s)$ is equal to the sum of
\begin{eqnarray}\label{eqn:anothertwoterms}
%\begin{split}
	\frac{1}{N(T_2-T_1)}\phi_{L}^\top(a,s)(\widehat{W}_{[T_1,T_2]}^{-1}-W_{[T_1,T_2]}^{-1}) \sum_{i=1}^N \sum_{t=T_1}^{T_2-1} \phi_{L}(A_{i,t},S_{i,t})\delta_{i,t}(\widehat{\beta}_{[T_1,T_2]}) e_{i,t}	
\end{eqnarray}
and
\begin{eqnarray}\label{eqn:anothertwoterms2}
    \frac{1}{N(T_2-T_1)}\phi_{L}^\top(a,s)W_{[T_1,T_2]}^{-1} \sum_{i=1}^N \sum_{t=T_1}^{T_2-1} \phi_{L}(A_{i,t},S_{i,t})(\delta_{i,t}(\widehat{\beta}_{[T_1,T_2]})-\delta_{i,t}^*) e_{i,t}.
\end{eqnarray}
Consider the first term. In Lemma \ref{lemmamatrixnonstat}, we establish a uniform upper error bound for $\|\widehat{W}_{[T_1,T_2]}-W_{[T_1,T_2]}\|_2$ and show that $(1-\gamma)\|W_{[T_1,T_2]}^{-1}\|_2$ is upper bounded by some constant. Using similar arguments in Part 3 of the proof of Lemma 3 in \cite{shi2022statistical}, we can show that $\|\widehat{W}_{[T_1,T_2]}^{-1}-W_{[T_1,T_2]}^{-1}\|_2$ is of the same order of magnitude to $(1-\gamma)^{-2}\|\widehat{W}_{[T_1,T_2]}-W_{[T_1,T_2]}\|_2$. The boundedness of rewards (implied by A2) implies that the Q-function is bounded by $O((1-\gamma)^{-1})$. This together with Lemma \ref{lemmaQ} implies the estimated Q-function is bounded by $O((1-\gamma)^{-1})$ as well, and so is $\delta_{i',t'}(\widehat{\beta}_{[T_1,T_2]})$. By \eqref{eqn:basis2}, the conditional variance of \eqref{eqn:anothertwoterms} given the data is upper bounded by
\begin{eqnarray*}
	\frac{C L^2\log(NT)}{\epsilon^2 N^2T^2(1-\gamma)^6} \lambda_{\max}\left\{\frac{1}{N(T_2-T_1)}\sum_{i=1}^N \sum_{t=T_1}^{T_2-1} \phi_L(A_{i,t},S_{i,t})\phi_L^\top(A_{i,t},S_{i,t}) \right\}.
\end{eqnarray*}
Similar to Lemma \ref{lemmamatrixnonstat}, we can show that the maximum eigenvalue of the matrix inside the curly brackets converges to $\lambda_{\max} \{(T_2-T_1)^{-1} \sum_{t=T_1}^{T_2-1}  \Mean\phi_L(A_{t},S_{t})\phi_L^\top(A_{t},S_{t})\}$, which is bounded by some finite constant according to \eqref{eqn:basis2} and the boundedness of the transition function $p_t$s in Condition A4(iii). Hence, the conditional variance of \eqref{eqn:anothertwoterms} given the data is of the order $O(L^2\epsilon^{-2}N^{-2}T^{-2}(1-\gamma)^{-6}\log(NT))$, with probability at least $1-O(N^{-1}T^{-1})$. 

Notice that the probability of a standard normal random variable exceeding $z$ is bounded by $\exp(-z^2/2)$ for any $z>1$; see e.g., the inequality for the Gaussian Mill's ratio \citep{birnbaum1942inequality}.
%\begin{eqnarray*}
%	\prob(N(0,1)> z)=\int_{z}^{+\infty} \frac{1}{\sqrt{2\pi}}\exp(-y^2/2)dy\le \int_{z}^{+\infty} y\exp(-y^2/2)dy=\exp(-z^2/2),
%\end{eqnarray*}
%for any $z>1$. 
Since \eqref{eqn:anothertwoterms} is a mean-zero Gaussian random variable given the data, it is of the order
\begin{eqnarray*}
    O\Big(\frac{L\log(NT)}{(1-\gamma)^3\epsilon NT}\Big),
\end{eqnarray*}
%$O\{L (1-\gamma)^{-1}(\epsilon NT)^{-1}\sqrt{\log (NT)}\}$, 
with probability at least $1-O\{(NT)^{-C}\}$ for any sufficiently large constant $C>0$. This together with Bonferroni's inequality yields the desired uniform upper error bound for the first term. 

As for the second term, notice that according to the finite-sample error bound established in Lemma \ref{lemmaQ}, the difference $\delta_{i,t}(\widehat{\beta}_{[T_1,T_2]})-\delta_{i,t}^*$ decays to zero at a rate of 
\begin{eqnarray*}
    O\Big(\frac{L^{-p/d}}{(1-\gamma)^2}\Big)+O\Big(\frac{\sqrt{L\log(NT)}}{(1-\gamma)^2\sqrt{\epsilon NT}}\Big),
\end{eqnarray*}
uniformly in $i,t,T_1,T_2$, with probability at least $1-O(N^{-1}T^{-1})$. Based on this result and Condition A5(ii), we can similarly derive the upper error bound for the second term. This completes the proof of this step.

\noindent \textbf{Step 2}. Recall that $\mathcal{S}_{\varepsilon}$ is an $\varepsilon$-net of $\mathcal{S}=[0,1]^d$ such that for any $s\in [0,1]^d$, there exists some $s'\in \mathcal{S}_{\varepsilon}$ that satisfies $\|s-s'\|_2\le \varepsilon=\sqrt{d}/(NT)^4$. It follows from Lemma 2.2 of \cite{mendelson2008uniform} that there exist some  $\mathcal{S}_{\varepsilon}^0$ that covers the ball $\{s:\|s\|_2\le \sqrt{d}\}$ with number of elements upper bounded by $5^d (NT)^{4d}$. For any $s^0=(s_1,\cdots,s_d)^\top\in \mathcal{S}_{\varepsilon}^0$, consider its restriction to $\mathcal{S}$, denoted by $s'=(\max(\min(s_1,1),0), \cdots, \max(\min(s_d,1),0))^\top$. It is immediate to see that for any $s\in \mathcal{S}$, $\|s^0-s\|_2\le \|s'-s\|_2$. Let $\mathcal{S}_{\varepsilon}$ denote the set of these restricted vectors, i.e., $\{s': s^0\in \mathcal{S}_{\varepsilon}^0\}$. It follows that $\mathcal{S}_{\varepsilon}$ corresponds to an $\varepsilon$-net of $\mathcal{S}$ as well. Using similar arguments to Step 1 of the proof, we can show that
\begin{eqnarray*}
    \max_{(T_1,T_2):T_2-T_1\ge \epsilon T}\Big\| \frac{W_{[T_1,T_2]}^{-1}}{N(T_2-T_1)} \sum_{i=1}^N \sum_{t=T_1}^{T_2-1} \phi_{L}(A_{i,t},S_{i,t}) \delta_{i,t}^*\Big\|_2\preceq \frac{\sqrt{L}}{(1-\gamma)^2},
\end{eqnarray*}
Combining this together with \eqref{eqn:basis2.5} yields that $|Z_{u,a,s,j}-Z_{u,a,s',j}|=O((1-\gamma)^{-2}L \|s-s'\|_2)$ where the big-$O$ term is uniform in $u,a,s,s',j$ and $Z_{u,a,s,j}$ is defined in \eqref{eqn:maxrandomvectors}. Consequently, for any $s\in \mathcal{S}$, there exists some $s'\in \mathcal{S}_{\varepsilon}$ such that $|Z_{u,a,s,j}-Z_{u,a,s',j}|$ is upper bounded by
\begin{eqnarray}\label{eqn:maxStep2upperbound}
    O\Big(\frac{L}{(1-\gamma)^2 (NT)^4}\Big).
\end{eqnarray}
This allows us to upper bound $|\textrm{TS}_{\infty}^*-\textrm{TS}_{\infty}^{**}|$ by \eqref{eqn:maxStep2upperbound} as well. 

Using similar arguments, we can show that $|\textrm{TS}_{\infty}^{b,*}-\textrm{TS}_{\infty}^{b,**}|$ is of the order \eqref{eqn:maxStep2upperbound} as well, with probability at least $1-O(N^{-1}T^{-1})$. This completes the proof of this step. 

\noindent \textbf{Step 3}. As we have commented in the outline, the proof is based on the high-dimensional martingale central limit theorem developed by \cite{belloni2018high}. Let $Z$ and $Z^b$ denote the high-dimensional random vectors formed by stacking the random vectors in the set \eqref{eqn:maxrandomvectors} and \eqref{eqn:maxrandomvectors0}, respectively. Notice that $Z$ can be represented as $\sum_{i,t} Z_{i,t}$ where each $Z_{i,t}$ depends on the data tuple $(S_{i,t},A_{i,t},R_{i,t},S_{i,t+1})$. We first observe that it corresponds to a sum of high-dimensional martingale difference. Specifically, for any integer $1\le g\le NT$, let $i(g)$ and $t(g)$ be the quotient and the remainder of $g+T-1$ divided by $T$ that satisfy
\begin{eqnarray*}
	g=\{i(g)-1\} T+t(g)+1\,\,\,\,\hbox{and}\,\,\,\,0\le t(g)<T.
\end{eqnarray*}
Let $\mathcal{F}^{(0)}=\{S_{1,0},A_{1,0}\}$. Then we recursively define $\{\mathcal{F}^{(g)}\}_{1\le g\le NT}$ as follows:
\begin{eqnarray*}
	\mathcal{F}^{(g)}=\left\{\begin{array}{ll}
		\mathcal{F}^{(g-1)} \cup \{ R_{i(g),t(g)}, S_{i(g),t(g)+1}, A_{i(g),t(g)+1} \}, & \hbox{if}~t(g)<T-1;\\
		\mathcal{F}^{(g-1)} \cup \{ R_{i(g),T-1}, S_{i(g),T}, S_{i(g)+1,0}, A_{i(g)+1,0} \}, & \hbox{otherwise}.
	\end{array} 
	\right.
\end{eqnarray*}
This allows us to rewrite $Z$ as $\sum_{g=1}^{NT} Z^{(g)}=\sum_{g=1}^{NT} Z_{i(g),t(g)}$. Similarly, we can rewrite $Z^b$ as $\sum_{g=1}^{NT} Z^{b,(g)}$. Under the Markov assumption and the conditional (mean) independence assumptions in \eqref{eqn:reward} and \eqref{eqn:transition} (e.g., the future state and the conditional mean of the immediate reward are independent of the history given the current state-action pair), $Z$ corresponds to a sum of martingale difference sequence with respect to the filtration $\{\sigma(\mathcal{F}^{(g)})\}_{g\ge 0}$ where $\sigma(\mathcal{F})$ denotes the $\sigma$-algebra generated by $\mathcal{F}$. 

Due to the existence of the max operator, $\textrm{TS}_1^{**}$ is a non-smooth function of $Z$. We next approximate the maximum using a smooth surrogate. Let $\theta$ be a sufficiently large real number. Consider the following smooth approximation of the maximum function, 
\begin{eqnarray*}
	F_{\theta}(\{Z_{u,a,s,j}\}_{u,a,s,j})=\frac{1}{\theta} \log(\sum_{u,a,s,j} \exp( \theta Z_{u,a,s,j})).
\end{eqnarray*}
As we have restricted the state space to $\mathcal{S}_{\varepsilon}$, the number of cardinality of the quadruples $(u,a,s,j)$ can be upper bounded by $O(N^cT^c)$ for some constant $c>0$. 
It can be shown that
\begin{eqnarray}\label{eqn:smoothapp}
	\max_{u,a,s,j} Z_{u,a,s,j} \le F_{\theta}(\{Z_{u,a,s,j}\}_{u,a,s,j})\le \max_{u,a,s,j} Z_{u,a,s,j} + \frac{C\log (NT)}{\theta},
\end{eqnarray}
for some constant $C>0$. See e.g., Equation (37) of \cite{cherno2014}. 

For a given $z$, consider the probability $\prob(\textrm{TS}_{\infty}^{**}\le z)$. According to Section B.1 of \cite{belloni2018high}, for any $\delta>0$, there exists a thrice differentiable function $h$ that satisfies $|h'|\le \delta^{-1}$, $|h''|\le \delta^{-2} K$ and $|h'''|\le \delta^{-3}K$ for some universal constant $K>0$ such that
\begin{eqnarray}\label{eqn:smoothapp1}
	\mathbb{I}(x\le z+\delta)\le h(x)\le \mathbb{I}(x\le z+4\delta).
\end{eqnarray}

Define a composite function $m(\{Z_{u,a,s,j}\}_{u,a,s,j})=h\circ F_{\theta}(\{Z_{u,a,s,j}\}_{u,a,s,j})$. By setting $\delta$ to $C\log(NT)/\theta$, it follows from \eqref{eqn:smoothapp} and \eqref{eqn:smoothapp1} that
\begin{eqnarray}\label{eqn:smoothapp2}
	\prob(\textrm{TS}_{\infty}^{**}\le z)\le \Mean m(\{Z_{u,a,s,j}\}_{u,a,s,j})\le 	\prob(\textrm{TS}_{\infty}^{**}\le z+4\delta).
\end{eqnarray}
Similarly, we have
\begin{eqnarray*}
	\prob(\textrm{TS}_{\infty}^{b,**}\le z|\textrm{Data})\le \Mean [m(\{Z_{u,a,s,j}^b\}_{u,a,s,j})|\textrm{Data}]\le 	\prob(\textrm{TS}_{\infty}^{b,**}\le z+4\delta|\textrm{Data}),
\end{eqnarray*}
where $Z_u^b$ is defined in \eqref{eqn:maxrandomvectors0}. This together with \eqref{eqn:smoothapp2} yields that
\begin{eqnarray}\label{eqn:probbound}
\begin{aligned}
	\max\{\max_z |\prob(\textrm{TS}_{\infty}^{b,**}\le z|\textrm{Data})-\prob(\textrm{TS}_{\infty}^{**}\le z-4\delta)|,\\
	\max_z |\prob(\textrm{TS}_{\infty}^{b,**}\le z|\textrm{Data})-\prob(\textrm{TS}_{\infty}^{**}\le z+4\delta)|\}\\\le |\Mean m(\{Z_{u,a,s,j}\}_{u,a,s,j})- \Mean [m(\{Z_{u,a,s,j}^b\}_{u,a,s,j})|\textrm{Data}]| .
\end{aligned}	
\end{eqnarray}

We next apply Corollary 2.1 of \cite{belloni2018high} to establish an upper bound for the last line of \eqref{eqn:probbound}. Similar to Lemma 4.3 of \cite{cherno2014}, we can show that $c_0\equiv \sup_{z,z'} |m(z)-m(z')|\le 1$, 
\begin{eqnarray*}
	c_2&\equiv& \sup_{z}\sum_{j_1,j_2} \left|\frac{\partial^2 m(z)}{\partial z_{j_1}\partial z_{j_2}}\right|\preceq \delta^{-2}+\delta^{-1} \theta,\\
	c_3&\equiv& \sup_{z}\sum_{j_1,j_2,j_3} \left|\frac{\partial^3 m(z)}{\partial z_{j_1}\partial z_{j_2}\partial z_{j_3}}\right|\preceq \delta^{-3}+\delta^{-2} \theta + \delta^{-1} \theta^2.
\end{eqnarray*}
In addition, similar to Lemma \ref{lemmamatrixnonstat}, we can show that the quadratic variation process \\$\sum_g \Mean \{Z^{(g)} (Z^{(g)})^\top |\mathcal{F}^{(g-1)} \}$ will converge to some deterministic matrix with elementwise approximation error upper bounded by $O((1-\gamma)^{-4}L^{3/2}\sqrt{\log (NT)}/(NT)^{3/2})$, with probability at least $1-O(N^{-1} T^{-1})$. So is the conditional covariance matrix of $Z^b$ given the data, e.g., $$\sum_g \Mean [Z^{b,(g)} (Z^{b,(g)})^\top |\{S_{i,t},A_{i,t},R_{i,t}\}_{1\le i\le N,0\le t\le T} ].$$
Moreover, by \eqref{eqn:basis2}, the sum of third absolute moments of each element in $Z^{(g)}$ is upper bounded by $O((1-\gamma)^{-6}\epsilon^{-1/2} (NT)^{-2}L^2)$. %Let $\delta=\theta^{-1}$. 
It follows from Corollary 2.1 of \cite{belloni2018high} that the last line of \eqref{eqn:probbound} is upper bounded by
\begin{eqnarray*}
	 \frac{\theta^{2}L^{3/2} \sqrt{\log (NT)}\log^2(NT)}{(1-\gamma)^4( NT)^{3/2}}+\frac{\theta^3 L^2\log^3(NT)}{(1-\gamma)^6\sqrt{\epsilon} (NT)^2}
\end{eqnarray*}
with probability at least $1-O(N^{-1} T^{-1})$. 

Meanwhile, under Assumption A4(iv) that the conditional variance of temporal difference error $\delta_{i,t}^*$ is lower bounded by $c(1-\gamma)^{-2}$ for some constant $c>0$. Additionally, by setting $\gamma'$ in A5(ii) to $0$, we obtain that the minimum eigenvalue of $\Mean [\phi_L(A_t,S_t)\phi_L^\top(A_t,S_t)]$ is bounded away from zero. Using similar arguments to Step 1 of the proof, we can show that the conditional variance of each $Z_{u,a,s,j}^b$ given the data is lower bounded by $C(1-\gamma)^{-2}\|\phi_L(a,s)\|_2^2/(NT)$. By \eqref{eqn:basis2.5}, it can be uniformly lower bounded by $C(1-\gamma)^{-2}/(NT)$ for some constant $C>0$. As such, it follows from the anti-concentration inequality for the maximum of Gaussian random vector \citep[Theorem 1]{chernozhukov2017detailed} that
\begin{eqnarray*}
	\sup_z |\prob(\textrm{TS}_{\infty}^{b,**}\le z-4\delta|\textrm{Data})-\prob(\textrm{TS}_{\infty}^{b,**}\le z+4\delta|\textrm{Data})|\preceq \frac{\sqrt{NT}\log^{3/2}(NT)}{(1-\gamma)^{-1}\theta},
\end{eqnarray*}
with probability at least $1-O(N^{-1}T^{-1})$. %as long as $\theta=O(N^CT^C)$ for some sufficiently large $C>0$. 

In view of \eqref{eqn:probbound}, by setting $\theta=(1-\gamma)^{7/4}L^{-1/2}\epsilon^{1/8}(NT)^{3/8}\log^{-3/8}(NT)$, %for some constant $1/2<c<2/3-2c_4/3$, both the RHS of \eqref{eqn:bound1} and \eqref{eqn:bound2} decay to zero. , 
we have
\begin{eqnarray*}
	\sup_{z} |\prob(\textrm{TS}_{\infty}^{b,**}\le z|\textrm{Data})-\prob(\textrm{TS}_{\infty}^{**}\le z)|\preceq %\frac{\sqrt{L}\log^{3/2}(NT)}{(1-\gamma)(NT)^{1/4}}\\+
    \frac{\sqrt{L}\log^{15/8}(NT)}{(1-\gamma)^{3/4}(\epsilon NT)^{1/8}}, 
\end{eqnarray*}
with probability at least $1-O(N^{-1}T^{-1})$. Similarly, based on \eqref{eqn:step3initial}, we can show that
\begin{eqnarray*}
	\sup_{z} |\prob(\textrm{TS}_1^{b}\le z|\textrm{Data})-\prob(\textrm{TS}_1\le z)|\preceq \frac{\sqrt{L}\log^{15/8}(NT)}{(1-\gamma)^{3/4}(\epsilon NT)^{1/8}}+\frac{L^{-p/d}\sqrt{NT\log(NT)} }{(1-\gamma)^2}.
\end{eqnarray*}
The proof is hence completed. 

\subsubsection{Normalized Maximum-Type Tests}\label{sec:proofmax}
For any $u$, $a$ and $s$, define the variance estimator $\widehat{\sigma}_u^2(a,s)$ by
\begin{eqnarray*}
    \frac{\phi_L^\top(a,s) \widehat{W}_{[T-\kappa,T-u]}^{-1}}{N^2 (\kappa-u)^2}  \left[\sum_{i=1}^N \sum_{t=T-\kappa}^{T-u-1} \phi_L(A_{i,t},S_{i,t}) \phi_L^\top(A_{i,t},S_{i,t}) \delta_{i,t}^2(\widehat{\beta}_{[T-\kappa,T-u]}) \right] \{\widehat{W}_{[T-\kappa,T-u]}^{-1}\}^\top \phi_L^\top(a,s)\\+\frac{1}{N^2 u^2} \phi_L^\top(a,s) \widehat{W}_{[T-u,T]}^{-1} \left[\sum_{i=1}^N \sum_{t=T-u}^{T-1} \phi_L(A_{i,t},S_{i,t}) \phi_L^\top(A_{i,t},S_{i,t}) \delta_{i,t}^2(\widehat{\beta}_{[T-u,T]}) \right] \{\widehat{W}_{[T-u,T]}^{-1}\}^\top \phi_L^\top(a,s).
\end{eqnarray*}
We next show that the normalized maximum-type test has good size property. The proof is very similar to that in Section \ref{sec:proofunmax}. We provide a sketch of the proof and outline a key step only. Define $\textrm{TS}_n^*$ and $\textrm{TS}_n^{b,*}$ to be versions of $\textrm{TS}_n$ and $\textrm{TS}_n^{b}$, respectively, where the variance estimator $\widehat{\sigma}_u^2(a,s)$ is replaced with its oracle value $\sigma_u^2(a,s)$. The key step here is to upper bound the differences $|\textrm{TS}_n^*-\textrm{TS}_n|$ and $|\textrm{TS}_n^{b,*}-\textrm{TS}_n^{b}|$. 

Toward that end, we first upper bound the difference between $\widehat{\sigma}_u^2(a,s)$ and $\sigma_u^2(a,s)$. Using similar arguments to Step 1 of the proof in Section \ref{sec:proofunmax}, we can show that with probability at least $1-O(N^{-1}T^{-1})$, (i)$\max_{T_2-T_1\ge \epsilon T} \|\widehat{W}_{[T_1,T_2]}^{-1}-W_{[T_1,T_2]}^{-1}\|_2=O((1-\gamma)^{-2}\sqrt{L(\epsilon NT)^{-1}\log(NT)})$; (ii) $|\delta_{i,t}^2(\widehat{\beta}_{[T_1,T_2]})-(\delta_{i,t}^*)^2|=O((1-\gamma)^{-3}L^{-p/d})+O((1-\gamma)^{-3}\sqrt{L(\epsilon NT)^{-1}\log(NT)})$ where the bound is uniform in $i,t,T_1,T_2$; 
\begin{eqnarray*}
    (\textrm{iii}) \max_{T_2-T_1\ge \epsilon T}\lambda_{\max}\left\{ \frac{1}{N(T_2-T_2)}\sum_{i=1}^N \sum_{t=T_1}^{T_2-1} \phi_L(A_{i,t},S_{i,t})\phi_L^\top(A_{i,t},S_{i,t})\right\}=O(1),
\end{eqnarray*}
(iv) the matrix $N^{-1}(T_2-T_2)^{-1}\sum_{i=1}^N \sum_{t=T_1}^{T_2-1} \phi_L(A_{i,t},S_{i,t})\phi_L^\top(A_{i,t},S_{i,t})(\delta_{i,t}^*)^2$ converges uniformly to its expectation across all pairs $(T_1,T_2)$ such that $T_2-T_1\ge \epsilon T$, with the uniform approximation error given by $O((1-\gamma)^{-2}\sqrt{L(\epsilon NT)^{-1}\log(NT)})$. These allow us to establish the following error bound
\begin{eqnarray}\label{eqn:sigmauerrorbound}
    |\widehat{\sigma}_u^2(a,s)-\sigma_u^2(a,s)|\le \frac{\|\phi_L(a,s)\|_2^2}{NT\tau_u^2} \Big[O\Big(\frac{L^{-p/d}}{(1-\gamma)^{5}}\Big)+O\Big(\frac{\sqrt{L(\epsilon NT)^{-1}\log(NT)}}{(1-\gamma)^{5}}\Big)\Big],
\end{eqnarray}
with probability at least $1-O(N^{-1}T^{-1})$, where the big-$O$ terms on the RHS are uniform in $a,s,u$.

Similar to Step 3 of the proof for the unnormalized test in Section \ref{sec:proofunmax}, we can show that $\sigma_u^2(a,s)\ge c(1-\gamma)^{-2}\|\phi_L(a,s)\|_2^2/(NT\tau_u^2)$ for some constant $c>0$ and hence
\begin{eqnarray*}
    \sup_{a,s,u}\frac{|\widehat{\sigma}_u^2(a,s)-\sigma_u^2(a,s)|}{\sigma_u^2(a,s)}=O\Big(\frac{L^{-p/d}}{(1-\gamma)^3}\Big)+O\Big(\frac{\sqrt{L(\epsilon NT)^{-1}\log(NT)}}{(1-\gamma)^3}\Big),
\end{eqnarray*}
with probability at least $1-O(N^{-1}T^{-1})$, and hence, 
\begin{eqnarray}\label{eqn:variancebound}
    \sup_{a,s,u}\frac{|\widehat{\sigma}_u(a,s)-\sigma_u(a,s)|}{\sigma_u(a,s)}=O\Big(\frac{L^{-p/d}}{(1-\gamma)^3}\Big)+O\Big(\frac{\sqrt{L(\epsilon NT)^{-1}\log(NT)}}{(1-\gamma)^3}\Big).
\end{eqnarray}
Meanwhile, it follows from \eqref{eqn:sigmauerrorbound} that there exists some constant $C>0$ such that $\widehat{\sigma}^2_u(a,s)\ge C(1-\gamma)^{-2}\|\phi_L(a,s)\|_2^2/(NT\tau_u^2)$ for any $u,a,s$, with probability at least $1-O(N^{-1}T^{-1})$. 

In view of these, using similar arguments to Steps 1 and 2 of the proof for the unnormalized test, we can show that the bootstrapped normalized test statistics can be upper bounded by
\begin{eqnarray}\label{eqn:normalizedbound}
    O(\sqrt{\log(NT)}),
\end{eqnarray}
with probability at least $1-O(N^{-1}T^{-1})$. Additionally, as discussed in Step 3 of the proof for the unnormalized test, each $Z_{u,a,s,j}/\sigma_u(a,s)$ can be represented as a sum of high-dimensional martingale difference sequence. We can first discretize the state space in a manner similar to Step 2 of the proof, and then apply the martingale concentration inequality \citep[see e.g.,][Theorem 1.1]{tropp2011freedman} and the Bonferroni's inequality to obtain a uniform upper error bound for $\{Z_{u,a,s,j}/\sigma_u(a,s)\}_{u,a,s,j}$. This %together with the lower bound of $\widehat{\sigma}^2_u(a,s)$ yields 
implies that the normalized test statistics is of the order of magnitude \eqref{eqn:normalizedbound} as well, with probability at least $1-O(N^{-1}T^{-1})$. 

To the contrary, the scaled unnormalized test $(1-\gamma)^2\sqrt{NT}\textrm{TS}_{\infty}$ and its bootstrap version $(1-\gamma)^2\sqrt{NT}\textrm{TS}_{\infty}^b$ are of the order of magnitude $O(\sqrt{L\log(NT)})$. When compared to the error bound in \eqref{eqn:normalizedbound}, it can be seen that the normalized test achieves a smaller order of  magnitude by a factor of $L^{-1/2}$. This reduction results from the normalization, which makes the order of each $Z_{u,a,s,j}$ independent of $\phi_L(a,s)$, thus avoiding multiplying with an additional supremum term $\sup_{a,s}\|\phi_L(a,s)\|_2$ (which is of the order $O(\sqrt{L})$ according to \eqref{eqn:basis2}) when taking the maximum over $Z_{u,a,s,j}$. 

Combining \eqref{eqn:normalizedbound} together with \eqref{eqn:variancebound}, we obtain that
\begin{eqnarray*}
    |\textrm{TS}_n-\textrm{TS}_n^*|=O\Big(\frac{L^{-p/d}\sqrt{\log(NT)}}{(1-\gamma)^3}\Big)+O\Big(\frac{\sqrt{L(\epsilon NT)^{-1}}\log(NT)}{(1-\gamma)^3}\Big).
\end{eqnarray*}
With this upper bound, the rest of the proof can be established in a similar manner to the proof for the unnormalized test. We omit the details to save space. 

\subsubsection{\texorpdfstring{$\ell_1$}--type Test}
\label{sec:ell1test}
The proof for the $\ell_1$-type test is more involved. It is divided into three steps. 

In Step 1, we show there exists some constant $C>0$ such that 
\begin{eqnarray}\label{eqn:step1ell1}
\prob(|\textrm{TS}_1-\textrm{TS}_1^*|\le C\kappa \sqrt{L N^{-1}T^{-1}\log(NT)} ) = 1-O\Big(\frac{1}{NT}\Big).
\end{eqnarray}
where
\begin{eqnarray*}
	\textrm{TS}_1^*=\max_{\epsilon T<u<(1-\epsilon) T} \sqrt{\frac{u(T-u)}{T^2}}\left\{\frac{1}{T}\sum_{t=0}^{T-1}\sum_a \int_s |\widehat{Q}_{[0,u]}(a,s)-\widehat{Q}_{[u,T]}(a,s) | \pi^b_t(a|s) p_t^b(s) ds \right\},
\end{eqnarray*}
where $\pi_t^b$ denotes the behavior policy at time $t$, $p_t^b$ denotes the marginal distribution of $S_t$ under the behavior policy, and $\kappa$ is a shorthand for $(1-\gamma)^{-2}L^{-p/d}+(1-\gamma)^{-2}\sqrt{L\log(NT)/(\epsilon NT)}$. 
By definition, $\textrm{TS}_1^*$ corresponds to a version of $\textrm{TS}_1$ assuming the marginal distribution of the observed state-action pairs is known to us. 

In the second step, we notice that we have established a uniform upper error bound for $\max_{i,t,T_1,T_2} |\widehat{Q}_{[T_1,T_2]}^{b}(A_{i,t},S_{i,t})-\widehat{Q}_{[T_1,T_2]}^{b,0}(A_{i,t},S_{i,t})|$ in Step 2 of the proof for the unnormalized test. This error bound in turn leads to the following bound for the difference between $\textrm{TS}_1^{b}$ and $\textrm{TS}_1^{b,0}$,
\begin{eqnarray}\label{eqn:step2initial}
\prob\Big(|\textrm{TS}_1^{b}-\textrm{TS}_1^{b,0}|\le \frac{C L^{3/2}\log(NT)}{(1-\gamma)^3\epsilon NT}+\frac{CL^{-p/d}\sqrt{\log(NT)}}{(1-\gamma)^3\sqrt{\epsilon NT}}\Big)= 1-O\Big(\frac{1}{NT}\Big),
\end{eqnarray}
for some constant $C>0$, where
\begin{eqnarray*}
	\textrm{TS}_1^{b,0}=\max_{\epsilon T<u<(1-\epsilon) T} \sqrt{\frac{u(T-u)}{T^2}}\left\{\frac{1}{NT}\sum_{i=1}^N\sum_{t=0}^{T-1} |\widehat{Q}_{[0,u]}^{b,0}(A_{i,t},S_{i,t})-\widehat{Q}_{[u,T]}^{b,0}(A_{i,t},S_{i,t})| \right\}.
\end{eqnarray*}
Using similar arguments to the proof of \eqref{eqn:step1ell1} in Step 1, we can show that $|\textrm{TS}_1^{b}-\textrm{TS}_1^{b,0}|$ is of the order $O(\kappa \sqrt{L N^{-1}T^{-1}\log(NT)})$ as well, with probability at least $1-O(N^{-1}T^{-1})$, where
\begin{eqnarray*}
    \textrm{TS}_1^{b,*}=\max_{\epsilon T<u<(1-\epsilon) T} \sqrt{\frac{u(T-u)}{T^2}}\left\{\frac{1}{T}\sum_{t=0}^{T-1}\sum_a \int_s |\widehat{Q}_{[0,u]}^{b,0}(a,s)-\widehat{Q}_{[u,T]}^{b,0}(a,s) | \pi^b_t(a|s) p_t^b(s) ds \right\}.
\end{eqnarray*}
This together with \eqref{eqn:step2initial} yields that
\begin{eqnarray}\label{eqn:step2ell1}
\prob\Big(|\textrm{TS}_1^b-\textrm{TS}_1^{b,*}|\le \frac{c L^{3/2}\log(NT)}{(1-\gamma)^3\epsilon NT}+\frac{cL^{-p/d}\sqrt{L\log(NT)}}{(1-\gamma)^3\sqrt{NT}}\Big) = 1-O\Big(\frac{1}{NT}\Big),
\end{eqnarray}
for some constants $c>0$, under the given conditions on $L$ (see A9) and $\epsilon$ (see A1). %It also implies that
%\begin{eqnarray}\label{eqn:step2}
%	\prob(|\sqrt{NT}(\textrm{TS}_1^b-\textrm{TS}_1^{b,*})|\le C (NT)^{-c} |\textrm{Data} ) \stackrel{P}{\to} 1.
%\end{eqnarray}

Meanwhile, we define $\textrm{TS}_1^{**}$ to be a version of $\textrm{TS}_1^*$ with $\widehat{Q}_{[T_1,T_2]}(a,s)$ replaced by %the leading term in Assumption (A1). 
\begin{eqnarray*}
    \frac{1}{N(T_2-T_1)}\sum_{i=1}^N \sum_{t=T_1}^{T_2-1} \phi_L^\top(a,s)W_{[T_1,T_2]}^{-1}\phi_L(A_{i,t},S_{i,t})\delta_{i,t}^*.
\end{eqnarray*}
By \eqref{eqn:asymptoticexpansion}, we have
\begin{eqnarray}\label{eqn:step3ell1}
\begin{aligned}
\prob\Big(|\textrm{TS}_1^*-\textrm{TS}_1^{**}|\le \frac{C L^{-p/d}}{(1-\gamma)^3}+\frac{C L^{3/2} \log (NT)}{(1-\gamma)^3\epsilon  NT} \Big) =1-O\Big(\frac{1}{NT}\Big),%\\
%&\prob(|\sqrt{NT}(\textrm{TS}_1^{b,*}-\textrm{TS}_1^{b,**})|\le C (NT)^{-c} |\textrm{Data} ) \stackrel{P}{\to} 1.
\end{aligned}
\end{eqnarray}
for some constant $C>0$.

%$\prob(|\sqrt{NT}(\textrm{TS}_1-\textrm{TS}_1^*)|\le C (NT)^{-c} ) \to 1$. 
%Using similar arguments, one can show that 
%\begin{eqnarray}\label{eqn:step1b}
%\prob(|\sqrt{NT}(\textrm{TS}_1^b-\textrm{TS}_1^{b,*})|\le C (NT)^{-c} | \textrm{Data} ) \to 1.
%\end{eqnarray}
Combining the results in \eqref{eqn:step1ell1}-\eqref{eqn:step3ell1}, we have shown that
\begin{eqnarray}\label{eqn:step4initial}
\begin{aligned}
&\prob\Big(|\textrm{TS}_1-\textrm{TS}_1^{**}|\le \frac{c L^{-p/d}}{(1-\gamma)^3}+\frac{c L^{3/2} \log (NT)}{(1-\gamma)^3\epsilon  NT} \Big)= 1-O\Big(\frac{1}{NT}\Big),\\
&\prob\Big(|\textrm{TS}_1^{b}-\textrm{TS}_1^{b,*}|\le \frac{c L^{-p/d}}{(1-\gamma)^3}+\frac{c L^{3/2} \log (NT)}{(1-\gamma)^3\epsilon  NT} \Big)= 1-O\Big(\frac{1}{NT}\Big),
\end{aligned}
\end{eqnarray}
for some constant $c>0$. Define
\begin{eqnarray}\label{eqn:randomvectors}
	\Big\{Z_u\stackrel{\Delta}{=}\tau_u\Big[\frac{W_{[0,u]}^{-1} }{Nu}\sum_{i=1}^N \sum_{t=0}^{u-1} \phi_{L}(A_{i,t},S_{i,t})\delta_{i,t}^*-\frac{W_{[u,T]}^{-1}}{N(T-u)} \sum_{i=1}^N \sum_{t=u}^{T-1} \phi_{L}(A_{i,t},S_{i,t})\delta_{i,t}^*\Big]: u \Big\},\quad\\\label{eqn:randomvectors0}
    \left\{Z_u^b\stackrel{\Delta}{=}\tau_u \Big[\frac{W_{[0,u]}^{-1} }{Nu} \sum_{i=1}^N \sum_{t=0}^{u-1} \phi_{L}(A_{i,t},S_{i,t})\delta_{i,t}^*e_{i,t}-\frac{W_{[u,T]}^{-1}}{N(T-u)} \sum_{i=1}^N \sum_{t=u}^{T-1} \phi_{L}(A_{i,t},S_{i,t})\delta_{i,t}^*e_{i,t}\Big]: u \right\}.\quad
\end{eqnarray}
Notice that $\textrm{TS}_1^{**}$ and $\textrm{TS}_1^{b,*}$ can be represented as functions of $\{Z_u\}_u$ and $\{Z_u^b\}_u$, respectively. The last step is again to apply the high-dimensional martingale central limit theorem to bound their Kolmogorov distance.

%Finally, in the last step, we show that Theorem \ref{thm:size} holds, based on \eqref{eqn:step4initial} and \eqref{eqn:step4}. 
%Finally, notice that $\sqrt{NT} \textrm{TS}_1^{b,**}$ has a density function that is bounded below 

We next detail the proofs for Steps 1 and 3. 

\noindent \textbf{Step 1}. For each $u$, we aim to develop a concentration inequality to bound the difference
\begin{eqnarray}\label{eqn:diff}
\begin{aligned}
	\left|\sqrt{\frac{u(T-u)}{T^2}}\left\{\frac{1}{NT}\sum_{t=0}^{T-1}\sum_{i=1}^N\sum_a \int_s [|\widehat{Q}_{[0,u]}(A_{i,t},S_{i,t})-\widehat{Q}_{[u,T]}(A_{i,t},S_{i,t})|\right.\right.\\-\left.\left.|\widehat{Q}_{[0,u]}(a,s)-\widehat{Q}_{[u,T]}(a,s) |] \pi^b_t(a|s) p_t^b(s)ds \right\}\right|.
\end{aligned}	
\end{eqnarray}
According to Lemma \ref{lemmaQ} (see \eqref{eqn:betaK-betastarell2bound}), we have that $\sup_{T_2-T_1\ge \epsilon T} \|\widehat{\beta}_{[T_1,T_2]}-\beta^*\|_2=O(\kappa)$, with probability at least $1-O(N^{-1}T^{-1})$. 

Define the set $\mathcal{B}(C)=\{\beta\in \mathbb{R}^L: \| \beta-\beta^*\|_2 \le C\kappa \}$. It follows that there exists some sufficiently large constant $C>0$ such that $\widehat{\beta}_{[T_1,T_2]} \in \mathcal{B}(C)$ with probability at least $1-O(N^{-1}T^{-1})$. \eqref{eqn:diff} can thus be upper bounded by 
\begin{eqnarray}\label{eqn:upperbound}
	\sup_{\substack{\beta_1\in \mathcal{B}(C)\\ \beta_2\in \mathcal{B}(C)}}\left|\frac{1}{2NT}\sum_{t=0}^{T-1} \sum_{i=1}^N \{|\phi_L^\top(A_{i,t},S_{i,t}) (\beta_1-\beta_2)|-\Mean |\phi_L^\top(A_{i,t},S_{i,t}) (\beta_1-\beta_2)|\} \right|.
\end{eqnarray}
The upper bound for \eqref{eqn:upperbound} can be established using similar arguments to in the proof of Lemma \ref{lemmamatrixnonstat}. To save space, we only provide a sketch of the proof here. Please refer to the proof of Lemma \ref{lemmamatrixnonstat} for details. 

Notice that the suprema in \eqref{eqn:upperbound} are taken with respect to infinitely many $\beta$s. As such, standard concentration inequalities are not applicable to bound \eqref{eqn:upperbound}. Toward that end, we first take an $\varepsilon$-net of $\mathcal{B}(C)$ for some sufficiently small $\varepsilon>0$, denote by  $\mathcal{B}^*(C)$, such that for any $\beta\in \mathcal{B}(C)$, there exists some $\beta^*\in \mathcal{B}^*(C)$ that satisfies $\|\beta-\beta^*\|_2\le \varepsilon$. The purpose of introducing an $\varepsilon$-net is to approximate these sets by collections of finitely many $\beta$s so that concentration inequalities are applicable to establish the upper bound. Set $\varepsilon=C\kappa(NT)^{-2}$. It follows from Lemma 2.2 of \cite{mendelson2008uniform} that there exist some $\mathcal{B}^*(C)$ with number of elements upper bounded by $5^L (NT)^{2L}$. 

By Lemma \ref{lemmabasis}, we have $\sup_{a,s}\|\phi_L(a,s)\|_2=O(\sqrt{L})$. Thanks to this uniform bound, the quantity within the absolute value symbol in \eqref{eqn:upperbound} is a Lipschitz continuous function of $(\beta_1,\beta_2)$, with the Lipschitz constant upper bounded by $O(\sqrt{L})$. As such, \eqref{eqn:upperbound} can be approximated by 
\begin{eqnarray}\label{eqn:upperbound1}
\sup_{\beta_1,\beta_2\in \mathcal{B}^*(C)}\Bigg|\underbrace{\frac{1}{2NT}\sum_{t=0}^{T-1} \sum_{i=1}^N \sum_a \{|\phi_L^\top(A_{i,t},S_{i,t}) (\beta_1-\beta_2)|-\Mean |\phi_L^\top(A_{i,t},S_{i,t}) (\beta_1-\beta_2)|\}}_{I(\beta_1,\beta_2)~~(\textrm{without~absolute~value})}  \Bigg|,
\end{eqnarray}
with the approximation error given by $O(C\sqrt{L}N^{-2}T^{-2} \kappa)$. %where the big-$O$ term is uniform in $u$. 

It remains to develop a concentration inequality for \eqref{eqn:upperbound1}. Since the number of elements in $\mathcal{B}^*(C)$ are bounded, we could develop a tail inequality for the quantity within the absolute value symbol in \eqref{eqn:upperbound1} for each combination of $\beta_1$ and $\beta_2$, and then apply Bonferroni's inequality to establish a uniform upper error bound. More specifically, for each pair $(\beta_1,\beta_2)$, let %notice that $\phi$
\begin{eqnarray*}
	I^*(\beta_1,\beta_2)=\frac{1}{2NT}\sum_{t=0}^{T-1} \sum_{i=1}^N [\Mean \{|\phi_L^\top(A_{i,t},S_{i,t}) (\beta_1-\beta_2)| |S_{i,t-1} \}-\Mean |\phi_L^\top(A_{i,t},S_{i,t}) (\beta_1-\beta_2)|],
\end{eqnarray*}
with the convention that $S_{i,-1}=\emptyset$. Notice that $I(\beta_1,\beta_2)-I^*(\beta_1,\beta_2)$ forms a mean-zero martingale under \eqref{eqn:reward} and \eqref{eqn:transition}, we can first apply the martingale concentration inequality to show that 
\begin{eqnarray}\label{eqnI}
|I(\beta_1,\beta_2)-I^*(\beta_1,\beta_2)|=O(\varepsilon \sqrt{L N^{-1} T^{-1}\log(NT)}),
\end{eqnarray}
with probability at least $1-O\{(NT)^{-CL}\}$ for some sufficiently large constant $C>0$. 
Here, the upper bound $O(\kappa \sqrt{L N^{-1} T^{-1}\log(NT)})$ decays faster than the parametric rate, due to the fact that the variance of the summand decays to zero. Specifically, notice that $\Var\{ \phi_L^\top(A_{t},S_{t})(\beta_1-\beta_2) |S_{t-1}\}$ is upper bounded by
\begin{eqnarray*}
    \max_{a,a',s} \lambda_{\max} \left\{\int_{s'} \phi_L(a',s')\phi_L^\top(a',s') p(s'|a,s)ds'\right\} \|\beta_1-\beta_2\|_2^2 =O(\kappa^2),
\end{eqnarray*} 
where the equality is due to Lemma \ref{lemmabasis} and the fact that $p_t$s are uniformly bounded. %, and the last equality is due to the condition that $L$ grows at a rate slower than $NT$. 

Next, under A4(i) and (ii), the transition functions $\{\mathcal{T}_t\}_t$ satisfy the conditions in the statement of Theorem 3.1 in \cite{alquier2019exponential}. In addition, each summand in the definition of $I^*(\beta_1,\beta_2)$ is upper bounded by $\kappa$. We can thus apply the concentration inequality for non-stationary Markov chains developed therein to show that $|I^*(\beta_1,\beta_2)|=O(\kappa \sqrt{L N^{-1} T^{-1} \log(NT)})$, with probability at least $1-O\{(NT)^{-CL}\}$ for some sufficiently large constant $C>0$.  
This together with the upper bound for $|I(\beta_1,\beta_2)-I^*(\beta_1,\beta_2)|$ in \eqref{eqnI} and Bonferroni's inequality yields the desired uniform upper bound for \eqref{eqn:upperbound1}. This completes Step 1 of the proof.

\smallskip

\noindent \textbf{Step 3}. The proof is very similar to that for the unnormalized test. Let $Z$ and $Z^b$ denote the high-dimensional random vectors formed by stacking the random vectors in the set \eqref{eqn:randomvectors} and \eqref{eqn:randomvectors0}, respectively. Again, $Z$ can be represented as a sum of high-dimensional martingale difference  $\sum_{g=1}^{NT} Z^{(g)}$ with respect to the filtration $\{\sigma(\mathcal{F}^{(g)})\}_{g\ge 0}$. Similarly, we can rewrite $Z^b$ as $\sum_{g=1}^{NT} Z^{b,(g)}$ and $Z_u=\sum_{g=1}^{NT} Z_u^{(g)}$. The test statistic can be represented as
\begin{eqnarray*}
	\textrm{TS}_1^{**}=\max_{\epsilon T<u<(1-\epsilon) T} \underbrace{%\sqrt{%\frac{u(T-u)}{T^2}} 
    \frac{1}{T}\sum_{t=0}^{T-1} \sum_a \int_s |\phi_L^\top(a,s) Z_u| p_t^b(s)\pi_t^b(a|s)ds}_{\psi_u}.
\end{eqnarray*}

Next, we approximate the absolute value function $|x|=\max(x,0)+\max(-x,0)$ by $\theta^{-1} \{\log (1+\exp(\theta x))+\log(1+\exp(-\theta x))\}$. Define the corresponding smooth function $f_{\theta}(Z_u)$ as
\begin{eqnarray*}
	\sqrt{\frac{u(T-u)}{T^2}} \frac{1}{T\theta}\sum_{t=0}^{T-1} \sum_a \int_s \{\log(1+\exp(\theta \phi_L^\top(a,s) Z_u))+\log (1+\exp(-\theta\phi_L^\top(a,s) Z_u))\} \\\times p_t^b(s)\pi_t^b(a|s)ds.
\end{eqnarray*}
Similarly, we have $\psi_u\le f_{\theta}(Z_u)\le \psi_u+\theta^{-1}\log 2$. This together with \eqref{eqn:smoothapp} yields 
\begin{eqnarray}\label{eqn:smoothapp-1}
	\textrm{TS}_1^{**}\le F_{\theta}(\{f_{\theta}(Z_u)\}_u)\le F_{\theta}(\{\psi_u+\theta^{-1}\log 2\}_u)\le \textrm{TS}_1^{**}+\frac{\log (2T)}{\theta}.
\end{eqnarray}

Define a composite function $m(\{Z_u\}_u)=h\circ F_{\theta}(\{f_{\theta}(Z_u)\}_u)$ where the smooth function $h$ is defined in Step 3 of the proof for the unnormalized test. By setting $\delta$ to $\log(2T)/\theta$, it follows from \eqref{eqn:smoothapp-1} and \eqref{eqn:smoothapp1} that
\begin{eqnarray}\label{eqn:smoothapp3}
	\prob(\textrm{TS}_1^{**}\le z)\le \Mean m(\{Z_u\}_u)\le 	\prob(\textrm{TS}_1^{**}\le z+4\delta).
\end{eqnarray}
Similarly, we have
\begin{eqnarray*}
	\prob(\textrm{TS}_1^{b,*}\le z|\textrm{Data})\le \Mean [m(\{Z_u^b\}_u)|\textrm{Data}]\le 	\prob(\textrm{TS}_1^{b,*}\le z+4\delta|\textrm{Data}),
\end{eqnarray*}
where $Z_u^b$ is defined in \eqref{eqn:randomvectors0}. This together with \eqref{eqn:smoothapp3} yields that
\begin{eqnarray}\label{eqn:probbound1}
\begin{aligned}
	\max\{\max_z |\prob(\textrm{TS}_1^{b,*}\le z|\textrm{Data})-\prob(\textrm{TS}_1^{**}\le z-4\delta)|,\\
	\max_z |\prob(\textrm{TS}_1^{b,*}\le z|\textrm{Data})-\prob(\textrm{TS}_1^{**}\le z+4\delta)|\}\\\le |\Mean m(\{Z_u\})- \Mean [m(\{Z_u^b\}_u)|\textrm{Data}]| .
\end{aligned}	
\end{eqnarray}

Meanwhile, we can show that for this choice of $m$, $c_0\equiv \sup_{z,z'} |m(z)-m(z')|\le 1$, 
\begin{eqnarray*}
	c_2&\equiv& \sup_{z}\sum_{j_1,j_2} \left|\frac{\partial^2 m(z)}{\partial z_{j_1}\partial z_{j_2}}\right|\preceq \delta^{-2} L+\delta^{-1} \theta L,\\
	c_3&\equiv& \sup_{z}\sum_{j_1,j_2,j_3} \left|\frac{\partial^3 m(z)}{\partial z_{j_1}\partial z_{j_2}\partial z_{j_3}}\right|\preceq \delta^{-3} L^{3/2}+\delta^{-2} \theta L^{3/2} + \delta^{-1} \theta^2 L^{3/2}.
\end{eqnarray*}
In addition, similar to Lemma \ref{lemmamatrixnonstat}, we can show that both $\sum_g \Mean \{Z^{(g)} (Z^{(g)})^\top |\mathcal{F}^{(g-1)} \}$ and $\sum_g \Mean \{Z^{b,(g)} (Z^{b,(g)})^\top|\textrm{Data}\}^\top$ will converge to the same deterministic matrix with elementwise approximation error upper bounded by $C(1-\gamma)^{-4}\sqrt{L\log (NT)}/(NT)^{3/2}$ for some constant $C>0$, with probability at least $1-O(N^{-1} T^{-1})$. 

However, the anti-concentration inequality is not applicable here, since the test statistic is no longer a maximum of $\{Z_u\}_u$. Toward that end, assuming $(1-\gamma)\sqrt{NT}\textrm{TS}_1^{b,*}/\sqrt{\log(NT)}$ has a bounded density function, it follows that
\begin{eqnarray*}
	\sup_z |\prob(\textrm{TS}_1^{b,*}\le z-4\delta|\textrm{Data})-\prob(\textrm{TS}_1^{b,*}\le z+4\delta|\textrm{Data})|\preceq \frac{\sqrt{NT}\log^{3/2}(NT)}{(1-\gamma)^{-1}\theta}.
\end{eqnarray*}
%with probability at least $1-O(N^{-1}T^{-1})$. 
In view of \eqref{eqn:probbound1}, by similarly setting $\theta=(1-\gamma)^{7/4}L^{-1/2}\epsilon^{1/8}(NT)^{3/8}\log^{-3/8}(NT)$, %for some constant $1/2<c<2/3-2c_4/3$, both the RHS of \eqref{eqn:bound1} and \eqref{eqn:bound2} decay to zero. , 
we obtain 
\begin{eqnarray*}
	\sup_{z} |\prob(\textrm{TS}_1^{b,**}\le z|\textrm{Data})-\prob(\textrm{TS}_1^{**}\le z)|\preceq %\frac{\sqrt{L}\log^{3/2}(NT)}{(1-\gamma)(NT)^{1/4}}\\+
    \frac{\sqrt{L}\log^{15/8}(NT)}{(1-\gamma)^{3/4}(\epsilon NT)^{1/8}}.
\end{eqnarray*}
Finally, based on \eqref{eqn:step4initial}, one can show that
\begin{eqnarray*}
	\sup_{z} |\prob(\textrm{TS}_1^{b}\le z|\textrm{Data})-\prob(\textrm{TS}_1\le z)|\preceq \frac{\sqrt{L}\log^{15/8}(NT)}{(1-\gamma)^{3/4}(\epsilon NT)^{1/8}}+\frac{L^{-p/d}\sqrt{NT\log(NT)} }{(1-\gamma)^2}.
\end{eqnarray*}
The proof is hence completed.

\subsection{Proof of Lemmas \ref{lemmabasis} and \ref{lemmaQ}}\label{sec:prooflemmaB1}
First, we notice that \eqref{eqn:basis2} can be proven based on the proof of Theorem 3.3 of \cite{Burman1989}. Second, the B-spline basis function is non-negative and sum up to $1$. Following \citep{chen2015optimal}, we rescale the basis by multiplying it by $L^{1/2}$. This leads to the equality in \eqref{eqn:basis2.5}. Next, according to Cauchy-Schwarz inequality, the $\ell_2$-norm of $\phi_L(a,s)$ is lower bounded by its $\ell_1$-norm divided by $\sqrt{L}$. This yields the inequality in \eqref{eqn:basis2.5}. 
%The second part follows from the fact that the number of nonzero elements in the vector $\Phi(s)$ is bounded by some universal constant and that each of the basis
%function is bounded by $O(\sqrt{L})$. 
Additionally, each function in $\Phi$ is Lipschitz continuous. This yields \eqref{eqn:basis3}. 

%\subsection{Proof of Lemma \ref{lemmaQ}}\label{sec:prooflemmaB1}
The rest of the proof is organized as follows. We first outline the challenge in establishing the uniform convergence rate of the Q-function estimators under the nonstandard stationarity assumption SA3 where the reward and transition functions may remain nonstationary over time and illustrate our main idea in addressing this challenge. Next, we conduct a population-level analysis with infinitely many data to prove \eqref{eqn:basis1}. Next, we consider the finite-sample scenario and
formally establish the uniform convergence rate of the Q-estimators in \eqref{eqn:rateofconvergence}. Finally, we conduct an asymptotic analysis to prove \eqref{eqn:asymptoticexpansion}. 

\noindent \textbf{Challenge}: In FQI, we iteratively update the Q-function according to the following formula,
\begin{eqnarray*}
    Q^{(k+1)}=\argmin_Q \sum_{i=1}^N \sum_{t=T_1}^{T_2-1} \Big[R_{i,t}+ \gamma \max_a Q^{(k)}(a, S_{i,t+1})-Q(A_{i,t},S_{i,t}) \Big]^2.
\end{eqnarray*}
Our goal is to establish the uniform rate of convergence of these estimators across different time intervals $[T_1,T_2]$, under the nonstandard stationarity assumption SA3 where the reward and transition functions can remain nonstationary over time. 

Under the more conventional stationarity assumption SA1, it is immediate to see that $Q^{(k+1)}(a,s)$ is to converge to 
\begin{eqnarray}\label{eqn:someconvergentquantity}
    g_t(a,s,Q^{(k)})\stackrel{\Delta}{=}r_t(a,s)+ \gamma \int_{s'}\max_a Q^{(k)}(a, s')p_t(s'|a,s)ds',
\end{eqnarray}
where the subscript $t$ could be removed due to the stationarity of the reward and transition functions. However, this no longer holds under SA3 where the reward and state transition functions may remain nonstationary over time. Our key insight is that although $Q^{(k+1)}(a,s)$ might not converge to \eqref{eqn:someconvergentquantity}, it converges to its weighted average with weights $w_t(a,s)=\pi_t^b(a|s)p_t^b(s)$ dependent upon the state-action visitation probability (recall that $p_t^b$ denotes the density function of $S_t$), given by
\begin{eqnarray}\label{eqn:anotherconvergentquantity}
    Q^{(k+1),*}(a,s)=\Big[\sum_{t=T_1}^{T_2-1}w_t(a,s)\Big]^{-1}\Big[\sum_{t=T_1}^{T_2-1} g_t(a,s,Q^{(k)})w_t(a,s)\Big].
\end{eqnarray}
 
Meanwhile, under SA3, the optimal Q-function $Q^{\tiny{opt}}$ is stationary and satisfies $Q^{\tiny{opt}}(a,s)=g_t(a,s,Q^{\tiny{opt}})$ for any $t$, and hence, we have
\begin{eqnarray*}
    Q^{\tiny{opt}}(a,s)= \Big[\sum_{t=T_1}^{T_2-1}w_t(a,s)\Big]^{-1}\Big[\sum_{t=T_1}^{T_2-1} g_t(a,s,Q^{\tiny{opt}})w_t(a,s)\Big].
\end{eqnarray*}
This together with \eqref{eqn:anotherconvergentquantity} enables us to apply the standard error analysis in FQI to show that
\begin{align}\label{eqn:FQIerroranalysis}
\begin{split}
\sup_{a,s} |Q^{\tiny{opt}}(a,s)-\widehat{Q}_{[T_1,T_2]}(a,s)|\le \sum_{1\le k\le K} \gamma^{K-k} \sup_{a,s} |Q^{(k),*}(a,s)-Q^{(k)}(a,s)|\\+\gamma^{K} \sup_{a,s} |Q^{\tiny{opt}}(a,s)-Q^{(0)}(a,s)|,  
\end{split}
\end{align}	
where the first term on the right-hand-side (RHS) denotes the finite-sample estimation error between the Q-function estimator $Q^{(k+1)}$ and its population-level target $Q^{(k+1),*}$, and the second term measures the bias due to initialization. 

In the following, we first conduct a population-level analysis by assuming $N=\infty$. We next conduct the finite-sample analysis to formally establish the rate of convergence. 

\noindent \textbf{Population-level analysis}. Recall that $\Sigma_{[T_1,T_2]}=(T_2-T_1)^{-1}\sum_{t=T_1}^{T_2-1} \Mean \phi_L(A_t,S_t)\phi_L^\top(A_t,S_t)$. As $N=\infty$, the regression coefficients $\beta^{(k)}$ computed at each $k$th iteration equals
\begin{eqnarray}\label{eqn:betak}
    \beta^{(k)}=\Sigma^{-1}_{[T_1,T_2]}\Big[\frac{1}{T_2-T_1}\sum_{t=T_1}^{T_2-1}\Mean \phi_L(A_{t}, S_{t}) \{r_t(A_{t},S_{t})+\gamma \max_{a}\phi_L^\top(a,S_{t+1})\beta^{(k-1)}\}\Big],
\end{eqnarray}
with $\beta^{(0)}$ initialized to a zero vector. This yields the Q-function estimator $Q^{(k)}(a,s)=\phi_L^\top(a,s) \beta^{(k)}$. 

To illustrate the rationale behind the definition of $Q^{(k),*}$ in \eqref{eqn:anotherconvergentquantity}, consider what would occur if we replace $r_t(A_{t},S_{t})+\gamma \max_{a}\phi_L^\top(a,S_{t+1})\beta^{(k-1)}$ in \eqref{eqn:betak} with $Q^{(k),*}(A_t,S_t)$. Equation \eqref{eqn:betak} remains true. In other words, we have
\begin{eqnarray}\label{eqn:betak1}
    \beta^{(k)}=\Sigma^{-1}_{[T_1,T_2]}\Big[\frac{1}{T_2-T_1}\sum_{t=T_1}^{T_2-1}\Mean \phi_L(A_{t}, S_{t}) Q^{(k),*}(A_t,S_t)\Big]. 
\end{eqnarray}
This implies that $\phi_L^\top \beta^{(k)}$ is the projection of $Q^{(k),*}$ onto the linear function class. Consequently, $Q^{(k),*}$ is indeed the population limit of $\phi_L^\top \beta^{(k)}$ in the presence of nonstationary reward and transition functions. 

To effectively control the difference between $Q^{(k)}$ and $Q^{(k)*}$, we need to approximate $Q^{(k)*}$ using the linear function class. This can be achieved using the realizability assumption in A2 and the completeness assumption in A3. Specifically, under the $p$-smoothness condition in A2, there exists some $\beta_t^*$ such that $\phi_L^\top\beta_t^*$ can approximate $r_t$ uniformly with the uniform approximation error upper bounded by $O(L^{-p/d})$ \citep[see][Section 2.2]{Huang1998}. Similarly, under A3, there exists some $\beta_t^{(k)*}$ whose $\ell_2$-norm is bounded by $\|\beta^{(k-1)}\|_2$ such that $\phi_L^\top \beta_t^{(k)*}$ can approximate $\mathcal{B}_t \phi_L^\top \beta^{(k-1)}$ uniformly with the error upper bounded by $O(L^{-p/d} \|\beta^{(k-1)}\|_2)$. This allows us to use $\phi_L^\top (\beta_t^{*}+\gamma\beta_t^{(k)*})$ to approximate $g_t(a,s,Q^{(k-1)})$, which further leads to the approximation of $Q^{(k)*}$ by $\phi_L^\top \beta^{(k)*}$ where
\begin{eqnarray*}
    \beta^{(k)*}=\Big[\sum_{t=T_1}^{T_2-1}w_t(a,s)\Big]^{-1}\Big[\sum_{t=T_1}^{T_2-1} (\beta_t^{*}+\gamma\beta_t^{(k)*}) w_t(a,s)\Big],
\end{eqnarray*}
such that
\begin{eqnarray*}
    \textrm{err}(a,s)\stackrel{\Delta}{=}|\phi_L^\top(a,s) \beta^{(k)*}-Q^{(k)*}(a,s)|=O\{L^{-p/d}(1+\|\beta^{(k-1)}\|_2)\},
\end{eqnarray*}
where the big-$O$ term is uniform in $(a,s)$. It follows from Equation \eqref{eqn:betak1} that 
\begin{eqnarray}\label{eqn:betak-betakstarpopulation0}
    \beta^{(k)}-\beta^{(k)*}=\frac{1}{T_2-T_1}\sum_{t=T_1}^{T_2-1}\Mean \Sigma^{-1}_{[T_1,T_2]} \phi_L(A_{t}, S_{t})\{Q^{(k),*}(A_{t},S_{t})-\phi_L^\top(A_{t}, S_{t})\beta^{(k)*}\},
\end{eqnarray}
and hence, $\|\beta^{(k)}-\beta^{(k)*}\|_2$ can be upper bounded by
\begin{eqnarray*}
    \sup_{\theta\in \mathbb{R}^{mL}:\|\theta\|_2=1} \Big|\frac{1}{T_2-T_1}\sum_{t=T_1}^{T_2-1}\Mean \theta^\top\Sigma^{-1}_{[T_1,T_2]} \phi_L(A_{t}, S_{t})\{Q^{(k),*}(A_{t},S_{t})-\phi_L^\top(A_{t}, S_{t})\beta^{(k)*}\}\Big|\\
    \le \sup_{a,s}\textrm{err}(a,s) \times \sup_{\theta\in \mathbb{R}^{mL}:\|\theta\|_2=1}\frac{1}{T_2-T_1}\sum_{t=T_1}^{T_2-1}\Mean |\theta^\top\Sigma^{-1}_{[T_1,T_2]} \phi_L(A_{t}, S_{t})|
\end{eqnarray*}
Using Cauchy-Schwarz inequality, the second term in the last line can be upper bounded by
\begin{eqnarray*}
    \sup_{\theta\in \mathbb{R}^{mL}:\|\theta\|_2=1}\sqrt{\frac{1}{T_2-T_1}\sum_{t=T_1}^{T_2-1}\Mean |\theta^\top\Sigma^{-1}_{[T_1,T_2]} \phi_L(A_{t}, S_{t})|^2}=\sup_{\theta\in \mathbb{R}^{mL}:\|\theta\|_2=1} \sqrt{\theta^\top \Sigma_{[T_1,T_2]}^{-1}\theta}.
\end{eqnarray*}
Under A5(ii), the minimum eigenvalue of $\Sigma_{[T_1,T_2]}$ is bounded away from zero, which in turn suggests that the maximum eigenvalue of $\Sigma_{[T_1,T_2]}^{-1}$ is bounded away from infinity. It follows that the above expression is of the order $O(1)$, and hence
\begin{eqnarray}\label{eqn:betak-betakstarpopulation}
    \|\beta^{(k)}-\beta^{(k)*}\|_2=O\{L^{-p/d}(1+\|\beta^{(k-1)}\|_2)\}.
\end{eqnarray}

Next, we focus on providing an upper bound for $\sup_k \|\beta^{(k)}\|_2$ to further simplify the RHS of \eqref{eqn:betak-betakstarpopulation}. To begin with, we first provide an uniform upper bound for $\sup_t \|\beta_t^*\|_2$. Notice that 
\begin{eqnarray}\label{eqn:betatstarbound}
\begin{split}
    \beta_t^*=\Sigma^{-1}_{[t,t+1]} \Sigma_{[t,t+1]}\beta^*_t=\Sigma^{-1}_{[t,t+1]} \Mean [\phi_L(A_{t},S_{t})r_t(A_t,S_t)]\\-\Sigma^{-1}_{[t,t+1]} \Mean \phi_L(A_{t},S_{t})[r_t(A_t,S_t)-\phi_L^\top(A_t,S_t)\beta^*_t].
\end{split}
\end{eqnarray}
As $r_t$s are uniformly bounded (implied by A2), using similar arguments to the proof of \eqref{eqn:betak-betakstarpopulation}, it can be shown that both the rightmost term in the first line of \eqref{eqn:betatstarbound} and the term in the second line of \eqref{eqn:betatstarbound} are upper bounded by $O(1)$ and $O(L^{p/d})$, respectively. Consequently, we have $\sup_t \|\beta_t^*\|_2=O(1)$. This together with the definition of $\beta^{(k)*}$ yields that $\|\beta^{(k)*}\|_2\le C+\gamma \|\beta^{(k-1)}\|_2$ for some constant $C>0$. It follows from \eqref{eqn:betak-betakstarpopulation} that
\begin{eqnarray*}
    \|\beta^{(k)}\|_2\le \bar{C}+\gamma (1+\bar{C}L^{-p/d}) \|\beta^{(k-1)}\|_2,
\end{eqnarray*}
for some constant $\bar{C}>0$. 
As $\beta^{(0)}$ is initialized to a zero vector, we obtain that $\sup_k \|\beta^{(k)}\|_2\le c(1-\gamma)^{-1}$ for some constant $c>0$. It follows that the uppper bound in \eqref{eqn:betak-betakstarpopulation} can be simplified into
\begin{eqnarray}\label{eqn:betak-betakstarpopulation1}
    \|\beta^{(k)}-\beta^{(k)*}\|_2=O\Big(\frac{L^{-p/d}}{1-\gamma}\Big).
\end{eqnarray}

Finally, we focus on providing an upper bound for $\sup_{a,s}|\phi_L^\top(a,s) (\beta^{(k)}-\beta^{(k)*})|$. A na{\"i}ve approach is to employ the Cauchy-Schwarz inequality to show that
\begin{eqnarray*}
    \sup_{a,s} |\phi_L^\top(a,s) (\beta^{(k)}- \beta^{(k)*})|\le \sup_{a,s} \|\phi_L^\top(a,s)\|_2 \|\beta^{(k)}- \beta^{(k)*}\|_2=O\Big(\frac{L^{1/2-p/d}}{1-\gamma}\Big),
\end{eqnarray*}
by \eqref{eqn:basis2} and \eqref{eqn:betak-betakstarpopulation1}. However, this bound is not sharp. In the following, we conduct a refined analysis to show that the upper bound can be sharpened to
\begin{eqnarray}\label{eqn:phibetak-betakstarpopulation}
    \sup_{a,s} |\phi_L^\top(a,s) (\beta^{(k)*}-\beta^{(k)})|=O\Big(\frac{L^{-p/d}}{1-\gamma}\Big),
\end{eqnarray}
which is faster by a factor of $L^{-1/2}$. This together with the upper bound on the approximation error $\sup_{a,s} |Q^{(k),*}(a,s)-\phi_L^\top(a,s) \beta^{(k)*}|=O((1-\gamma)^{-1}L^{-p/d})$ yields the following error bound for the difference between $Q^{(k)}$ and its population limit $Q^{(k),*}$,
\begin{eqnarray}\label{eqn:Qkstar-Qkpopulation}
    \sup_{a,s} |Q^{(k),*}(a,s)-Q^{(k)}(a,s)|=O\Big(\frac{L^{-p/d}}{1-\gamma}\Big).
\end{eqnarray}
Notice that the boundedness of $r_t$s yields that $\sup_{a,s}|Q^{\tiny{opt}}(a,s)|=O((1-\gamma)^{-1})$. As the initial Q-estimator is initialized to zero, it follows from \eqref{eqn:Qkstar-Qkpopulation} and the error analysis in \eqref{eqn:FQIerroranalysis} that
\begin{eqnarray*}
    \sup_{a,s} |Q^{\tiny{opt}}(a,s)-Q^{(k)}(a,s)|=O\Big(\frac{L^{-p/d}}{(1-\gamma)^2}\Big)+O\Big(\frac{\gamma^k}{1-\gamma}\Big). 
\end{eqnarray*}
By letting $k\to \infty$, we obtain \eqref{eqn:basis1}. 

It remains to prove \eqref{eqn:phibetak-betakstarpopulation}. The main idea is to employ the bias control techniques developed by \citet{huang2003local} (see Lemma 5.1 and Theorem A.1 therein), based on which we can show that the basis function $\phi_L$ satisfies
\begin{eqnarray}\label{eqn:ellinfinitynorm}
    \Big[\sup_{a,s} |h(A_t,S_t)|\Big]^{-1}\sup_{a,s} \Big|\phi_L^\top(a,s) \frac{\Sigma_{[T_1,T_2]}^{-1}}{T_2-T_1}\sum_{t=T_1}^{T_2-1}\Mean \phi_L(A_t,S_t)h(A_t,S_t)\Big|=O(1),
\end{eqnarray}
where the big-$O$ term on the RHS is uniform in any nonzero function $h$. 

Notice that our setting differs from the one considered in \citet{huang2003local} in two ways: (i) the basis function is not only a function of the continuous state, but a function of the discrete action as well; (ii) the state-action pairs at different times may have nonstationary distribution functions. Despite these differences, \eqref{eqn:ellinfinitynorm} remains valid as (i) we estimate the coefficients for different actions separately using different data subsets grouped by the action; (ii) we can treat the data as if it were sampled from a stationary distribution, in which each state-action pair follows a mixture of distributions of $\{(S_t,A_t):T_1\le t<T_2-1\}$with equal weights. 

Now, in view of \eqref{eqn:betak-betakstarpopulation0},  \eqref{eqn:phibetak-betakstarpopulation} can be readily obtained by setting the function $h$ in \eqref{eqn:ellinfinitynorm} to $\textrm{err}$. This completes the population-level analysis. 

\noindent \textbf{Finite-sample analysis}. Next, we proceed to conduct the finite-sample analysis by considering a finite $N$. The arguments are very similar to those used in the population-level analysis. The difference lies in that, with a finite $N$, the equality in \eqref{eqn:betak1} no longer holds. Rather, the estimator $\beta^{(k)}$ is defined by 
\begin{eqnarray}\label{eqn:betakdefinition}
    \frac{\widehat{\Sigma}_{[T_1,T_2]}^{-1}}{N(T_2-T_1)}\Big[\sum_{i=1}^N\sum_{t=T_1}^{T_2-1} \phi_L(A_{i,t},S_{i,t}) \{R_{i,t}+\gamma \max_{a'}\phi_L^\top(a',S_{i,t+1})\beta^{(k-1)}\}\Big],
\end{eqnarray}
where we recall that 
\begin{eqnarray*}
    \widehat{\Sigma}_{[T_1,T_2]}=\frac{1}{N(T_2-T_1)} \sum_{i=1}^N\sum_{t=T_1}^{T_2-1} \phi_L(A_{i,t},S_{i,t})\phi_L^\top(A_{i,t},S_{i,t}).
\end{eqnarray*}
As such, we have
\begin{eqnarray}\label{betak-betakstar}
\begin{split}
    \beta^{(k)}-\beta^{(k)*}&=\widehat{\Sigma}_{[T_1,T_2]}^{-1}
    \Big[\frac{1}{N(T_2-T_1)} \sum_{i=1}^N\sum_{t=T_1}^{T_2-1} \phi_L(A_{i,t},S_{i,t})\{R_{i,t}\\&+\gamma\max_{a'}\phi_L^\top(a',S_{i,t+1})\beta^{(k-1)}-\phi_L(A_{i,t},S_{i,t}) \beta^{(k)*} \}  \Big],
\end{split}
\end{eqnarray}
where the population limit $\beta^{(k)*}$ is defined in the population-level analysis.

Using similar arguments to the proof of Lemma \ref{lemmamatrixnonstat}, we can show that 
\begin{eqnarray}\label{eqn:set0}
\begin{aligned}
	\Big\|\widehat{\Sigma}_{[T_1,T_2]}-\Sigma_{[T_1,T_2]} \Big\|_2 
	\le c \sqrt{L (\epsilon NT)^{-1} \log (NT)},  
\end{aligned}	
\end{eqnarray}
for some constant $c>0$, where the inequality holds uniformly across all pairs $(T_1,T_2)$ with probability at least $1- O(N^{-1} T^{-1})$. Under A5(ii), $\lambda_{\min}[\Sigma_{[T_1,T_2]}]$ is uniformly bounded away from zero. On the event set defined by \eqref{eqn:set0}, for sufficiently large $NT$, there exists some $\bar{c}>0$ such that $\lambda_{\min}[\widehat{\Sigma}_{[T_1,T_2]}]\ge \bar{c}$. 
This together with \eqref{betak-betakstar} yields that
\begin{eqnarray}\label{eqn:betak-betakstar}
\begin{split}
    \|\beta^{(k)}-\beta^{(k)*}\|_2\le \bar{c}^{-1} \Big\| \sum_{i=1}^N\sum_{t=T_1}^{T_2-1} \frac{\phi_L(A_{i,t},S_{i,t})}{N(T_2-T_1)}\{R_{i,t}+\gamma\max_{a'}\phi_L^\top(a',S_{i,t+1})\beta^{(k-1)}\\-\phi_L^\top(A_{i,t},S_{i,t}) \beta^{(k)*} \}  \Big\|_2.
\end{split}
\end{eqnarray}
Assume for now,
\begin{eqnarray}\label{eqn:betakbetakstarupperbound}
    \sup_k \|\beta^{(k)}\|_2\le \frac{c}{1-\gamma}\,\,\hbox{and}\,\,\sup_k \|\beta^{(k)*}\|_2\le \frac{c}{1-\gamma},
\end{eqnarray}
for some constant $c>0$. These assertions can be proven using similar arguments to the population-level analysis. 

By definition, if $\beta^{(k-1)}$ and $\beta^{(k)*}$ were computed using an independent dataset, the expected value of the vector on the RHS of \eqref{eqn:betak-betakstar} would be of the order $O((1-\gamma)^{-1}L^{-p/d})$, based on our population-level analysis. However, since they are computed by the same dataset, they become dependent upon each other. To handle such dependence, we upper bound the $\ell_2$-norm in the RHS of \eqref{eqn:betak-betakstar} by
\begin{eqnarray}\label{eqn:upperbound0}
    \sup_{\beta_1,\beta_2}\Big\| \sum_{i=1}^N\sum_{t=T_1}^{T_2-1} \frac{\phi_L(A_{i,t},S_{i,t})}{N(T_2-T_1)}\{R_{i,t}+\gamma\max_{a'}\phi_L^\top(a',S_{i,t+1})\beta_1-\phi_L^\top(A_{i,t},S_{i,t}) \beta_2 \}  \Big\|_2
\end{eqnarray}
where the supremum is taken over all the pairs $(\beta_1,\beta_2)$ such that $\|\beta_1\|_2\le (1-\gamma)^{-1}c$, $\|\beta_2\|_2\le (1-\gamma)^{-1}c$ and that
\begin{eqnarray*}
    \Big|\frac{1}{T_2-T_1}\Mean \Big[\sum_{t=T_1}^{T_2-1} \phi_L(A_{t},S_{t})\{R_{t}+\gamma\max_{a'}\phi_L^\top(a',S_{t+1})\beta_1-\phi_L^\top(A_{t},S_{t}) \beta_2 \}\Big]\Big|_2=O\Big(\frac{L^{-p/d}}{1-\gamma}\Big),
\end{eqnarray*}
as the above equation would have been satisfied if we were to set $\beta_1=\beta^{(k-1)}$ and $\beta_2=\beta^{(k)*}$. Using similar arguments to Step 3 of the proof of Theorem \ref{thm:size}, we can show that \eqref{eqn:upperbound0} is upper bounded by $O((1-\gamma)^{-1}\sqrt{L(\epsilon NT)^{-1}\log (NT)})+O((1-\gamma)^{-1}L^{-p/d})$ where the big-$O$ terms are uniform in $(T_1,T_2)$, with probability at least $1-O((NT)^{-1})$. 

This together with \eqref{eqn:betak-betakstar} yields the following upper bound for the difference $\beta^{(k)*}-\beta^{(k)}$, 
\begin{eqnarray*}
    \|\beta^{(k)*}-\beta^{(k)}\|_2= O\Big(\frac{\sqrt{L(\epsilon NT)^{-1}\log (NT)}}{1-\gamma}\Big)+O\Big(\frac{L^{-p/d}}{1-\gamma}\Big).
\end{eqnarray*}
Notice that compared to the error bound in the population-level analysis \eqref{eqn:betak-betakstarpopulation}, the RHS above includes an additional term $O((1-\gamma)^{-1}\sqrt{L(\epsilon NT)^{-1}\log(NT)})$, which accounts for the finite-sample estimation error. 

Finally, to upper bound $\sup_{a,s}|\phi_L^\top(a,s)(\beta^{(k)*}-\beta^{(k)})|$, we define two intermediate quantities
\begin{eqnarray*}
    I_1=\frac{\Sigma_{[T_1,T_2]}^{-1}}{N(T_2-T_1)} \sum_{i=1}^N\sum_{t=T_1}^{T_2-1} \phi_L(A_{i,t},S_{i,t})\{R_{i,t}+\gamma\max_{a'}\phi_L^\top(a',S_{i,t+1})\beta^{(k-1)}-\phi_L(A_{i,t},S_{i,t}) \beta^{(k)*} \},\\
    I_2=\frac{\Sigma_{[T_1,T_2]}^{-1}}{T_2-T_1} \sum_{t=T_1}^{T_2-1} \Mean \phi_L(A_{t},S_{t})\{R_{t}+\gamma\max_{a'}\phi_L^\top(a',S_{t+1})\beta^{(k-1)}-\phi_L(A_{t},S_{t}) \beta^{(k)*} \}.
\end{eqnarray*}
It follows from \eqref{eqn:set0} that
\begin{eqnarray*}
    \|\beta^{(k)*}-\beta^{(k)}-I_1\|_2=O\Big(\frac{ L(\epsilon NT)^{-1}\log (NT)}{(1-\gamma)^2}\Big)+O\Big(\frac{L^{1/2-p/d}\sqrt{(\epsilon NT)^{-1}\log (NT)}}{(1-\gamma)^2}\Big),
\end{eqnarray*}
with probability at least $1-O(N^{-1}T^{-1})$. 

Meanwhile, using similar arguments to bounding the $\ell_2$ norm in the RHS of \eqref{eqn:betak-betakstar}, we obtain with probability at least $1-O(N^{-1}T^{-1})$ that
\begin{eqnarray*}
    \|I_1-I_2\|_\infty=O\Big(\frac{\sqrt{L(\epsilon NT)^{-1}\log (NT)}}{1-\gamma}\Big).
\end{eqnarray*}
Combining these bounds together with \eqref{eqn:basis2.5}, we obtain that
\begin{eqnarray*}
    \sup_{a,s}|\phi_L^\top(a,s)(\beta^{(k)*}-\beta^{(k)})|\le \sup_{a,s}\|\phi_L(a,s)\|_2 \|\beta^{(k)*}-\beta^{(k)}-I_1\|_2+\sup_{a,s}\|\phi_L(a,s)\|_1 \|I_1-I_2\|_\infty\\
    +\sup_{a,s} |\phi_L^\top(a,s)I_2|=O\Big(\frac{\sqrt{L(\epsilon NT)^{-1}\log (NT)}}{1-\gamma}\Big)+\sup_{a,s} |\phi_L^\top(a,s)I_2|,
\end{eqnarray*}
under the conditions on $L$ in A9. 

Finally, it follows from the arguments used to the proof of \eqref{eqn:phibetak-betakstarpopulation} in the population-level analysis that $\sup_{a,s} |\phi_L^\top(a,s)I_2|=O((1-\gamma)^{-1}L^{-p/d})$. Consequently, we have
\begin{eqnarray*}
    \sup_{a,s}|\phi_L^\top(a,s)(\beta^{(k)*}-\beta^{(k)})|=O\Big(\frac{\sqrt{L(\epsilon NT)^{-1}\log (NT)}}{1-\gamma}\Big)+O\Big(\frac{L^{-p/d}}{1-\gamma}\Big),
\end{eqnarray*}
and hence $\sup_{a,s}|Q^{(k)*}(a,s)-Q^{(k)}(a,s)|$ is of the same order of magnitude. It follows from the error analysis in \eqref{eqn:FQIerroranalysis} that
\begin{eqnarray*}
    \sup_{a,s} |Q^{\tiny{opt}}(a,s)-Q^{(k)}(a,s)|=O\Big(\frac{\sqrt{L(\epsilon NT)^{-1}\log (NT)}}{(1-\gamma)^2}\Big)+O\Big(\frac{L^{-p/d}}{(1-\gamma)^2}\Big)+O\Big(\frac{\gamma^k}{1-\gamma}\Big),
\end{eqnarray*}
with probability at least $1-O(N^{-1}T^{-1})$. This establishes the rate of convergence in \eqref{eqn:rateofconvergence} by noting that the number of FQI iterations much larger than $\log (NT)$ (see Assumption A9). 

To conclude this part, notice that
\begin{eqnarray*}
    \beta^{(k)}-\beta^*= \frac{\Sigma_{[T_1,T_2]}^{-1}}{T_2-T_1}\sum_{t=T_1}^{T_2-1} \Mean \phi_L(A_t,S_t)\phi_L^\top(A_t,S_t)(\beta^{(k)}-\beta^*)\\
    =\frac{\Sigma_{[T_1,T_2]}^{-1}}{T_2-T_1}\sum_{t=T_1}^{T_2-1} \Mean \phi_L(A_t,S_t)\{\phi_L^\top(A_t,S_t)\beta^{(k)}-Q^{\tiny{opt}}(A_t,S_t)\}\\
    +\frac{\Sigma_{[T_1,T_2]}^{-1}}{T_2-T_1}\sum_{t=T_1}^{T_2-1} \Mean \phi_L(A_t,S_t)\{Q^{\tiny{opt}}(A_t,S_t)-\phi_L^\top(A_t,S_t)\beta^*\}.
\end{eqnarray*}
Based on the established rate of convergence and \eqref{eqn:basis1}, one can similarly prove that
\begin{eqnarray}\label{eqn:betaK-betastarell2bound}
    \|\beta^{(k)}-\beta^*\|_2=O\Big(\frac{L^{-p/d}}{(1-\gamma)^2}\Big)+O\Big(\frac{\sqrt{L(\epsilon NT)^{-1}\log(NT)}}{(1-\gamma)^2}\Big)+O\Big(\frac{\gamma^k}{1-\gamma}\Big),
\end{eqnarray}
We omit the detail proof to save space. 

\noindent \textbf{Asymptotic analysis}. To begin with, we notice that  
\begin{eqnarray*}
    \beta^*=\frac{\Sigma^{-1}_{[T_1,T_2]}}{T_2-T_1} \Mean \Big[\sum_{t=T_1}^{T_2-1} \phi_L(A_t,S_t)\phi_L^\top(A_t,S_t) \beta^*\Big]=\frac{\Sigma^{-1}_{[T_1,T_2]}}{T_2-T_1} \Mean \Big[\sum_{t=T_1}^{T_2-1} \phi_L(A_t,S_t)Q^*(A_t,S_t)\Big]\\
    -\underbrace{\frac{\Sigma^{-1}_{[T_1,T_2]}}{T_2-T_1} \Mean \Big[\sum_{t=T_1}^{T_2-1} \phi_L(A_t,S_t)\textrm{reminder}(A_t,S_t)\Big]}_{\theta_{r}^{(1)}},
\end{eqnarray*}
where the reminder function is bounded by $O((1-\gamma)^{-2}L^{-p/d})$, due to \eqref{eqn:basis1}. This together with the Bellman optimality equation $Q^*=r_t+\gamma \mathcal{B}_t Q^*$ implies that $\beta^*$ is equal to
\begin{eqnarray*}
    \underbrace{\frac{\Sigma^{-1}_{[T_1,T_2]}}{T_2-T_1} \Mean \Big[\sum_{t=T_1}^{T_2-1} \phi_L(A_t,S_t)r_t(A_t,S_t)\Big]}_{\beta_r}+ \frac{\gamma\Sigma^{-1}_{[T_1,T_2]}}{T_2-T_1} \Mean \Big[\sum_{t=T_1}^{T_2-1} \phi_L(A_t,S_t) Q^{\tiny{opt}}(\pi^{\tiny{opt}}(S_{t+1}),S_{t+1})\Big]-\theta_r^{(1)},
\end{eqnarray*}
and hence,
\begin{eqnarray*}
    \beta^*=\beta_r+\gamma\underbrace{\frac{\Sigma^{-1}_{[T_1,T_2]}}{T_2-T_1} \Mean \Big[\sum_{t=T_1}^{T_2-1} \phi_L(A_t,S_t) \phi_L^\top(\pi^{\tiny{opt}}(S_{t+1}),S_{t+1})\Big]}_{\mathcal{P}}\beta^*-\theta_r^{(1)}\\
    +\gamma \underbrace{\frac{\Sigma^{-1}_{[T_1,T_2]}}{T_2-T_1} \Mean \Big[\sum_{t=T_1}^{T_2-1} \phi_L(A_t,S_t) \textrm{reminder}(\pi^{\tiny{opt}}(S_{t+1}),S_{t+1})\Big]}_{\theta_r^{(2)}}.
\end{eqnarray*}
Using the same arguments, we obtain that $\beta^*=\beta_r +\gamma \mathcal{P} \beta_r +\gamma^2 \mathcal{P}^2 \beta^*-\theta_r-\mathcal{P}\theta_r$ 
where $\theta_r=\theta_r^{(1)}-\gamma \theta_r^{(2)}$. Repeating this process, we obtain that $\beta^*=\sum_{j=1}^k (\gamma \mathcal{P})^{j-1}\beta_r +\gamma^k \mathcal{P}^k \beta^*-\sum_{j=1}^{k-1} (\gamma \mathcal{P})^{j} \theta_r$ and hence 
\begin{eqnarray}\label{eqn:phibetastariterative1}
    \phi_L^\top(a,s)\beta^*=\sum_{j=1}^k \gamma^{j-1}\phi_L^\top(a,s)\mathcal{P}^{j-1}\beta_r +\gamma^k \phi_L^\top(a,s)\mathcal{P}^k \beta^*-\sum_{j=1}^{k-1} \gamma^j \phi_L^\top(a,s) \mathcal{P}^{j} \theta_r
\end{eqnarray}

Below, we focus on providing an upper bound on the rightmost reminder term in \eqref{eqn:phibetastariterative1}. First, as the reminder function is of the order $O((1-\gamma)^{-2}L^{-p/d})$, it follows from \eqref{eqn:ellinfinitynorm} that $\sup_{a,s} |\phi_L^\top(a,s) \theta_r^{(1)}|=O((1-\gamma)^{-2}L^{-p/d})$. Second, when upper bounding the function $\phi_L^\top(a,s) \theta_r^{(2)}$, we notice that it can be represented as
\begin{eqnarray}\label{eqn:phithetar2}
    \phi_L^\top(a,s)\frac{\Sigma^{-1}_{[T_1,T_2]}}{T_2-T_1} \Mean \Big[\sum_{t=T_1}^{T_2-1} \phi_L(A_t,S_t) (\mathcal{B}_t^*\textrm{reminder})(A_t,S_{t})\Big],
\end{eqnarray}
where $\mathcal{B}_t^*$ is defined such that $\mathcal{B}_t^*g(a,s)=\int_{s'} g(\pi^{\tiny{opt}}(s'),s')p_t(s'|a,s)ds'$. 

However, $\mathcal{B}_t\textrm{reminder}$ can vary over time, as the transition function may not be stationary. As a result, \eqref{eqn:ellinfinitynorm} is not directly applicable to upper bound \eqref{eqn:phithetar2}. Nonetheless, we may apply the techniques in \eqref{eqn:betak1} to stationarize the reminder term in \eqref{eqn:phithetar2}. Specifically, define
\begin{eqnarray*}
    \textrm{reminder}^*(a,s)= \Big[\sum_{t=T_1}^{T_2-1} w_t(a,s)\Big]^{-1} \Big[\sum_{t=T_1}^{T_2-1} w_t(a,s) (\mathcal{B}_t\textrm{reminder})(a,s)\Big]
\end{eqnarray*}
With some calculations, it can be shown that \eqref{eqn:phithetar2} is equal to
\begin{eqnarray*}
    \phi_L^\top(a,s)\frac{\Sigma^{-1}_{[T_1,T_2]}}{T_2-T_1} \Mean \Big[\sum_{t=T_1}^{T_2-1} \phi_L(A_t,S_t) \textrm{reminder}^*(A_t,S_{t})\Big],
\end{eqnarray*}
with $\mathcal{B}_t\textrm{reminder}$ being replaced by a stationary function $\textrm{reminder}^*$ so that we can use \eqref{eqn:ellinfinitynorm} to show that $\sup_{a,s} |\phi_L^\top(a,s) \theta_r^{(2)}|=O((1-\gamma)^{-2}L^{-p/d})$. Third, we will show below that
\begin{eqnarray}\label{eqn:operatorgammaSigmaPSigma}
    \sup_{\beta\in\mathbb{R}^{mL}:\|\beta\|_2=1} \|\gamma\Sigma^{1/2}_{[T_1,T_2]}\mathcal{P}\Sigma^{-1/2}_{[T_1,T_2]} \beta\|_2 \le \rho=\sqrt{1-c(1-\gamma)},
\end{eqnarray}
for some constant $c>0$. This suggests that the operator norm of $\gamma\Sigma^{1/2}_{[T_1,T_2]}\mathcal{P}\Sigma^{-1/2}_{[T_1,T_2]}$ is strictly smaller than $1$, which together with the boundedness of $\|\Sigma_{[T_1,T_2]}^{-1/2}\|_2$ (implied by A5(ii)) yields
\begin{eqnarray}\label{eqn:gammajPjthetar}
    \|\gamma^{j}\mathcal{P}^j \theta_r\|_2\le \|\Sigma_{[T_1,T_2]}^{-1/2}\|_2 \|\gamma\Sigma^{1/2}_{[T_1,T_2]}\mathcal{P} \Sigma^{-1/2}_{[T_1,T_2]}\|_2^j \|\Sigma^{1/2}_{[T_1,T_2]}\theta_r\|_2\le C \rho^j \|\Sigma^{1/2}_{[T_1,T_2]}\theta_r\|_2.
\end{eqnarray}
Meanwhile, using similar arguments to the proof of \eqref{eqn:betak-betakstarpopulation}, we can show that $\|\Sigma^{1/2}_{[T_1,T_2]}\theta_r\|_2=O((1-\gamma)^{-2}L^{-p/d})$, which together with \eqref{eqn:gammajPjthetar} yields that $\|\gamma^j \mathcal{P}^j \theta_r\|_2\le \bar{c} (1-\gamma)^{-2} \rho^j L^{-p/d}$ for some constant $c>0$ and any $j\ge 0$. Consequently, for each $j$, we can represent $\gamma^j \phi_L^\top(a,s) \mathcal{P}^{j} \theta_r$ as $\gamma \phi_L^\top(a,s) \mathcal{P} \theta_{r,j}$, i.e.,
\begin{eqnarray}\label{eqn:phiPthetarj}
    \gamma \phi_L^\top(a,s) \frac{\Sigma^{-1}_{[T_1,T_2]}}{T_2-T_1} \Mean \Big[\sum_{t=T_1}^{T_2-1} \phi_L(A_t,S_t) \mathcal{B}_t^*\phi_L^\top(A_t, S_{t})\theta_{r,j}\Big]
\end{eqnarray}
for some $\theta_{r,j}$ whose $\ell_2$-norm is bounded by $\bar{c} (1-\gamma)^{-2} \rho^{j-1} L^{-p/d}$. As $p_t$s are bounded, it follows from \eqref{eqn:basis2} that $\sup_{a,s}\|\mathcal{B}_t^*\phi_L^\top(a,s)\|_2=O(1)$. Using similar arguments in bounding $\theta_L^\top \theta_r^{(2)}$, we can show that \eqref{eqn:phiPthetarj} is of the order $O((1-\gamma)^{-2} \rho^{j-1} L^{-p/d})$. 

Hence, the rightmost reminder term in \eqref{eqn:phibetastariterative1} is of the order
\begin{eqnarray}\label{eqn:betastarreminder}
    O\Big(\frac{L^{-p/d}}{(1-\gamma)^2(1-\rho)}\Big)=O\Big(\frac{L^{-p/d}}{(1-\gamma)^3}\Big),
\end{eqnarray}
since 
\begin{eqnarray*}
    \frac{1}{1-\rho}=\frac{1+\sqrt{1-c(1-\gamma)}}{[1-\sqrt{1-c(1-\gamma)}][1+\sqrt{1-c(1-\gamma)}]}=\frac{1+\sqrt{1-c(1-\gamma)}}{c(1-\gamma)}\le \frac{2}{c(1-\gamma)}.
\end{eqnarray*}

It remains to prove \eqref{eqn:operatorgammaSigmaPSigma}. Toward that end, notice that the left-hand-side of \eqref{eqn:operatorgammaSigmaPSigma} can be represented by
\begin{eqnarray*}
    \sup_{\substack{\beta_1\in\mathbb{R}^{mL}:\|\beta_1\|_2=1\\ \beta_2\in\mathbb{R}^{mL}:\|\beta_2\|_2=1}} \Big\|\gamma\beta_1^\top\frac{\Sigma^{-1/2}_{[T_1,T_2]}}{T_2-T_1}\sum_{t=T_1}^{T_2-1}\Mean \phi_L(A_t,S_t)\phi_L^\top(\pi^{\tiny{opt}}(S_{t+1}),S_{t+1})\Sigma^{-1/2}_{[T_1,T_2]} \beta_2\Big\|_2.
\end{eqnarray*}
It follows from the Cauchy-Schwarz inequality that the $\ell_2$ norm of in the above expression can be upper bounded by
\begin{eqnarray}\nonumber
    &\displaystyle\frac{\gamma}{T_2-T_1} \sum_{t=T_1}^{T_2-1} \Mean |\beta_1^\top \Sigma^{-1/2}_{[T_1,T_2]} \phi_L(A_t,S_t)||\phi_L^\top(\pi^{\tiny{opt}}(S_{t+1}),S_{t+1})\Sigma^{-1/2}_{[T_1,T_2]} \beta_2|\\\nonumber
    &\displaystyle\le \gamma \sqrt{\frac{1}{T_2-T_1}\sum_{t=T_1}^{T_2-1} \Mean | \beta_1^\top \Sigma^{-1/2}_{[T_1,T_2]} \phi_L(A_t,S_t)|^2}\\\nonumber &\displaystyle\times \sqrt{\frac{1}{T_2-T_1}\sum_{t=T_1}^{T_2-1} \Mean |\phi_L^\top(\pi^{\tiny{opt}}(S_{t+1}),S_{t+1})\Sigma^{-1/2}_{[T_1,T_2]} \beta_2|^2}.
\end{eqnarray}
By the definition of $\Sigma_{[T_1,T_2]}$, the second line equals $\gamma$. Hence, to prove \eqref{eqn:operatorgammaSigmaPSigma}, it suffices to show the maximum eigenvalue of the matrix 
\begin{eqnarray}\label{eqn:maximumeigenmatrix}
    \frac{\gamma^2}{T_2-T_1}\Sigma^{-1/2}_{[T_1,T_2]} \sum_{t=T_1}^{T_2-1} \Big[\Mean \phi_L(\pi^{\tiny{opt}}(S_{t+1}),S_{t+1})\phi_L^\top(\pi^{\tiny{opt}}(S_{t+1}),S_{t+1})\Big] \Sigma^{-1/2}_{[T_1,T_2]}
\end{eqnarray}
is strictly smaller than $1$, or equivalently, the matrix
\begin{eqnarray*}
    \frac{\Sigma^{-1/2}_{[T_1,T_2]}}{T_2-T_1} \sum_{t=T_1}^{T_2-1} \Big[\Mean \phi_L(A_t,S_t)\phi_L^\top(A_t,S_t) -\gamma^2\Mean \phi_L(\pi^{\tiny{opt}}(S_{t+1}),S_{t+1})\phi_L^\top(\pi^{\tiny{opt}}(S_{t+1}),S_{t+1})\Big] \Sigma^{-1/2}_{[T_1,T_2]}
\end{eqnarray*}
is positive semi-definite. The latter condition is implied by \eqref{eqn:basis2}, the boundedness of $p_t$ and A5(ii), which suggest that (i) the maximum eigenvalue of $\Sigma_{[T_1,T_2]}$ is bounded away from infinity, or equivalently, the minimum eigenvalue of $\Sigma^{-1}_{[T_1,T_2]}$ is bounded away from zero, and (ii) the minimum eigenvalue of the matrix
\begin{eqnarray*}
    \frac{1}{T_2-T_1}\sum_{t=T_1}^{T_2-1} \Big[\Mean \phi_L(A_t,S_t)\phi_L^\top(A_t,S_t)-\gamma^2\Mean \phi_L(\pi^{\tiny{opt}}(S_{t+1}),S_{t+1})\phi_L^\top(\pi^{\tiny{opt}}(S_{t+1}),S_{t+1})\Big]
\end{eqnarray*}
is lower bounded by $C(1-\gamma)$ for some constant $C>0$. Consequently, the maximum eigenvalue of \eqref{eqn:maximumeigenmatrix} is upper bounded by $1-c(1-\gamma)$
for some constant $c>0$. This completes the proof of  \eqref{eqn:operatorgammaSigmaPSigma}. 

To summarize, so far, we have shown that
\begin{eqnarray}\label{eqn:phibetastariterative2}
     \phi_L^\top(a,s)\beta^*=\sum_{j=1}^k \gamma^{j-1}\phi_L^\top(a,s)\mathcal{P}^{j-1}\beta_r +\gamma^k \phi_L^\top(a,s)\mathcal{P}^k \beta^*+O\Big(\frac{L^{-p/d}}{(1-\gamma)^3}\Big),
\end{eqnarray}
by \eqref{eqn:phibetastariterative1} and \eqref{eqn:betastarreminder}.  

Next, we aim to obtain a similar expression for $\beta^{(k)}$. Let $m_0$ denote the margin defined in A7. Under the condition on $\epsilon$ in A1 and that on $L$ in A9, it follows from the uniform rate of convergence derived in the finite-sample analysis and the condition $K\gg \log(NT)$ that
\begin{eqnarray}\label{eqnevent1}
    \prob\Big(\max_{k\ge K/2}\max_{(T_1,T_2):T_2-T_1\ge \epsilon T}\sup_{a,s} |\phi_L^\top(a,s) \beta^{(k)}-Q^{opt}(a,s)|\le \frac{m_0}{3}\Big)\ge 1-O\Big(\frac{1}{NT}\Big). 
\end{eqnarray}
Under the event defined in \eqref{eqnevent1}, it follows from the uniqueness of the optimal policy (see A6) that
\begin{eqnarray*}
    &&\phi_L^\top(\pi^{opt}(s),s))\widehat{\beta}_{[T_1,T_2]}-\max_{a\neq \pi^{opt}(s)}\phi_L^\top(a,s))\widehat{\beta}_{[T_1,T_2]}\\
    &\ge &Q^{opt}(\pi^{opt}(s),s)-\max_{a\neq \pi^{opt}(s)}Q^{opt}(a,s)-\frac{2m_0}{3}\ge \frac{m_0}{3},
\end{eqnarray*}
for any $a$, $s$, $T_1$ and $T_2$. This leads to
\begin{eqnarray}\label{eqn:widehatpiopt}
    \pi^{opt}(s)=\pi_{\beta^{(k)}}(s)=\argmax_{a} \phi_L^\top(a,s)\beta^{(k)},\qquad \forall k\ge K/2,
\end{eqnarray}  
with probability $1-O(N^{-1}T^{-1})$. On the event defined in \eqref{eqn:widehatpiopt}, it follows from the definition of $\beta^{(k)}$ (see \eqref{eqn:betakdefinition}) that 
\begin{eqnarray*}
    &&\beta^{(k)}=\underbrace{\sum_{i=1}^N\sum_{t=T_1}^{T_2-1}\frac{\widehat{\Sigma}_{[T_1,T_2]}^{-1}}{N(T_2-T_1)}\phi_L(A_{i,t},S_{i,t})R_{i,t}}_{\widehat{\beta}_r}\\
    &+&\gamma \underbrace{\sum_{i=1}^N\sum_{t=T_1}^{T_2-1}\frac{\widehat{\Sigma}_{[T_1,T_2]}^{-1}}{N(T_2-T_1)}\phi_L(A_{i,t},S_{i,t})\phi_L^\top(\pi^{\tiny{opt}}(S_{i,t+1}),S_{i,t+1})}_{\widehat{\mathcal{P}}} \beta^{(k-1)},
\end{eqnarray*}
for all $k>K/2$. By iteratively applying this argument, we obtain that $\beta^{(K)}=\sum_{k=1}^{K/2} \gamma^{k-1}\widehat{\mathcal{P}}^{k-1}\widehat{\beta}_r +\gamma^{K/2} \widehat{\mathcal{P}}^{K/2} \beta^{(K/2)}$ and hence
\begin{eqnarray*}
    \phi_L^\top(a,s)\beta^{(K)}=\sum_{k=1}^{K/2} \gamma^{k-1}\phi_L^\top(a,s)\widehat{\mathcal{P}}^{k-1}\widehat{\beta}_r +\gamma^{K/2} \phi_L^\top(a,s)\widehat{\mathcal{P}}^{K/2} \beta^{(K/2)}.
\end{eqnarray*}
In view of \eqref{eqn:phibetastariterative2}, we obtain that
\begin{eqnarray}\label{eqn:phibetaK-betastar}
\begin{split}
    \phi_L^\top(a,s)(\beta^{(K)}-\beta^*)=\sum_{k=1}^{K/2} \gamma^{k-1}\phi_L^\top(a,s)(\widehat{\mathcal{P}}^{k-1}\widehat{\beta}_r-\mathcal{P}^{k-1}\beta_r)\\
    +\gamma^{K/2} \phi_L^\top(a,s)(\widehat{\mathcal{P}}^{K/2} \beta^{(K/2)}-\mathcal{P}^{K/2} \beta^{\tiny{*}})+O\Big(\frac{L^{-p/d}}{(1-\gamma)^3}\Big).
\end{split}
\end{eqnarray}
Using similar arguments to the proof of \eqref{eqn:operatorgammaSigmaPSigma}, it can be shown that $\|\gamma \Sigma_{[T_1,T_2]}^{1/2}\widehat{\mathcal{P}}\Sigma_{[T_1,T_2]}^{-1/2}\|_2$ is also strictly smaller than $1$, with probability at least $1-O(N^{-1}T^{-1})$. As $K\gg \log(NT)$ (Assumption A8) and $\beta^{(k)}$s are bounded (see \eqref{eqn:betakbetakstarupperbound}), $\beta^{(K/2)}$ in the first term on the second line of \eqref{eqn:phibetaK-betastar} can be replaced with $\beta^*$, and the resulting error can be made arbitrarily small, specifically of the order $O((NT)^{-c})$ for any sufficiently large constant $c>0$. In view of the condition on $L$ in A9, we obtain that
\begin{eqnarray*}
    \phi_L^\top(a,s)(\beta^{(K)}-\beta^*)=\sum_{k=1}^{K/2} \gamma^{k-1}\phi_L^\top(a,s)(\widehat{\mathcal{P}}^{k-1}\widehat{\beta}_r-\mathcal{P}^{k-1}\beta_r)\\
    +\gamma^{K/2} \phi_L^\top(a,s)(\widehat{\mathcal{P}}^{K/2} \beta^{*}-\mathcal{P}^{K/2} \beta^{*})+O\Big(\frac{L^{-p/d}}{(1-\gamma)^3}\Big).
\end{eqnarray*}
With some calculations, the first two terms on the RHS can be shown to equal 
\begin{eqnarray}\label{eqn:alpharhat-alphar}
    \sum_{k=1}^{K/2} \gamma^{k-1}\phi_L^\top(a,s)(\widehat{\mathcal{P}}^{k-1}\widehat{\alpha}_r-\mathcal{P}^{k-1}\alpha_r)
\end{eqnarray}
where
\begin{eqnarray*}
    \widehat{\alpha}_r&=&\sum_{i=1}^N\sum_{t=T_1}^{T_2-1}\frac{\widehat{\Sigma}_{[T_1,T_2]}^{-1}}{N(T_2-T_1)}\phi_L(A_{i,t},S_{i,t})[R_{i,t}+\gamma \phi_L^\top(\pi^{\tiny{opt}}(S_{i,t+1}), S_{i,t+1})\beta^*-\phi_L^\top(A_{i,t},S_{i,t})\beta^*],\\
    \alpha_r&=&\sum_{t=T_1}^{T_2-1}\frac{\Sigma_{[T_1,T_2]}^{-1}}{T_2-T_1}\Mean \phi_L(A_{t},S_{t})[R_t+\gamma \phi_L^\top(\pi^{\tiny{opt}}(S_{t+1}), S_{t+1})\beta^*-\phi_L^\top(A_{t},S_{t})\beta^*].
\end{eqnarray*}
Using similar arguments in bounding the rightmost reminder term in \eqref{eqn:phibetastariterative1}, it can be show that $\sup_{a,s}|\phi_L^\top(a,s)\alpha_r|=O((1-\gamma)^{-2}L^{-p/d})$ and $\sup_{a,s}|\phi_L^\top(a,s)\sum_{k} \gamma^{k-1}\mathcal{P}^{k-1}\alpha_r|=O((1-\gamma)^{-3}L^{-p/d})$. Similarly, we can further replace $\widehat{\alpha}_r$ in \eqref{eqn:alpharhat-alphar} with 
\begin{eqnarray*}
    \widehat{\alpha}_r^*=\sum_{i=1}^N\sum_{t=T_1}^{T_2-1}\frac{\widehat{\Sigma}_{[T_1,T_2]}^{-1}}{N(T_2-T_1)}\phi_L(A_{i,t},S_{i,t})\delta_{i,t}^*,
\end{eqnarray*}
with the approximation error upper bounded by $O((1-\gamma)^{-3}L^{-p/d})$, where we recall that $\delta_{i,t}^*$ denotes the temporal difference error $R_{i,t}+\gamma Q^{\tiny{opt}}(\pi^{\tiny{opt}}(S_{i,t+1}),S_{i,t+1})-Q^{\tiny{opt}}(A_{i,t},S_{i,t})$. It follows that
\begin{eqnarray*}
    \phi_L^\top(a,s)(\beta^{(K)}-\beta^*)=\sum_{k=1}^{K/2} \gamma^{k-1}\phi_L^\top(a,s) \widehat{\mathcal{P}}^{k-1}\widehat{\alpha}_r^*+O\Big(\frac{L^{-p/d}}{(1-\gamma)^3}\Big).
\end{eqnarray*}
Meanwhile, using similar arguments to bounding $\|\beta^{(k)*}-\beta^{(k)}-I_1\|_2$ in the finite-sample analysis, we can further approximate the leading term in the above expression by
\begin{eqnarray*}
    \sum_{k=1}^{K/2} \gamma^{k-1}\phi_L^\top(a,s) \mathcal{P}^{k-1}\widehat{\alpha}_r^{**},
\end{eqnarray*}
with the approximation error bounded by $O((1-\gamma)^{-3}L^{3/2}\log(NT)/(\epsilon NT))$, where
\begin{eqnarray*}
    \widehat{\alpha}_r^{**}=\sum_{i=1}^N\sum_{t=T_1}^{T_2-1}\frac{\Sigma_{[T_1,T_2]}^{-1}}{N(T_2-T_1)}\phi_L(A_{i,t},S_{i,t})\delta_{i,t}^*.
\end{eqnarray*}
Now, the assertion in \eqref{eqn:asymptoticexpansion} can be readily obtained by noting that the matrix $\sum_{k=1}^{K/2} \gamma^{k-1} \mathcal{P}^{k-1}$ can be well-approximated by $(I-\gamma \mathcal{P})^{-1}$, and that $\sup_{a,s}|\phi_L^\top(a,s)\beta^*-Q^{\tiny{opt}}(a,s)|=O((1-\gamma)^{-2}L^{-p/d})$.

\subsection{Proof of Lemma \ref{lemmamatrixnonstat}}\label{sec:prooflemma2}
We focus on establishing a uniform upper error bound for $\{\|\widehat{W}_{[T_1,T_2]}-W_{[T_1,T_2]}\|_2: T_2-T_1\ge \epsilon T \}$ in this section. The assertion that $\|W_{[T_1,T_2]}^{-1}\|_2\le \bar{c}(1-\gamma)^{-1}$ can be proven by Lemma 3 of \cite{shi2022statistical}.

In Step 3 of the proof of Theorem \ref{thm:size}, we have shown that 
$\argmax_{a} \phi_L^\top(a,s) \widehat{\beta}_{[T_1,T_2]}=\argmax_a Q^{\tiny{opt}}(a,s)$ and hence $\pi_{\widehat{\beta}_{[T_1,T_2]}}=\pi^{\tiny{opt}}$, WPA1. It follows that
\begin{eqnarray}\label{eqn:firststep}
	\widehat{W}_{[T_1,T_2]}=\frac{1}{N(T_2-T_1)}\sum_{i=1}^N\sum_{t=T_1}^{T_2-1}  \phi_{L}(A_{i,t},S_{i,t})\{\phi_{L}(A_{i,t},S_{i,t})-\gamma \phi_{L}(\pi^{\tiny{opt}}(S_{i,t+1}),S_{i,t+1})\}^\top.
\end{eqnarray}

We next provide an upper bound on the difference between the RHS of \eqref{eqn:firststep} and $W_{[T_1,T_2]}$. Define $\widehat{W}_{[T_1,T_2]}^*$ as 
\begin{eqnarray*}
	\frac{1}{N(T_2-T_1)}\sum_{i=1}^N\sum_{t=T_1}^{T_2-1} \sum_a \pi^b(a|S_{i,t}) \phi_{L}(a,S_{i,t})[\phi_{L}(a,S_{i,t})-\gamma \Mean \{\phi_{L}(\pi^{\tiny{opt}}(S_{i,t+1}),S_{i,t+1})|A_{i,t}=a,S_{i,t}\}]^\top.
\end{eqnarray*}
The difference $\|\widehat{W}_{[T_1,T_2]}-W_{[T_1,T_2]}\|_2$ can be upper bounded by $\|\widehat{W}_{[T_1,T_2]}-\widehat{W}^*_{[T_1,T_2]}\|_2+\|\widehat{W}^*_{[T_1,T_2]}-W_{[T_1,T_2]}\|_2$. 

Under the conditional mean independent assumption on the reward \eqref{eqn:reward}, the first term $\widehat{W}_{[T_1,T_2]}-\widehat{W}^*_{[T_1,T_2]}$ corresponds to a sum of martingale difference. Using similar arguments to the proof of Lemma 3 of \cite{shi2022statistical}, we can show that the first term is of the order $O(\sqrt{(\epsilon NT)^{-1}L \log (NT)})$, with probability at least $1-O\{(NT^{-3})\}$, under the condition that $T_2-T_1\ge \epsilon T$. See also, Freedman's inequality for matrix martingales developed by \cite{tropp2011freedman}. 
It follows from Bonferroni's inequality that $\sup_{T_1,T_2} \|\widehat{W}_{[T_1,T_2]}-\widehat{W}^*_{[T_1,T_2]}\|_2=O(\sqrt{(\epsilon NT)^{-1}L \log (NT)})$, with probability at least $1-O\{(NT)^{-1}\}$. 

It remains to bound $\|\widehat{W}^*_{[T_1,T_2]}-W_{[T_1,T_2]}\|_2$. %Without loss of generality, assume $N=1$. 
Let $\Gamma_0$ be an $\varepsilon$-net of the unit sphere in $\mathbb{R}^L$ that satisfies the following: for any $\nu\in \mathbb{R}^L$ with unit $\ell_2$ norm, there exists some $\nu_0\in \Gamma_0$ such that $\|\nu-\nu_0\|_2\le \varepsilon$. Set $\varepsilon=(NT)^{-2}$. According to Lemma 2.3 of \cite{mendelson2008uniform}, there exists such an $\varepsilon$-net $\Gamma_0$ that belongs to the unit sphere and satisfies $|\Gamma_0|\le 5^L (NT)^{2L}$.

For any $\nu_1,\nu_2$ with unit $\ell_2$ norm, define 
\begin{eqnarray*}
	\Psi_t(a,s,\nu_1,\nu_2)=\pi_t^b(a|s) \nu_1^\top \phi_L(a,s) [\phi_L(a,s)-\gamma \Mean \{ \phi_L(\pi^{\tiny{opt}}(S_{t+1}),S_{t+1})|A_t=a,S_t=s\} ]^\top \nu_2.
\end{eqnarray*}
The difference $\|\widehat{W}^*_{[T_1,T_2]}-W_{[T_1,T_2]}\|_2$ can be represented as
\begin{eqnarray*}
	\sup_{\|\nu_1\|_2=\|\nu_2\|_2=1} \left|\frac{1}{N(T_2-T_1)}\sum_{i=1}^N\sum_{t=T_1}^{T_2-1}\sum_a \{\Psi_t(a,S_{i,t},\nu_1,\nu_2)-\Mean \Psi_t(a,S_{i,t},\nu_1,\nu_2)\} \right|.
\end{eqnarray*} 
We first show that $\Psi_t(a,s,\nu_1,\nu_2)$ 
%
%%It follows from Lemma 2.3 of \cite{mendelson2008uniform} that there exists such an $\varepsilon$-net such that $|\mathcal{S}_0|\le $
%
%We aim to apply the concentration inequality developed by \cite{alquier2019exponential}. Toward that end, we aim to show that 
%\begin{eqnarray*}
%	\Psi(a,s)=\pi^b(a|s) \phi_L(a,s) [\phi_L(a,s)-\gamma \Mean \{ \phi_L(\pi^{\tiny{opt}}(S_{t+1}),S_{t+1})|A_t=a,S_t=s\} ]^\top
%\end{eqnarray*}
is a Lipschitz continuous function of $\nu_1$ and $\nu_2$. For any $\nu_1,\nu_2,\nu_3,\nu_4$, the difference $\Psi_t(a,s,\nu_1,\nu_2)-\Psi_t(a,s,\nu_3,\nu_4)$ can be decomposed into the sum of the following two terms:
\begin{eqnarray*}
	\pi^b_t(a|s) (\nu_1-\nu_3)^\top \phi_L(a,s) [\phi_L(a,s)-\gamma \Mean \{ \phi_L(\pi^{\tiny{opt}}(S_{t+1}),S_{t+1})|A_t=a,S_t=s\} ]^\top \nu_2\\
	+\pi^b_t(a|s_2) \nu_3^\top \phi_L(a,s)  [\phi_L(a,s)-\gamma \Mean \{ \phi_L(\pi^{\tiny{opt}}(S_{t+1}),S_{t+1})|A_t=a,S_t=s\} ]^\top (\nu_2-\nu_4).
	%\\+\pi^b_t(a|s_2) \nu_1^\top \phi_L(a,s_2) \{\phi_L(a,s_1)-\phi_L(a,s_2)\}^\top\nu_2+\gamma \pi^b_t(a|s_2) \nu_1^\top \phi_L(a,s_2)  \\\times [\Mean \{ \phi_L(\pi^{\tiny{opt}}(S_{t+1}),S_{t+1})|A_t=a,S_t=s_2\}-\Mean \{ \phi_L(\pi^{\tiny{opt}}(S_{t+1}),S_{t+1})|A_t=a,S_t=s_1\} ]^\top \nu_2.
\end{eqnarray*}
The first term is $O(L)\|\nu_1-\nu_3\|_2$ according to \eqref{eqn:basis2.5}. Similarly, the second term is $O(L\|\nu_2-\nu_4\|_2)$. 
%and the fact that $\pi^b_t(a|\bullet)$ is a Lipschitz continuous function in (A5). Since $\Phi_L$ is Lipschitz continuous, so is $\phi_L$. By (A4), the second and the third terms are $O(L)$ as well. Finally, notice that the transition density function $p_t$ is Lipschitz continuous under (A2). It follows that the last term is $O(L)$ as well.
To summarize, we have shown that
%\begin{eqnarray}\label{eqn:Lip}
%	|\Psi_t(a,s_1,\nu_1,\nu_2)-\Psi_t(a,s_2,\nu_1,\nu_2)|\le c_2 L \|s_1-s_2\|_2,
%\end{eqnarray} 
%for some constant $c_2>0$. Similarly, we can show that 
\begin{eqnarray*}
	|\Psi_t(a,s_1,\nu_1,\nu_2)-\Psi_t(a,s_2,\nu_3,\nu_4)|\le c L (\|\nu_1-\nu_3\|_2 + \|\nu_2-\nu_4\|_2),
\end{eqnarray*}
for some constant $c>0$. 

For any $\nu_1,\nu_2$ with unit $\ell_2$-norm, there exist $\nu_{1,0}, \nu_{2,0}\in \Gamma_0$ that satisfy $\|\nu_1-\nu_{1,0}\|_2\le \varepsilon$ and $\|\nu_2-\nu_{2,0}\|_2\le \varepsilon$. As such, $\Psi_t(a,s_1,\nu_1,\nu_2)-\Psi_t(a,s_2,\nu_1,\nu_2)$ can be upper bounded by
\begin{eqnarray*}
	\sup_{\nu_{1,0},\nu_{2,0}\in \Gamma_0} \left|\frac{1}{N(T_2-T_1)}\sum_{i=1}^N\sum_{t=T_1}^{T_2-1}\sum_a \{\Psi_t(a,S_{i,t},\nu_{1,0},\nu_{2,0})-\Mean \Psi_t(a,S_{i,t},\nu_{1,0},\nu_{2,0})\} \right|+\frac{2c L}{(NT)^2}.
\end{eqnarray*}

It remains to establish a uniform upper bound for the first term. We aim to apply the concentration inequality developed by \cite{alquier2019exponential}. However, a direct application of Theorem 3.1 in \cite{alquier2019exponential} would yield a sub-optimal bound. This is because each summand $\Psi_t(a,S_t,\nu_{1,0},\nu_{2,0})$ is not bounded, since $\|\phi_L\|_2$ is proportional to $L^{1/2}$. To obtain a sharper bound, we further decompose the first term into the sum of the following two terms:
\begin{eqnarray}\label{eqn:twoterms}
\begin{aligned}
	\sup_{\nu_{1,0},\nu_{2,0}\in \Gamma_0} \left|\frac{1}{N(T_2-T_1)}\sum_{i=1}^N\sum_{t=T_1}^{T_2-1}\sum_a [\Psi_t(a,S_{i,t},\nu_{1,0},\nu_{2,0})-\Mean \{\Psi_t(a,S_{i,t},\nu_{1,0},\nu_{2,0})|S_{i,t-1}\}] \right|\\
	+\sup_{\nu_{1,0},\nu_{2,0}\in \Gamma_0} \left|\frac{1}{N(T_2-T_1)}\sum_{i=1}^N\sum_{t=T_1}^{T_2-1}\sum_a [\Mean \{\Psi_t(a,S_{i,t},\nu_{1,0},\nu_{2,0})|S_{i,t-1}\}-\Mean \Psi_t(a,S_{i,t},\nu_{1,0},\nu_{2,0})] \right|.
\end{aligned}	
\end{eqnarray}

The first term corresponds to a sum of martingale difference. Using similar arguments in showing $\sup_{T_1,T_2} \|\widehat{W}_{[T_1,T_2]}-\widehat{W}^*_{[T_1,T_2]}\|_2=O(\sqrt{(\epsilon NT)^{-1}L \log (NT)})$, we can show that the first term in \eqref{eqn:twoterms} is of the order $O(\sqrt{(\epsilon NT)^{-1}L \log (NT)})$, with probability at least $1-O(N^{-1} T^{-1})$, where the big-$O$ term is uniform in $\{(T_1,T_2):T_2-T_1\ge \epsilon T\}$. 

As for the second term, notice that by definition, $\Mean \{\Psi_t(a,S_t,\nu_{1,0},\nu_{2,0})|S_{t-1}=s\}$ equals
\begin{eqnarray*}
	\int_{s'} \pi^b_t(a|s') \nu_{1,0}^\top \phi_L(a,s') [\phi_L(a,s')-\gamma \Mean \{ \phi_L(\pi^{\tiny{opt}}(S_{t+1}),S_{t+1})|A_t=a,S_t=s'\} ]^\top \nu_{2,0} p_{t-1}(s'|a,s)ds'.
\end{eqnarray*}
Given that $p_t(s'|a,\bullet)$s are Lipschitz continuous on $\mathcal{S}$, $\Mean \{\Psi_t(a,S_t,\nu_{1,0},\nu_{2,0})|S_{t-1}=s\}$ is a Lipschitz continuous function of $s$. However, unlike $\Psi_t(a,s,\nu_{1,0},\nu_{2,0})$, the integrand 
\begin{eqnarray*}
	|\pi^b_t(a|s') \nu_{1,0}^\top \phi_L(a,s') [\phi_L(a,s')-\gamma \Mean \{ \phi_L(\pi^{\tiny{opt}}(S_{t+1}),S_{t+1})|A_t=a,S_t=s'\} ]^\top \nu_{2,0}|
\end{eqnarray*}
is upper bounded by a constant; see e.g., Equation (E.77) of \cite{shi2022statistical}. As such, the Lipschitz constant is uniformly bounded by some constant. 
%Without loss of generality, assume $N=1$. 
Consequently, the conditions in the statement of Theorem 3.1 in \cite{alquier2019exponential} are satisfied. We can apply Theorem 3.1 to the mean zero random variable 
\begin{eqnarray*}
	\frac{1}{N(T_2-T_1)}\sum_{i=1}^N\sum_{t=T_1}^{T_2-1}\sum_a [\Mean \{\Psi_t(a,S_{i,t},\nu_{1,0},\nu_{2,0})|S_{i,t-1}\}-\Mean \Psi_t(a,S_{i,t},\nu_{1,0},\nu_{2,0})],
\end{eqnarray*}
for each combination of $\nu_{1,0},\nu_{2,0},T_1,T_2$, and 
%$L^{-1}(\widehat{W}_{[T_1,T_2]}^*-W_{[T_1,T_2]})$ and 
show that it is of the order $O(\sqrt{\epsilon L (NT)^{-1} \log (NT)})$ with probability at least $1-O(N^{-CL}T^{-CL})$, for some sufficiently large constant $C>0$. By Bonferroni's inequality, we can show that
\begin{eqnarray*}
	\sup_{T_2-T_1\ge \epsilon T}\sup_{\nu_{1,0},\nu_{2,0}\in \Gamma_0} \left|\frac{1}{N(T_2-T_1)}\sum_{i=1}^N\sum_{t=T_1}^{T_2-1}\sum_a [\Psi_t(a,S_{i,t},\nu_{1,0},\nu_{2,0})-\Mean \{\Psi_t(a,S_{i,t},\nu_{1,0},\nu_{2,0})|S_{i,t-1}\}] \right|,
\end{eqnarray*}
is upper bounded by $O(\sqrt{ L \epsilon^{-1}(NT)^{-1} \log (NT)})$ with probability at least $1-O(N^{-1}T^{-1})$. 
 
\subsection{Proof of Theorem \ref{thm:power}}\label{sec:proofthmpower}
Without loss of generality, assume $T_0=0$. We first consider the $\ell_1$-type test. %The proof for the maximum-type tests can be similarly derived. 
%\subsubsection{$\ell_1$-Type Test}
Under the given conditions on $T^*$, we have
\begin{eqnarray}\label{eqn:TS1}
\begin{aligned}
	\textrm{TS}_1\ge \sqrt{\frac{T^*(T-T^*)}{T^2}}\left\{\frac{1}{NT}\sum_{i,t}|\widehat{Q}_{[0,T^*]}(A_{i,t},S_{i,t})-\widehat{Q}_{[T^*,T]}(A_{i,t},S_{i,t}) |\right\}\\
	\ge \sqrt{\epsilon (1-\epsilon)}\left\{\frac{1}{NT}\sum_{i,t}|\widehat{Q}_{[0,T^*]}(A_{i,t},S_{i,t})-\widehat{Q}_{[T^*,T]}(A_{i,t},S_{i,t}) |\right\}.
\end{aligned}	
\end{eqnarray}
Similar to Lemma \ref{lemmaQ}, we can show that
\begin{eqnarray}\label{eqn:QrateH1}
	\sup_{a,s}\max(|\widehat{Q}_{[0,T^*]}(a,s)-Q_{0}^{\tiny{opt}}(a,s)|,|\widehat{Q}_{[T^*,T]}(a,s)-Q_{T}^{\tiny{opt}}(a,s)|)=O(\kappa),%O\Big(\frac{L^{-p/d}}{(1-\gamma)^2}\Big)+O\Big(\frac{\sqrt{L(\epsilon NT)^{-1}\log (NT)}}{(1-\gamma)^2}\Big)
\end{eqnarray} 
where we recall that $\kappa=(1-\gamma)^{-2}(L^{-p/d}+\sqrt{L(\epsilon NT)^{-1}\log(NT)})$. 
This together with \eqref{eqn:TS1} yields that
\begin{eqnarray}\label{eqn:TS1lowerbound}
\begin{split}
    \textrm{TS}_1\ge \sqrt{\epsilon (1-\epsilon)}  \frac{1}{NT}\sum_{i=1}^N\sum_{t=0}^{T-1}|Q_0^{\tiny{opt}}(A_{i,t},S_{i,t}) -Q_T^{\tiny{opt}}(A_{i,t},S_{i,t})|\\+\sqrt{\epsilon (1-\epsilon)}O\Big(\frac{\sqrt{ L\log (NT)}}{(1-\gamma)^2\sqrt{\epsilon NT}}\Big)+\sqrt{\epsilon (1-\epsilon)}O\Big(\frac{L^{-p/d}}{(1-\gamma)^2}\Big),
\end{split}
\end{eqnarray}
with probability at least $1-O(N^{-1}T^{-1})$. Using similar arguments to proving the size property of the $\ell_1$-type test in Section \ref{sec:proofthmsize}, we can show that
\begin{eqnarray*}
     &&\frac{1}{T}\sum_{t=0}^{T-1}\sum_a \int_s |Q_0^{\tiny{opt}}(a,s)-Q_T^{\tiny{opt}}(a,s)|\pi_t^b(a|s)p_t^b(s) ds\\&=&
     \frac{1}{NT}\sum_{i=1}^N\sum_{t=0}^{T-1}|Q_0^{\tiny{opt}}(A_{i,t},S_{i,t})-Q_T^{\tiny{opt}}(A_{i,t},S_{i,t})|+O\Big(\frac{\sqrt{ \log (NT)}}{(1-\gamma)NT}\Big),
\end{eqnarray*}
with probability at least $1-O(N^{-1}T^{-1})$, which together with \eqref{eqn:TS1lowerbound} leads to
\begin{eqnarray*}
	\textrm{TS}_1\ge \frac{\sqrt{\epsilon (1-\epsilon)}}{T} \sum_{t=0}^{T-1}\sum_a \int_s |Q_0^{\tiny{opt}}(a,s)-Q_T^{\tiny{opt}}(a,s)|\pi_t^b(a|s)p_t^b(s) ds\\+O\Big(\frac{\sqrt{L(NT)^{-1}\log (NT)}}{(1-\gamma)^2}\Big)+O\Big(\frac{\sqrt{\epsilon} L^{-p/d}}{(1-\gamma)^2}\Big),
\end{eqnarray*}
with probability at least $1-O(N^{-1}T^{-1})$. In addition, using similar arguments to the proof of Theorem \ref{thm:size}, we can show that the bootstrapped test statistic $\textrm{TS}_1^b$ is upper bounded by $O((1-\gamma)^{-2}\sqrt{L(NT)^{-1}\log (NT)})$, with probability at least $1-O(N^{-1}T^{-1})$. Under the given condition on $\Delta_1$, $\textrm{TS}_1$ is much larger than the upper $\alpha$th quantile of $\textrm{TS}_1^b$, with probability at least $1-O(N^{-1}T^{-1})$. As such, the power of the proposed test is at least $1-O(N^{-1}T^{-1})$ under the alternative hypothesis. This completes the proof for the $\ell_1$-type test.

The power property of the unnormalized maximum-type test can be similarly established based on \eqref{eqn:QrateH1}. 

Finally, notice that the normalized test requires a weaker condition to detect the alternative hypothesis. This is due to normalization, which makes its bootstrapped test statistic upper bounded by $O(\sqrt{\log (NT)})$, as opposed to $O(\sqrt{L\log (NT)})$, 
with probability at least $1-O(N^{-1}T^{-1})$; see the proof of the size property of the normalized test in Section \ref{sec:proofthmsize}. Additionally, the normalized estimation error divided by the variance estimator
\begin{eqnarray*}
    \frac{1}{\widehat{\sigma}_{T^*}(a,s)}\max(|\widehat{Q}_{[0,T^*]}(a,s)-Q_{0}^{\tiny{opt}}(a,s)|,|\widehat{Q}_{[T^*,T]}(a,s)-Q_{T}^{\tiny{opt}}(a,s)|),
\end{eqnarray*}
can be upper bounded by $O(\sqrt{\log (NT)})+O(\widehat{\sigma}_{T^*}^{-1}(a,s) L^{-p/d})$, with probability at least $1-O(N^{-1}T^{-1})$ where the probability upper bound is uniform in $a$ and $s$. Using similar arguments in proving of the size property of the normalized test in Section \ref{sec:proofthmsize}, we can show that $\widehat{\sigma}_{T^*}(a,s)$ can be upper bounded by $\sqrt{L}(1-\gamma)^{-2}(\epsilon NT)^{-1/2}$. 
The proof is hence completed under the given condition on the signal strength. 
%\subsection{Proof of Theorem \ref{thmQ}}
%We first show the consistency of the estimated Q-function. 

%\subsection{Proof of Theorem \ref{thmQ1}}
%Theorem \ref{thmQ1} can be proven based on the uniform consistency in Theorem \ref{thmQ} and similar arguments to Step 3 of the proof of Theorem \ref{thm:size}. We omit the details to save space.
%Using similar arguments in establishing the uniform consistency of $\widehat{Q}$, we can show that 
%On the other hand, by definition, $\beta^*_{[T_1,T_2]}$ equals
%\begin{eqnarray*}
%	\left\{ \sum_{t=T_1}^{T_2-1} \Mean \phi_L(A_{t},S_{t})\phi_L(A_{t},S_{t})^\top \right\}^{-1}\left[ \sum_{t=T_1}^{T_2-1} \Mean \phi_L(A_{t},S_{t}) \{R_{t}+\gamma \phi_L^\top(\pi^{\tiny{opt}}(S_{t+1}),S_{t+1})\beta^*_{[T_1,T_2]}\} \right].
%\end{eqnarray*}

%Under (A2), for any $k$, there exists some $\beta^{(k),*}$ such that $\sup_{a,s} |Q^{(k),*}(a,s)-\phi_L^\top(a,s) \beta^{(k),*}|=O(L^{-c_2})$. Let $\widehat{\beta}^{(k)}$ be the estimated regression coefficients. It follows that
%\begin{eqnarray*}
%	\widehat{\beta}^{(k)}-\beta^{(k),*}=\left\{\sum_{i,t} \phi_L(a,s)\phi_L^\top(a,s) \right\}^{-1} 
%\end{eqnarray*}

%The assertion thus follows under the given conditions on $\Delta_1$.

%%%%%%%%%%%%%%%%%%%%%%%%%%%%%%%%%%%%%%%%%%%%%%%%%%%%%%%%%%%%%
\section{More on the numerical study}\label{sec:morenum}

\subsection{Implementation Details}\label{sec:imp}
To implement the proposed tests, the boundary removal parameter $\epsilon$ is set to 0.1; 2000 bootstrap samples are generated to compute $p$-values. The discount factor $\gamma$ is chosen from $\{0.9, 0.95\}$.  In our simulations, the state variables are continuous. 
We set the basis function $\Phi_L$ (see \eqref{eqn:basis})
% The set of basis functions $\phi_L$ is selected according to \eqref{eqn:basis}. %To save computational time, we approximate the basis expansion using random Fourier features 
% In particular, we set $\Phi$ 
to the random Fourier features following the Random Kitchen Sinks (RKS) algorithm \citep{rahimi2007random}, using \code{RBFsampler} function from the Python \code{scikit-learn} module for implementation. The bandwidth in the radial basis function (RBF) kernel is selected according to the median heuristic \citep{garreau2017large}. The number of basis functions $L$
% in $\Phi$ (denoted by $M=L/m$) 
is selected via 5-fold cross-validation. 
Specifically, 
% for each $M$, let $\Phi_M$ denote the resulting set of basis functions. 
we first divide all data trajectories into 5 non-overlapping equal-sized data subsets. Let $\mathcal{I}_{f}$ denote the $f$-th subsample and $\mathcal{I}_{f}^c$ denote its complement, $f=1,2,3,4,5$. For each combination of $f$, $L$ and the specified data interval $[T_1,T_2]$, we use FQI to compute an estimated optimal Q-function $\widehat{Q}_{f,L,[T_1,T_2]}$ using $L$ basis functions based on the data subsets in $\mathcal{I}_{f}^c\times [T_1,T_2]$. We next select $L$ that minimizes the FQI objective function,
\begin{eqnarray} \label{eq:cv:kerneldist}
	\begin{aligned}
		& \sum_{f=1}^5\sum\limits_{ \substack{ i \in \mathcal{I}_{f} }}\sum_{t=T_1}^{T_2-1}
		\left\{ R_{i, t}+ \gamma \max_a \widehat{Q}_{f,L,[T_1,T_2]} (a, S_{i,t+1}) - \widehat{Q}_{f,L,[T_1,T_2]} (A_{i,t}, S_{i,t}) \right\}^2.
	\end{aligned}
\end{eqnarray}
When the data is stationary over $[T_1,T_2]$, this criterion balances off the bias and standard deviation of the Q-function estimator.

To mitigate the randomness introduced by the random Fourier features, for each of the 100 simulation replications, we repeat our tests four times with different random seeds. This yields $p$-values $\{p_r, r = 1, \ldots, 4\}$. %for each candidate change point. 
We then employ the method developed by \citet{meinshausen2009p} to combine these $p$-values by defining
\begin{eqnarray}\label{combine:pvalue}
	p_0=\min \left(1, q_{\tau} \left\{ \tau^{-1} p_r, r = 1, \ldots, 4 \right\} \right),
\end{eqnarray}
to be the final $p$-value. Here, $\tau$ is some constant between 0 and 1, and $q_{\tau}$ is the empirical $\tau$-quantile of the $p$-values. Compared to using a single set of Fourier features, such an aggregation method reduces the type I error and increases the power of the resulting test. Our simulation results hardly change under $\tau=0.05,0.1, 0.15, 0.2$; hereafter, we report results under $\tau=0.1$.
All tests are conducted at significance level $\alpha = 0.05$.

% \begin{comment}
%%%%%%%%%%%%%%%%%%%%%%%%%%%%%%%%%%%%%%%%%%%%%%%%%%%%%%%%%%%%%

% \end{comment}
%}

\subsection{Four Synthetic Nonstationary Settings}\label{sec:smoothtransition}
The way we generate a smooth transition function is to first define a piecewise constant function, and smoothly connects the constant functions through a transformation. Specifically, let the piecewise constant function with two segments be $f(s, t) = f_1(s) I \{ t \leq T^* \} + f_2(s) I \{ t > T^* \}$, where $f_1$ and $f_2$ are functions not dependent on $t$. 

\begin{figure}[H]
    \centering
    \includegraphics[width=0.4\textwidth]{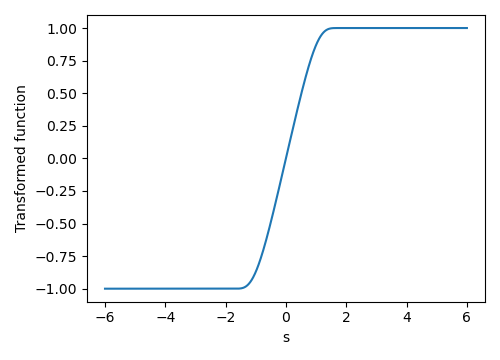}
    \caption{Smooth transformation of piecewise constant function: $f_1(s) = - I\{s \leq -2 \}$ and $f_2 (s) = I\{ s \geq 2\}$.}
    \label{fig:smoothtrans_pwconst_example}
\end{figure}

We now introduce a smooth transformation $\phi (s) = \frac{\psi(s)}{ \psi(s) + \psi(1-s) }$, where $\psi(s) = e^{-1/s} I \{ s > 0\}$. Then $g(s; f_1, f_2, s_0, s_1) := f_1 (s) + (f_2 (s) - f_1 (s)) \phi \left( \frac{s - s_0}{ s_1 - s_0} \right)$ is a smooth function from $f_1$ to $f_2$ on the interval $[s_0, s_1]$. In addition, the transformed function $\Tilde{f}(s, t) = f_1(s) I \{ t \leq T^* - \delta T \} + g(s; f_1, f_2, s_0, s_1) I \{ T^* - \delta T < t < T^* + \delta T \} + f_2(s) I \{ t > T^* \}$ is smooth in $s$. Here $\delta$ controls the smoothness of the transformation; smaller $\delta$ leads to more abrupt change and larger $\delta$ leads to smoother change. An example of $\Tilde{f}(s, t)$ is shown in Figure \ref{fig:smoothtrans_pwconst_example}.

The four simulations settings in Section \ref{sec:sim:1d} are specified as the following.
\begin{enumerate}[{(1)}]
%%%%
\item Time-homogeneous state transition function and piecewise constant reward function:
\begin{equation*} %\label{eq:transition:homo}
S_{0, t+1} = 0.5 A_{0,t} S_{0, t} + z_{0,t}, t \in [0,T].
\end{equation*}
\begin{equation*} %\label{eq:reward:constant}
R_{0,t} = 
\begin{cases}
	r_1 (S_{0,t}, A_{0,t}; t) \equiv -1.5 A_{0,t} S_{0,t}, & \text{if $t \in [0, T^*)$} \\
    r_2 (S_{0,t}, A_{0,t}; t) \equiv A_{0,t} S_{0,t}, & \text{if $t \in [T^*, T]$},
\end{cases}
\end{equation*}

%%%%
\item Time-homogeneous state transition function and smooth reward function:
\begin{equation*} %\label{eq:transition:homo}
S_{0, t+1} = 0.5 A_{0,t} S_{0, t} + z_{0,t}, t \in [0,T].
\end{equation*}
\begin{equation*} %\label{eq:reward:smooth}
R_{0,t} = \begin{cases}
    r_1 (S_{0,t}, A_{0,t}; t), & \text{if $t \in [0, T^* - \delta T)$,} \\
    g \left(S_{0,t}; r_1, r_2, T^* - \delta T, T^* \right),  & \text{if $t \in [T^* - \delta T, T^*)$,} \\
    r_2 (S_{0,t}, A_{0,t}; t), & \text{if $t \in [T^*, T]$.} \\
\end{cases}
\end{equation*}

%%%%
\item Piecewise constant state transition and time-homogeneous reward function:
\begin{equation*} %\label{eq:transition:constant}
S_{0, t+1} = 
\begin{cases}
	F_1 (S_{0,t}, A_{0,t}; t) \equiv - 0.5 A_{0,t} S_{0, t} + z_{0,t}, & \text{if $t \in [0, T^*)$,} \\
	F_2 (S_{0,t}, A_{0,t}; t) \equiv 0.5 A_{0,t} S_{0, t} + z_{0,t}, & \text{if $t \in [T^*, T]$.}
\end{cases}
\end{equation*} 
\begin{equation*} %\label{eq:rewards:homo}
R_{0,t} = 0.25 A_{0,t} S_{0,t}^2 + 4 S_{0,t}, t \in [0, T].
\end{equation*}

%%%%
\item Smooth state transition and time-homogeneous reward function:
\begin{equation*} %\label{eq:transition:constant}
S_{0, t+1} = 
\begin{cases}
	F_1 (S_{0,t}, A_{0,t}; t), & \text{if $t \in [0, T^*)$,} \\
	g \left(S_{0,t}; F_1, F_2, T^* - \delta T, T^* \right),  & \text{if $t \in [T^* - \delta T, T^*)$,} \\
	F_2 (S_{0,t}, A_{0,t}; t), & \text{if $t \in [T^*, T]$.}
\end{cases}
\end{equation*} 
\begin{equation*} %\label{eq:rewards:homo}
R_{0,t} = 0.25 A_{0,t} S_{0,t}^2 + 4 S_{0,t}, t \in [0, T].
\end{equation*}
\end{enumerate}

\subsection{Details of Baseline Methods} \label{sec:sim:evaluation:baseline}
%\textbf{Model-Based RL Context Detection}. 

\paragraph{MBCD}
The MBCD method tests whether $\tau \in [T_0, T_1]$ is a change point on the batch data collected on the interval $[T_0, T_1]$. Specifically, denote the conditional distribution of the next state and the current reward given the current state-action pair as $p_\theta (S_{i,t+1}, R_{i,t} \mid S_{i,t}, A_{i,t})$, where $\theta$ is a vector of model parameters. We let the parameter of the distribution of data on the left interval $[T_0, \tau]$ be $\theta_l$, and that of data on the right interval $[\tau, T_1]$ be $\theta_r$.
The test statistic is computed as $W_\tau = \max \left\{0, \log \frac{p_{\theta_l}(S_{i,t+1}, R_{i,t} \mid S_{i,t}, A_{i,t}; t \in [T_0, \tau]) }{ p_{\theta_r}(S_{i,t+1}, R_{i,t} \mid S_{i,t}, A_{i,t}; t \in [\tau, T_1]) } \right\}$. We reject the null hypothesis if $W_\tau > h$, where the threshold $h = |\log \alpha|$ is chosen to ensure that the false alarm rate is no larger than $\alpha$. In our implementation, we specify $h = 100$ such that $\alpha \approx 10^{-43}$ is negligible.

\sloppy To model the conditional distribution, we parameterize the conditional distribution $p_\theta (S_{i,t+1}, R_{i,t} | S_{i,t}, A_{i,t})$, using a multivariate normal distribution $\mathcal{N} \left( \mu (S_{i,t}, A_{i,t}), \Sigma (S_{i,t}, A_{i,t}) \right)$. The parameter $\theta$ contains the mean $\mu (S_{i,t}, A_{i,t})$ and the covariance $\Sigma (S_{i,t}, A_{i,t})$, which are estimated through a bootstrap ensemble of probabilistic neural networks. The number of bootstraps is set to 5, as recommended by \cite{lakshminarayanan2017simple}. Each neural network contains five layers, with 200 nodes in each hidden layer and ReLU activation function.

\paragraph{ODCP}
To implement the ODCP method \citep{padakandla2020reinforcement}, we first map the state-action pairs at each time point to compositional data, and second apply the Dirichlet likelihood test for change point detection. Specifically, for each $t$, we obtain transformed compositional data through the multi-dimensional expit function:
\begin{equation*}
X_{i,t} := \left( \frac{\exp(S_{i,t})}{1 + \exp(S_{i,t} + R_{i,t})}, \frac{\exp(R_{i,t})}{1 + \exp(S_{i,t} + R_{i,t})}, \frac{1}{1 + \exp(S_{i,t} + R_{i,t})} \right).
\end{equation*}

Next, for each candidate change point $\kappa_j \in (0, T)$, we test the null hypothesis $H_0$ that the data $\cD = \{X_{i,1}, \ldots, X_{i,T}\}_{1 \leq i \leq N}$ comes from a single Dirichlet distribution, versus the alternative hypothesis $H_1$ that the two data chunks $\cD_{:\kappa_j} = \{X_{i,1}, \ldots, X_{i,\kappa_j}\}_{1 \leq i \leq N}$ and $\cD_{\kappa_j:} = \{X_{i,\kappa_j+1}, \ldots, X_{i,T}\}_{1 \leq i \leq N}$ come from two Dirichlet distributions. Under $H_0$, we compute the Dirichlet loglikelihood as $L_0 = \log p(\{X_{i,1}, \ldots, X_{i,T}\}_{1 \leq i \leq N} \mid \eta_0)$, where $\eta_0$ is the Dirichlet maximum likelihood estimate (MLE). Under $H_1$, we compute the sum of Dirichlet loglikelihoods as $L_1 = \log p(\{X_{i,1}, \ldots, X_{i,:\kappa_j}\}_{1 \leq i \leq N} \mid \eta_L) + \log p(\{X_{i,1}, \ldots, X_{i,\kappa_j:}\}_{1 \leq i \leq N} \mid \eta_R)$, where $\eta_L$ and $\eta_R$ are the MLEs for the data chunks to the left and right of $\kappa_j$. The test statistic $Z_{\kappa_j}$ is given by $L_1 - L_0$. 

The significance of the test is performed through a permutation test. For the $w$-th permutation of the time points in $(0, T)$, $w = 1, \ldots, 500$, we compute a test statistic $Z^*_w$ following the same procedure described above. We reject the null hypothesis if the fraction $\frac{|\{w: Z^*_w > Z^*\}|}{500}$ exceeds the threshold 0.05.

\paragraph{Kernel}
To implement the kernel-based method \citep{domingues2021kernel}, at the $l$-th FQI iteration, we consider the following objective function,
\begin{eqnarray}\label{eqn:FQIkernel}
Q^{(l+1)}=\argmin_Q \sum_{i,t} K\left(\frac{T-t}{Th}\right)\left\{R_{i,t}+ \gamma \max_a Q^{(l)}(a, S_{i,t+1})-Q(A_{i,t},S_{i,t}) \right\}^2,
\end{eqnarray}
where $K(\cdot)$ denotes the Gaussian RBF basis and $h$ denotes the associated bandwidth parameter taken from the set $\{0, 0.1, 0.2,0.4,0.8,1.6\}$. According to \eqref{eqn:FQIkernel}, the kernel-based method assigns larger weights to more recent observations to deal with nonstationarity.
After we receive the $k$th data batch, we sample $B \gg T$ data slices across all individuals from $\{(S_{i,t},A_{i,t},R_{i,t},S_{i,t+1}; 1 \leq i \leq N)\}_{0 \le t < T+kL}$
with weights proportional to $K((T-t)/(Th))$ and apply the decision tree regression to these samples to solve \eqref{eqn:FQIkernel}.

%%%%%%%%%%%%%%%%%%%%%%%%%%%%%%%%%%%%%%%%%%%%%%%%%%%%%%%%%%%%%
\subsection{Additional Simulation Results} \label{app:sec:sim:result}

%\subsubsection{Additional Change Point Detection Results}
% \begin{figure}[H]
% 	\centering
% 	\includegraphics[width=\textwidth]{AoS/revision2/fig/1d_rejection_rates_N100.png} 
% 	% 		\caption{(a) \textbf{$N=25$.}}
% \caption{Empirical type I errors and powers of the proposed test and their associated 95\% confidence intervals under settings described in Section \ref{sec:sim:1d}, with $N=100$, using quantile-based aggregation method in \cite{meinshausen2009p}. Abbreviations: Homo for homogeneous, PC for piecewise constant, and Sm for smooth.}
% \label{fig:1d:rejN100}
% \end{figure}

\begin{figure}[htbp]
    \centering
    \begin{minipage}[c]{0.9\textwidth}
        \centering
        \includegraphics[width=\linewidth]{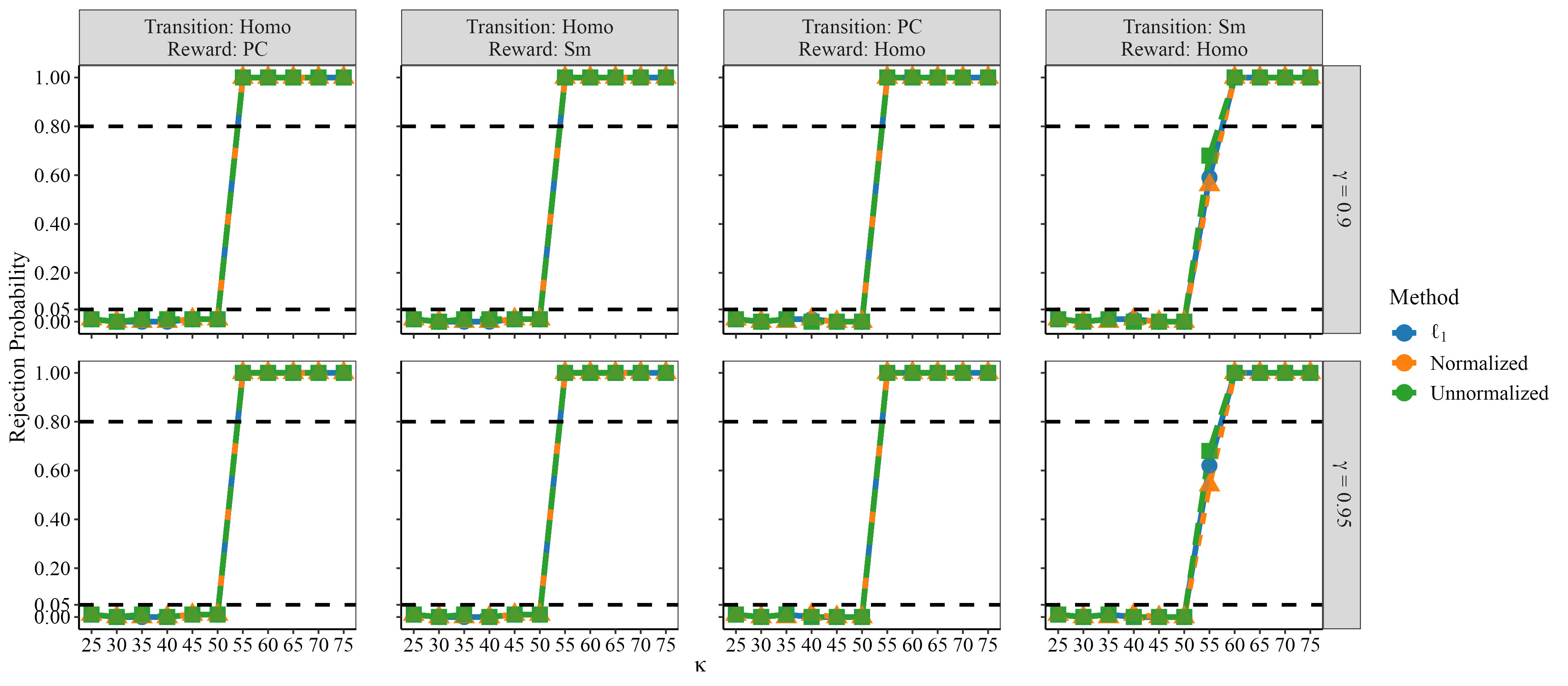}
    \end{minipage}
    \begin{minipage}[c]{0.9\textwidth}
        \centering \small{(a) Rejection probabilities.}
        \vspace{1ex}
    \end{minipage}
    \begin{minipage}[c]{0.9\textwidth}
        \centering
        \includegraphics[width=\linewidth]{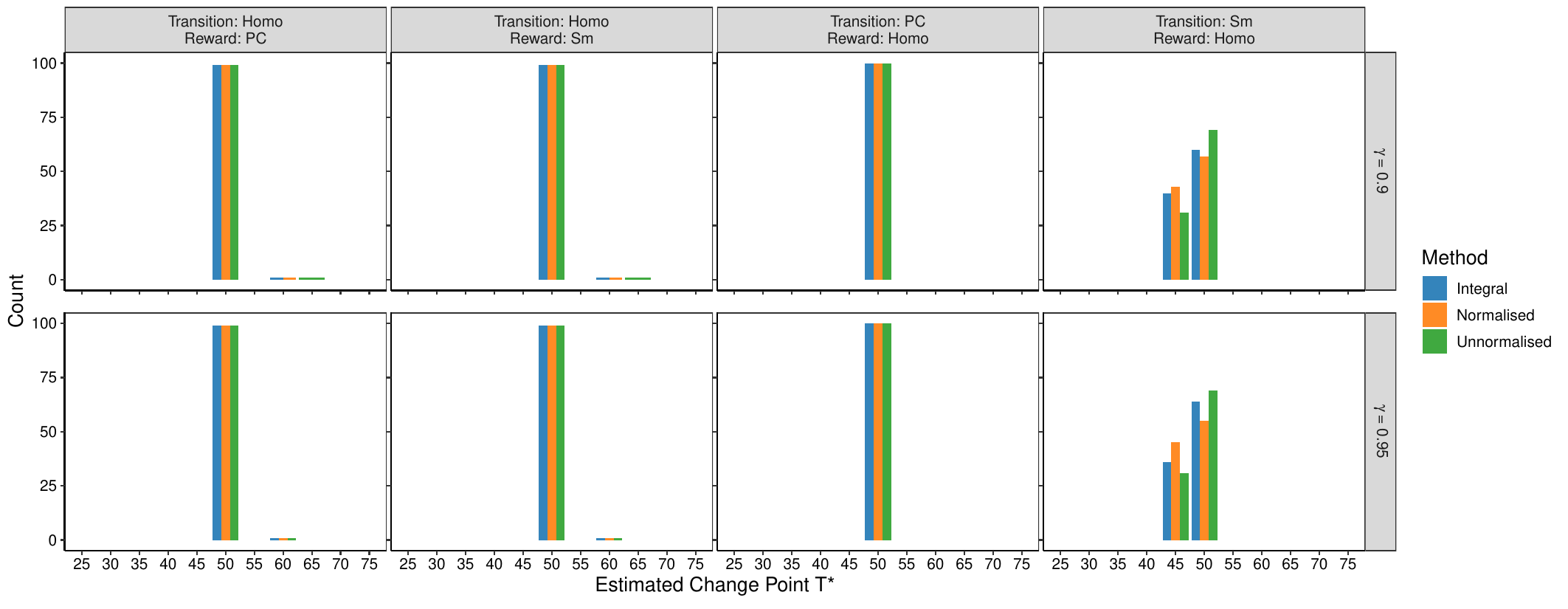}
    \end{minipage}
    \begin{minipage}[c]{0.9\textwidth}
        \centering \small{(b) Detected change points.}
        \vspace{1ex}
    \end{minipage}
	% \includegraphics[width=\textwidth]{AoS/revision2/fig/1d_bar_N100.pdf}
	%\caption{Distribution of detected change points $\hat{T}^* = T - \kappa_{j_0-1}$ with sample size $N = 25$ of each simulation scenario in Section \ref{sec:sim:1d}.}
    \caption{Empirical type I errors and powers of the proposed test and their associated 95\% confidence intervals under settings described in Section \ref{sec:sim:1d}, with $N=100$. Abbreviations: Homo for homogeneous, PC for piecewise constant, and Sm for smooth.}
    \label{fig:1d:N100}
\end{figure}

\begin{figure}[H]\label{app:fig:1d_value}
    \centering
    \begin{subfigure}{0.9\textwidth}
        \includegraphics[width=\linewidth]{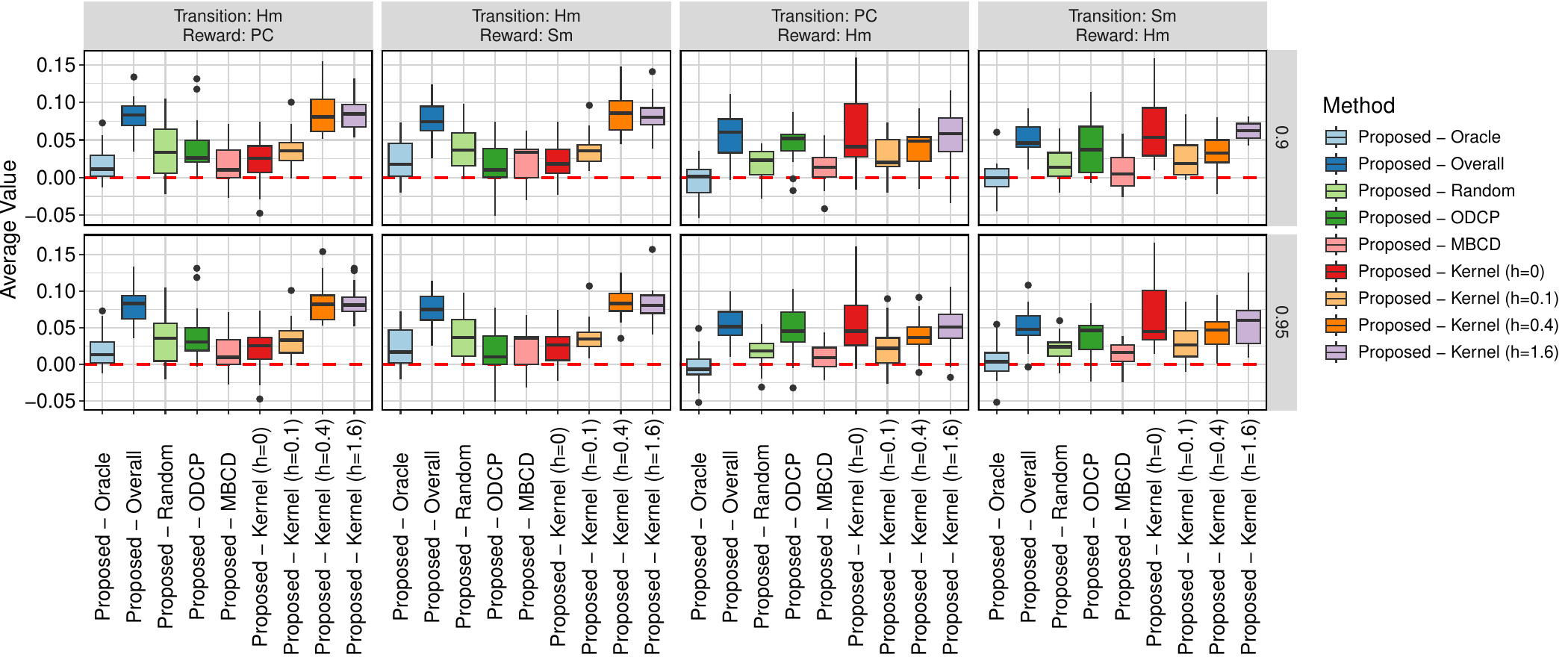}
        \subcaption{Moderate signal.}
    \end{subfigure}
\end{figure}%
\begin{figure}[H]\ContinuedFloat
    \centering
\medskip
    \begin{subfigure}{\textwidth}
        \includegraphics[width=\linewidth]{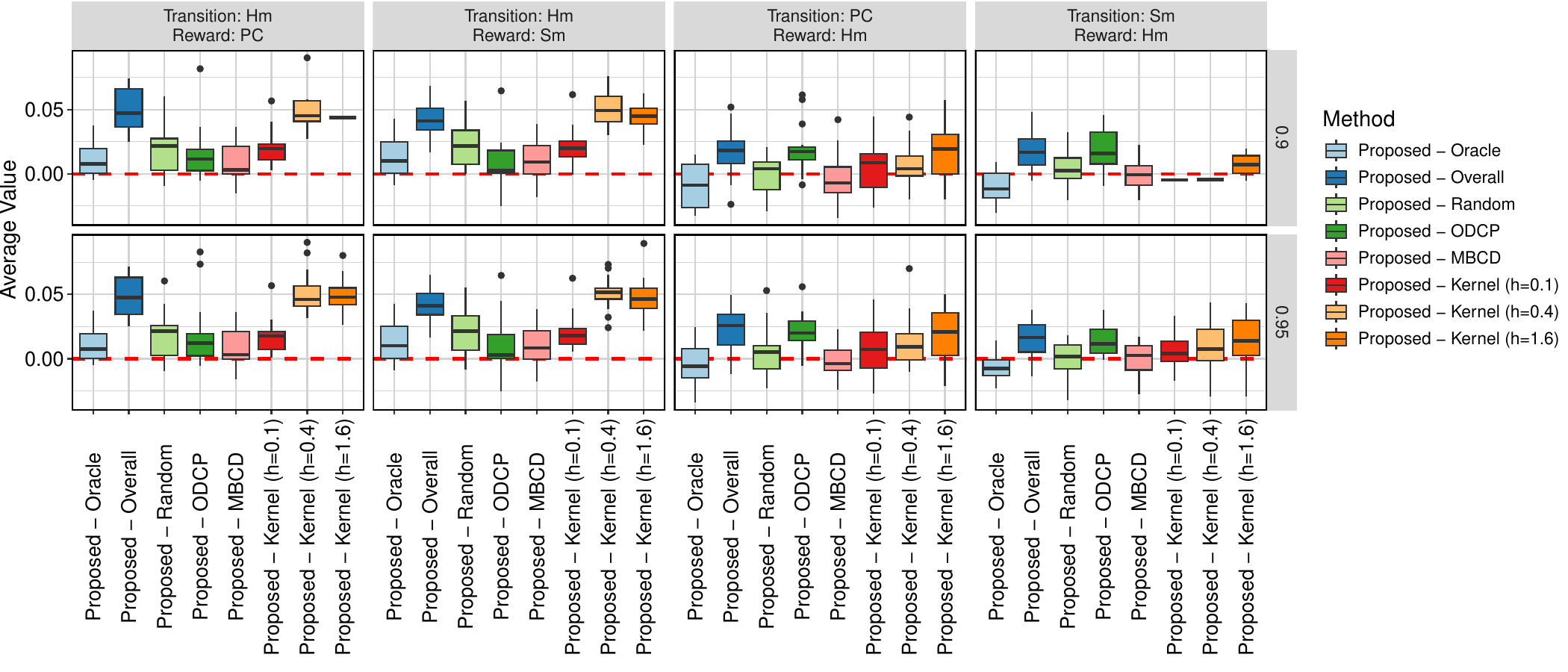}
        \subcaption{Weak signal.}
    \end{subfigure}
% \end{figure}%
% \begin{figure}[H]\ContinuedFloat
%     \centering
%     \begin{subfigure}{0.9\textwidth}
%         \includegraphics[width=\linewidth]{fig/1d_box_optvalue_online_dt_N100_moderate.pdf}
%         \subcaption{$N=100$ and moderate signal.}
%     \end{subfigure}
% \end{figure}%
% \begin{figure}[H]\ContinuedFloat
%     \centering
% \medskip
%     \begin{subfigure}{\textwidth}
%         \includegraphics[width=\linewidth]{fig/1d_box_optvalue_online_dt_N100_weak.pdf}
%         \subcaption{$N=100$ and weak signal.}
%     \end{subfigure}
\caption{Distribution of the difference between the expected return under the proposed policy and those under policies computed by other baseline methods, under settings in Section \ref{sec:sim:1d} with moderate and weak signal-to-noise ratios. The proposed policy is based on the change point detected by the $\ell_1$ type test statistic. In all scenarios, we find the value results based on the normalized or unnormalized test statistics are similar to those of the integral test statistic.}
% \label{fig:1d:value}
\end{figure}%

%\subsubsection{Comparisons of the Three tests}

\subsection{Real-data-based Simulation}\label{sec:realdatabasedsimu}
To mimic the IHS study, we simulate $N = 100$ subjects, each observed over $T = 50$ time points. Our aim is to estimate an optimal treatment policy to improve these interns' long-term physical activity levels. See Section \ref{sec:data} for more details about the study background. At time $t$, the state vector $S_{i,t}$ comprises four variables to mimic the actual IHS study: the square root of step count at time $t$ ($S_{i,t,1}$), cubic root of sleep minutes at time $t$ ($S_{i,t,2}$), mood score at time $t$ ($S_{i,t,3}$), and the square root of step count at time $t-1$ ($S_{i,t,4} = S_{i,t-1,1}$). These power transformations were applied in accordance with the approaches in \cite{necamp2020assessing}. %Under  the state transition is designed to follow an AR(2) process. 
% See Appendix \ref{sec:sim:ihs:setting} for the true parameter values that govern the dynamics. %The transformations performed on the state variables are made consistent with the real IHS data analysis in Section \ref{sec:data}. 
The actions are binary %$A_{it} \in \{-1, 1\}$ 
with $\mathbb{P}(A_{it} = 1) = 0.25$; $A_{it} = 1$ means the subject is randomized to receive activity messages at time $t$, and $A_{it} = 0$ means any other types of messages or no message at all. %To the study aim, 
Reward $R_{i,t} = S_{i,t,1}$ is defined as the step count at time $t$. We assume that the state transition function has an abrupt change point at time $T^* = 25$. %Specifically, the state transition changes from $\mathcal{T}_t (S_t, A_t, \delta)$ on $t \in [0, T^*)$ to $\mathcal{T}_t (S_t, -A_t, \delta)$ on $t \in [T^*, T]$. %; see Appendix \ref{sec:sim:ihs:setting} for the detailed setup. 
Specifically, we initiate the state variables as independent normal distributions with $S_{i,0,1} \sim \cN(20, 3)$, $S_{i,0,2} \sim \cN(20, 2)$, and $S_{i,0,3} \sim \cN(7, 1)$, and let them evolve according to
\begin{align*}
\begin{pmatrix}
S_{i,t+1,1} \\ S_{i,t+1,2} \\ S_{i,t+1,3}
\end{pmatrix} =
\bW_1 (A_{i,t}) \Tilde{S}_{i,t} I \{t \in [0, T^*)\} + \bW_2 (A_{i,t}) \Tilde{S}_{i,t} I \{t \in [T^*, T] \} + \bz_{i,t},
\end{align*}
where the transition matrices are
\begin{align*}
\bW_1 (A_{i,t}) &= \begin{pmatrix}
    10 + 0.6 A_{i,t} & 0.4 + 0.3 A_{i,t} & 0.1 - 0.1 A_{i,t} & -0.04 & 0.1 \\
    11 - 0.4 A_{i,t} & 0.05 & 0 & 0.4 & 0 \\
    1.2 - 0.5 A_{i,t} & -0.02 & 0 & 0.03 + 0.03A_{i,t} & 0.8
    \end{pmatrix}, \\
\bW_2 (A_{i,t}) &= \begin{pmatrix}
    10 - 0.6 A_{i,t} & 0.4 - 0.3 A_{i,t} & 0.1 + 0.1 A_{i,t} & 0.04 & -0.1 \\
    11 - 0.4 A_{i,t} & 0.05 & 0 & 0.4 & 0 \\
    1.2 + 0.5 A_{i,t} & -0.02 & 0 & 0.03 - 0.03A_{i,t} & 0.8
    \end{pmatrix},
\end{align*}
$\Tilde{S}_{i,t} = (1, S_{i,t,1}, S_{i,t,2}, S_{i,t,3}, S_{i,t-1,1})^\top$, and $\bz_{i,t} \sim \cN_3( 0, \mathrm{diag}(1,1,0.2))$ is random noise. 

Under this setting, the state transition function is nonstationary whereas the reward is a stationary function of the state. %Under the data generating mechanism, 
In addition, 
the data follow the null hypothesis when $\kappa = 1, \ldots, 25$ and follow the alternative hypothesis for $\kappa=26,\ldots,49$. %We start collecting data by following the specified process after an initial burn-in period of 60 time points to ensure that the simulated data are sampled from an approximately stationary distribution. 
The discount factor is set to $\gamma=0.9$ or $0.95$. We test the null hypothesis along a sequence of $\kappa = 10, 15,\ldots, 40$ for every five time points. The number of basis functions is chosen among $\{20,30,40,50,60\}$. %For each replication data set, the detected change point is set to be $\hat{T}^* = T-\kappa_{j_0-1}$, where the first rejection occurs at $\kappa_{j_0}$. %In terms of implementation of the proposed method (change point detection, post-change-point optimal policy estimation, and policy evaluation), we provide more details below for each data replication; other comparative methods are implemented the same as in Simulation I. During change point detection, RBF bases for normalized state variables are used in FQI (see Section \ref{sec:imp}). The number of basis $M$ is chosen among $\{10,15,20,25,30\}$ using 5-fold cross-validation to minimize the least squared errors. We then estimate an optimal policy using post-change-point data in $[\hat{T}^*, T]$. In particular, we use FQI but with decision tree regression for approximating the $Q$-function; denote the resulting estimated optimal policy by $\hat{\pi}^{opt}$. Finally, we calculate the value of $\hat{\pi}^{opt}$ via Monte Carlo. In particular, we simulate 300 new subjects for 200 time steps following the true post-change-point dynamics to calculate the average time-discounted cumulative rewards and treat it as the value of $\hat{\pi}^{opt}$.

\begin{figure}[t]
\centering
\begin{minipage}[t]{0.48\linewidth}
	\centering
	\includegraphics[width=\textwidth]{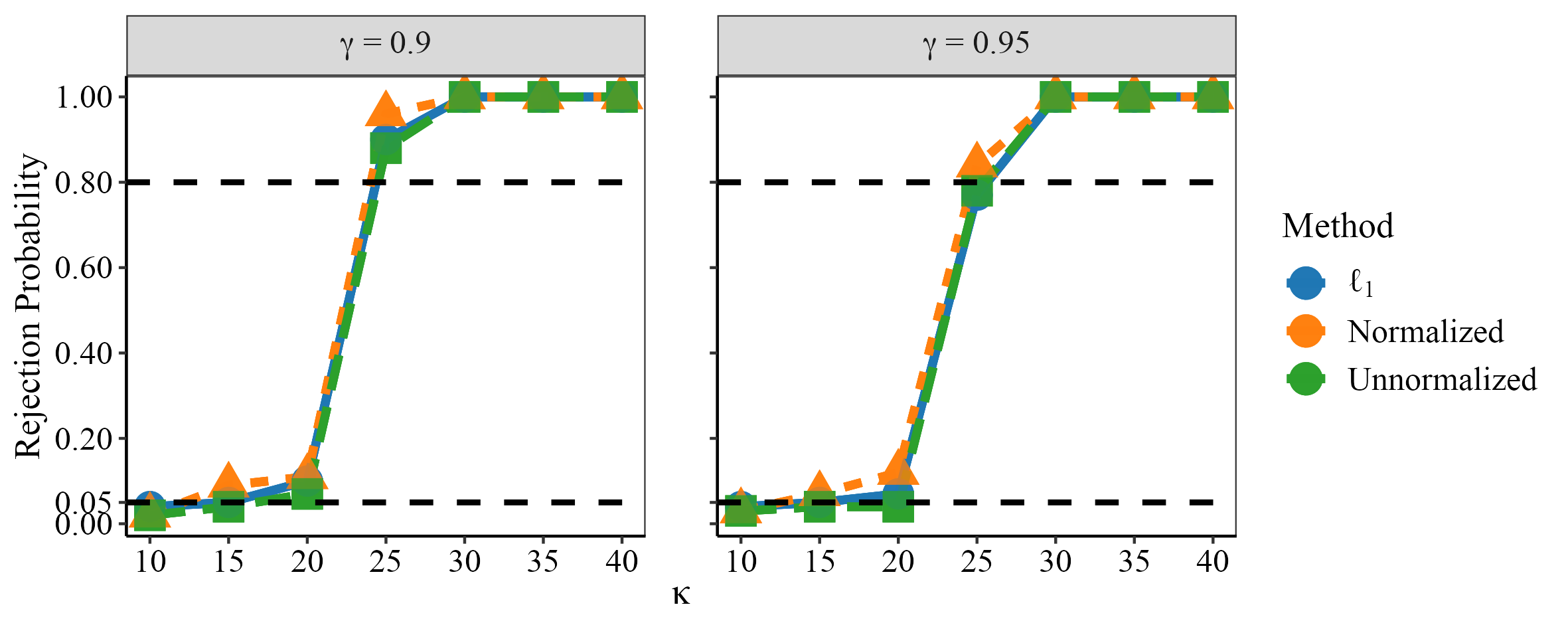} 
\end{minipage}
\begin{minipage}[t]{0.48\linewidth}
	\centering
	\includegraphics[width=\textwidth]{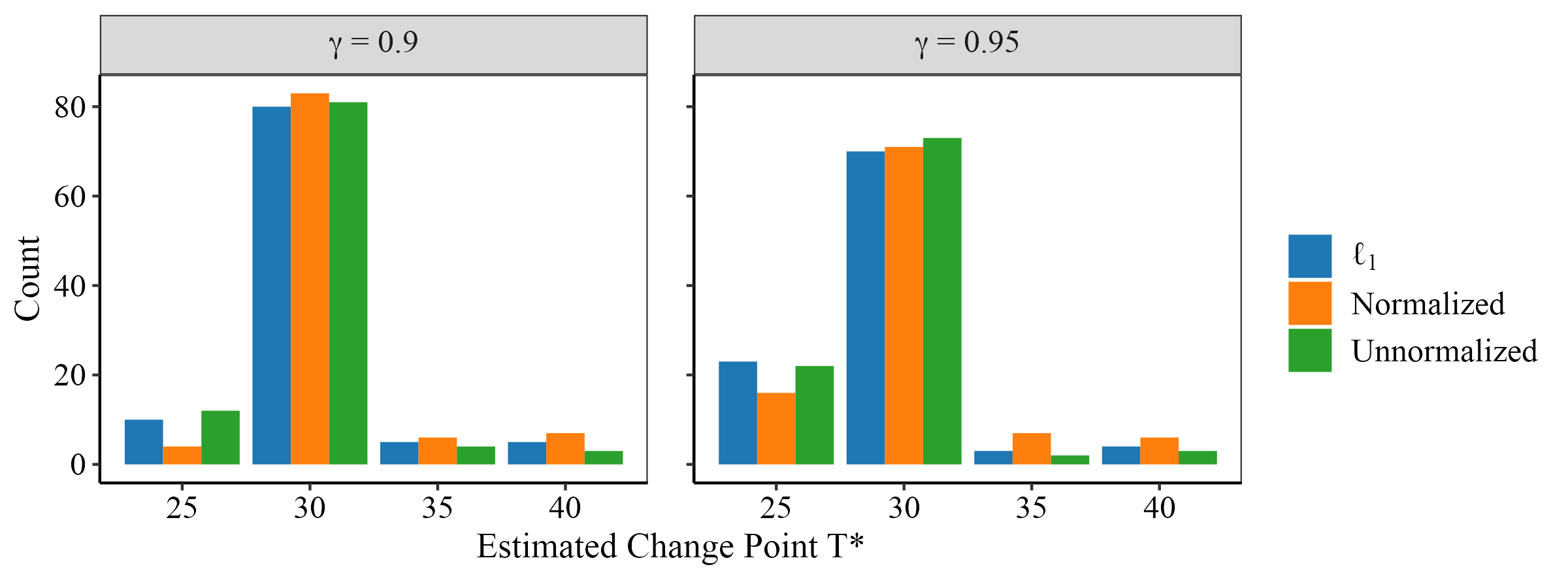} 
\end{minipage}
\begin{minipage}[t]{0.48\linewidth}
	\centering \small{(a) Type I error ($\kappa \leq 25$)
		% 		{\color{red} $\leq 25$?} 
		and power ($\kappa > 25$)
		% 		{\color{red} $> 25$?}
		.}
	\vspace{3ex}
\end{minipage}
\begin{minipage}[t]{0.48\linewidth}
	\small{(b) Estimated change point $\hat{T}^* = T - \kappa_{j_0-1}$ where the first rejection occurs at $\kappa_{j_0}$.}
	\vspace{3ex}
\end{minipage}
% 	\begin{minipage}[t]{0.8\linewidth}
	% 		\centering
	% 		\includegraphics[width=\textwidth]{fig/real_box_optreward_dt_N100.pdf} 
	% 	\end{minipage}
% 	\begin{minipage}[t]{\textwidth}
	% 		\centering \small{(c) Distribution of Monte Carlo differences of mean rewards and the overall mean rewards.}
	% 		\vspace{3ex}
	% 	\end{minipage}	
\caption{Real-data-based simulation: Empirical rejection rates of the proposed tests ($\ell_1$-type \eqref{eqn:teststat1}, normalized \eqref{eqn:teststatinf}, and unnormalized \eqref{eqn:teststatninf}) and the distribution of the estimated change points. }
\label{fig:real:result}
\end{figure}

Figure \ref{fig:real:result} shows the empirical rejection rates of the proposed tests as well as the distribution of the estimated change point location. Similar to the results in Section \ref{sec:sim:1d}, our proposed test controls the type I error at the nominal level (see $\kappa < 25$) and is powerful to detect the alternative hypothesis (see $\kappa >25$). At the true change point where $\kappa=25$ however, the proposed test fails to control the type I error. %Nonetheless, the proposed procedure yields a much better policy compared to the overall and random methods, as we show below. 
We also remark that the reason the proposed test fails at the boundary is because the marginal distribution of the first few states after the change point is very different from the stationary state distribution. After an initial burn-in period of 5 points, the proposed test is able to control the type I error at $\kappa=20$. 
%{\color{blue}[Both: please check the language here, $\kappa=25$ seems to be still under the null, but they type i error is not controlled. Need to address.]} 
%Importantly, note that in Figure \ref{fig:real:result}(b) the estimated change point $\hat{T}^* = 50$ or 60 in most cases and none reach the exact true change point $T^* = 53$, because the list of tested $\kappa$'s does not include $\kappa^* = T - T^* = 57$. However, our estimator is close enough to the truth.
In addition, the distribution of the estimated change point concentrates on $30$, which is very close to the oracle change point location $25$, implying the consistency of the proposed change point detection procedure. We remark that consistency here requires $T^{-1}|\widehat{T}^*-T^*|\stackrel{P}{\to} 0$ instead of $\prob(\widehat{T}^*=T^*)\to 1$, the latter being usually impossible to achieve in change point detection. It also demonstrates the detection delay (i.e., the estimated change point occurs later than its oracle value). %in a multi-dimensional state scenario.
% We remark that consistency here requires $T^{-1}|\widehat{T}^*-T^*|\stackrel{P}{\to} 0$ instead of $\prob(\widehat{T}^*=T^*)\to 1$, the latter being usually impossible to achieve in change-point settings.%[{\color{blue}Piotr, could you please take a look and make some edits here?}]
%{\color{blue}[Both: Need to address. Later than oracle location?]} 
%Since the approximate estimator is able to identify a subset of data on the interval $[\hat{T}^*, T]$, which are more stationary than data on the entire timeline, we obtain an optimal policy that has larger mean values than the overall or random method, as displayed in Table \ref{tab:real:value}.

%\violet{put two plots on the same row}

%%%%%%%%%%%%%%%%%%%%%%%%%%%%%%%%%%%%%
\subsection{More on the IHS Data} \label{sec:data:additional}

\begin{figure}[H]
    \centering
    \includegraphics[width=0.75\linewidth]{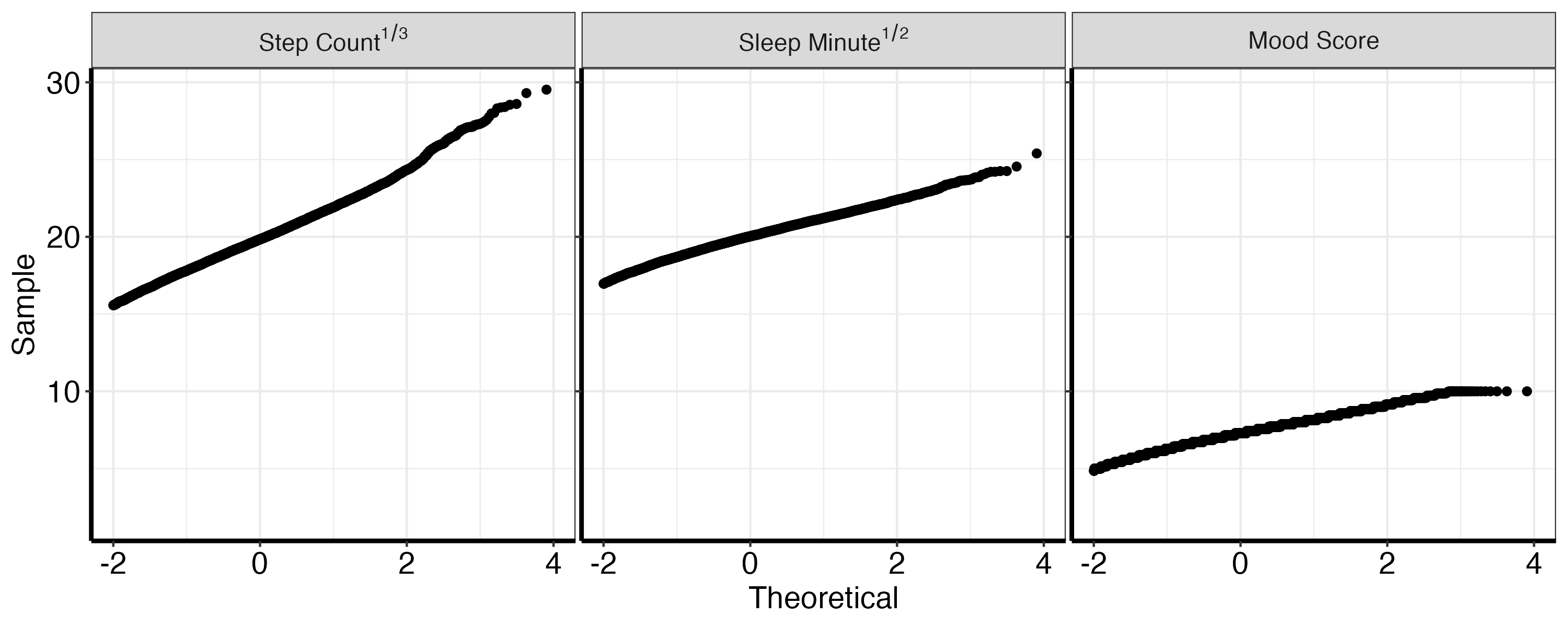}
    \caption{%\blue{
    Quantile-quantile plot of the state variables used in analyzing IHS data in Section \ref{sec:data}. After cubic root transformation of weekly average step count and square root transformation of weekly average sleep minutes, the three variables are approximately normally distributed.}%}
    \label{fig:data:qqplot}
\end{figure}

\end{document}